\documentclass[journal]{IEEEtran}

\ifCLASSOPTIONcompsoc
  \usepackage[nocompress]{cite}
\else
  \usepackage{cite}
\fi

\usepackage{amsmath}
\usepackage{algorithmic}
\usepackage{array}
\usepackage{multirow}
\usepackage[table,xcdraw]{xcolor}
\usepackage{soul}

\usepackage{url}

\usepackage{comment}
\usepackage{amsmath,amssymb} 
\usepackage{color}

\usepackage{pdflscape}

\usepackage{bm}
\usepackage{array}
\usepackage{arydshln}
\usepackage{booktabs}
\usepackage{graphicx}
\usepackage{amsmath}
\usepackage{amssymb}
\usepackage{booktabs}
\usepackage{caption}
\usepackage{subcaption}
\usepackage{float}
\definecolor{myGray}{RGB}{80,80,80}
\definecolor{myGreen}{RGB}{120, 170, 120}
\definecolor{myOrange}{RGB}{255, 120, 40}
\definecolor{myBlue}{RGB}{60, 130, 180}

\newcommand{\eg}{\textit{e.g., }}
\newcommand{\ie}{\textit{i.e., }}
\newcommand{\etal}{\textit{et al. }}

\begin{document}

\title{Training Better Deep Learning Models \\Using Human Saliency}

\author{Aidan~Boyd,~\IEEEmembership{Member,~IEEE,}
        Patrick~Tinsley,~\IEEEmembership{Member,~IEEE,} \\
        Kevin~Bowyer,~\IEEEmembership{Fellow,~IEEE,}
        and~Adam~Czajka,~\IEEEmembership{Senior~Member,~IEEE}
\IEEEcompsocitemizethanks{\IEEEcompsocthanksitem All authors associated with the Department of Computer Science Engineering, University of Notre Dame, Notre Dame,
IN, 46556.\protect\\
E-mail: aczajka @nd.edu}
}

\maketitle

\IEEEdisplaynontitleabstractindextext
\IEEEpeerreviewmaketitle

\begin{abstract}
This work explores how human judgement about salient regions of an image can be introduced into deep convolutional neural network (DCNN) training. Traditionally, training of DCNNs is purely data-driven. This often results in learning features of the data that are only coincidentally correlated with class labels. Human saliency can guide network training using our proposed new component of the loss function that ConveYs Brain Oversight to Raise Generalization (CYBORG) and penalizes the model for using non-salient regions. This mechanism produces DCNNs achieving higher accuracy and generalization compared to using the same training data without human salience. Experimental results demonstrate that CYBORG applies across multiple network architectures and problem domains (detection of synthetic faces, iris presentation attacks and anomalies in chest X-rays), while requiring significantly less data than training without human saliency guidance. Visualizations show that CYBORG-trained models' saliency is more consistent across independent training runs than traditionally-trained models, and also in better agreement with humans. To lower the cost of collecting human annotations, we also explore using deep learning to provide automated annotations. CYBORG training of CNNs addresses important issues such as reducing the appetite for large training sets, increasing interpretability, and reducing fragility by generalizing better to new types of data.
\end{abstract}


\section{Introduction}\label{sec:introduction}

\begin{figure*}[!ht]
    \centering
    \includegraphics[width=0.9\textwidth]{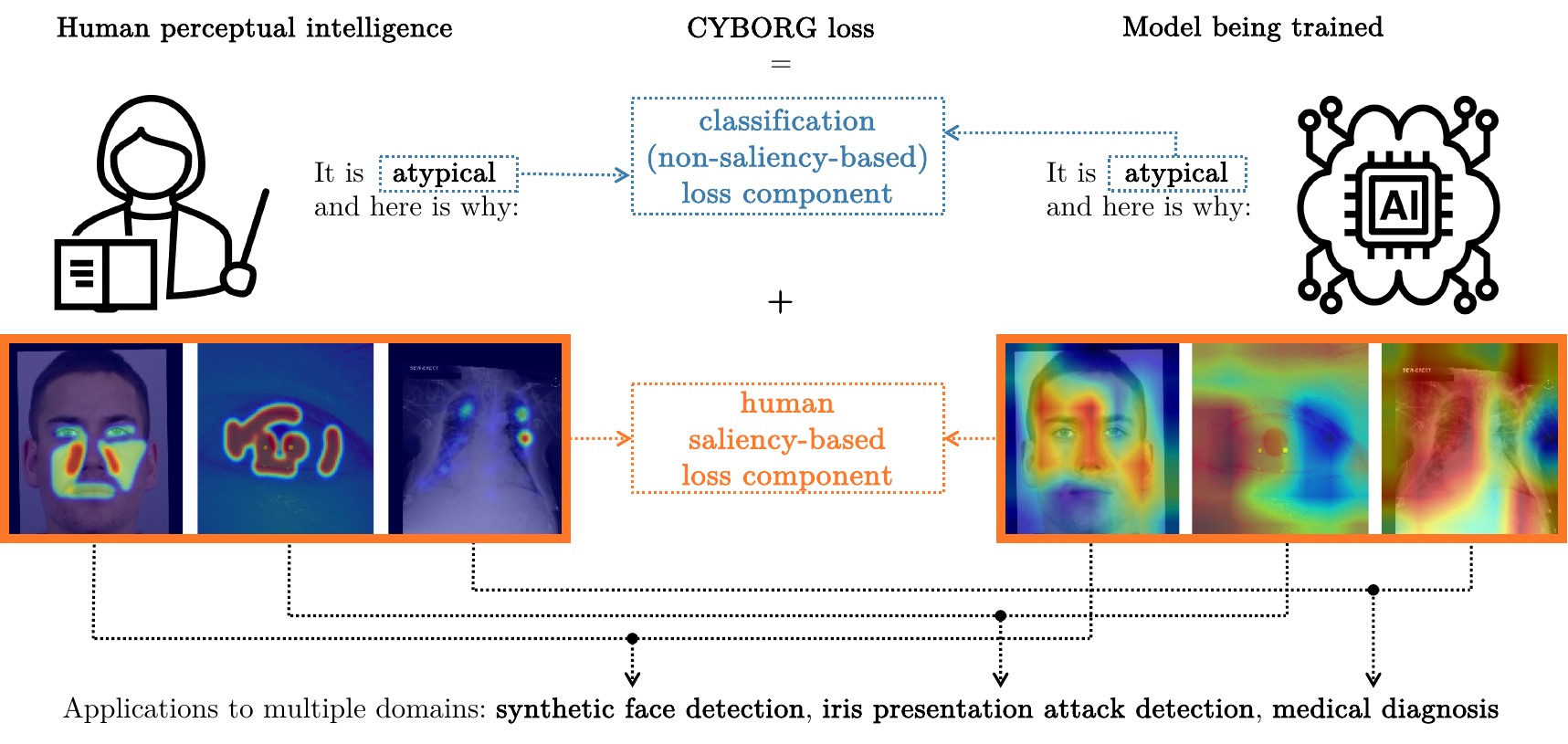}
    \caption{Our proposed training strategy to {\bf C}onve{\bf Y}s {\bf B}rain {\bf O}versight to {\bf R}aise {\bf G}eneralization. CYBORG  guides the network throughout training to learn features  
    using image regions judged as salient for human visual perception. This results in a model that is more likely to learn features from regions that are salient to humans, and less likely to learn features that are accidentally correlated with class labels. A boost in generalization performance is demonstrated.}
    \label{fig:teaser}
\end{figure*}

\IEEEPARstart{T}{he} 
quest for deep learning models that better generalize to new data calls for the ability to incorporate domain-specific expertise into model training, in addition to simply maximizing accuracy on training data.  Human perception is one of the most attractive sources of this domain-specific expertise that can improve model performance when compared to purely data-driven techniques. This is especially important in areas where data acquisition is too time-consuming, expensive, difficult, or sometimes even impossible. For instance, collecting new data for medical image analysis can be problematic due to privacy concerns that weigh even larger than concerns of time and cost. In biometric presentation attack detection, the attack landscape is constantly changing as new attacks are developed, and so collecting a comprehensive dataset representing all current and future attacks is infeasible. Without domain knowledge, data-driven few-shot learning methods have exhibited the tendency to latch onto features that are only coincidentally correlated with class categories.   

Fortunately, human saliency can be used to avoid learning accidental correlations (also known as \textit{spurious features} or \textit{dataset biases}) that reduce a model's ability to generalize
\cite{hovy2015tagging, pmlr-v139-zhou21g,buolamwini2018gender}.  Strategies that incorporate human perception into deep learning models are emerging, especially by guiding models ``where to look'' during training. It has been demonstrated that incorporating human salience into either training data \cite{boyd2022humanblur} or the loss functions \cite{boyd2021cyborg,thomasharmonizing} can guide models toward features that humans use when solving visual tasks. Saliency-based guidance produces models that achieve greater accuracy on unknown presentation attack (PA) types (in iris PAD) and on unknown methods for face image generation (in synthetic face detection). 
We build upon these earlier approaches to formulate a series of research questions, which we answer in this work with a cross-domain solution of incorporating human perceptual intelligence into open-set detection models:

\begin{itemize}
    \item{\bf RQ1} Does human salience-based guidance during training improve the generalization capabilities of the model?
    \item{\bf RQ2} How does human guidance influence models in terms of their salience on the test set and robustness against overfitting?
    \item{\bf RQ3} Is this approach domain-specific, or can it be successfully applied across various domains in which humans can offer visual perception-related expertise?
    \item{\bf RQ4} Can (and if so, how can) human salience be replaced by algorithm-generated salience, 
    or by increasing training dataset size in purely data driven approaches?
    \item{\bf RQ5} Which method of incorporating human salience is more effective: 
    
    \hspace{\parindent}   (a) through training data augmentations (\eg \cite{boyd2022humanblur}), or 
    
    \hspace{\parindent} (b) through components of the loss function penalizing for divergence of the model's salience from human's salience \cite{boyd2021cyborg}?
\end{itemize}

To answer the above questions, we carried out experiments across three domains in which we had access to human saliency data associated with the classification process: (i) synthetic face detection, (ii) iris presentation attack detection, and (iii) abnormality detection in chest X-rays. Fig. \ref{fig:teaser} outlines the proposed loss function, which {\bf c}onve{\bf y}s {\bf b}rain {\bf o}versight to {\bf r}aise {\bf g}eneralization (CYBORG) by comparing human saliency and model saliency, and penalizing large differences between the two. In the first two aforementioned domains, we asked non-experts to classify images of faces and irises as bona fide (real) or synthetic (fake). We simultaneously asked the participants to annotate regions that support their decisions. In the medical imaging domain, we used eye-tracking-sourced salience obtained from doctors as they evaluating X-ray scans to identify abnormalities. We show that:

\begin{itemize}
    \item Human saliency-based guidance during training improves models' generalization capabilities. Improvement is seen for all three domains: synthetic face detection, iris spoofing detection and abnormality detection in X-rays (re: {\bf RQ1}).
    \item Human-guided models show saliency on the test set that (a) more closely resembles human salience, and (b) is stable across training runs, demonstrating that the models are less prone to learn features accidentally correlated with class labels (re: {\bf RQ2}).
    \item The proposed approach can be applied to various domains in which humans can deliver features supporting their decisions through supplied annotations and eye-tracking recording (re: {\bf RQ3}).
    \item Human salience cannot be effectively replaced by simply generating more training samples in the case of synthetic face detection, and the amount of new data needed to match the performance of human-guided models is from 2.4$\times$ for iris presentation attack detection to 6.1$\times$ for the chest x-ray case. This demonstrates (a) effectiveness of CYBORG training in the case of limited data, and (b) the high value of human salience information (re: {\bf RQ4}). 
    \item Incorporating human salience into the loss function is a better approach than human-sourced training data augmentations (re: {\bf RQ5}).
\end{itemize}

\section{Related Work}
\label{sec:rel_work}

\subsection{Synthetic Face Detection}

Since Goodfellow \etal  introduced generative adversarial networks (GAN)~\cite{goodfellow2014generative}, many open-source, pre-trained, GAN-based  generators have been made available~\cite{karras2017progressive, karras2019style, karras2020analyzing, Karras2020ada, Karras2021, choi2020stargan, brock2018large, zhu2017unpaired, park2019gaugan}. Of the  possible types of images to synthesize, fake \textit{face} images have been very popular for both entertainment and research ~\cite{thispersondoesnotexist}. However, as these image generators have grown in popularity, there too grows a demand for fake image detector models for the sake of societal security, trust, and transparency. 

The authors of~\cite{qian2020thinking, frank2020leveraging} state that the frequency domain of images can reveal artifacts in GAN-generated images, regardless of generative model architecture, training dataset, and image resolution. However, as documented by Marra \etal~\cite{marra2018detection}, conventional, non-deep-learning methods (such as frequency analysis and steganalysis~\cite{cozzolino2014image}) show poor generalizability in the context of compressed images. Since there exists (virtually) no limit on the number of fake images to be seen in the training process, deep networks 
have achieved over 99\% accuracy in fake image detection~\cite{tariq2019gan}. 
As described above, the public release of the StyleGAN3~\cite{Karras2021} image generator was accompanied by the release of proactive detector models geared towards detecting StyleGAN3-generated images~\cite{polimi-ispl, wang2019cnngenerated}. 

Although the generation of never-before-seen images lends itself naturally to the creative process, the ability to generate new images and manipulate existing images poses a significant security problem~\cite{botha2020fake,chesney2019deepfakes}. 

\subsection{Iris Presentation Attack Detection (PAD)}

Iris PAD refers to the task of classifying whether or not an object (presented to a biometric sensor) is attempting to drive the system into an incorrect decision \cite{czajka2018presentation,Boyd_PRL_2020}. 
Given the prevalence  
of biometric systems at a national scale (such as in national identification \cite{mir2020realizing} and border control),  development of generalizable PAD models is crucial. 

Creation of models that 
generalize well against truly {\it unknown} attack types is an open research problem and an important aspect of deployable solutions \cite{Boyd_PRL_2020,boyd_pad_assessment_tifs}. Many modern iris PAD approaches rely on deep-learning to achieve state-of-the-art accuracy, as seen in submissions to the LivDet-Iris 2020 and 2023 competitions \cite{Das_IJCB_2020,tinsley2023iris}. In particular, Sharma and Ross \cite{Sharma_IJCB_2020} propose applying DenseNet-121  \cite{Huang_2017_CVPR} to iris PAD with a focus on human interpretability. More recently, Sharma and Chen developed a novel method of attention-guided training that uses class activation mappings and attention modules to further increase generalizability and interpretability \cite{Chen_WACVW_2021}. 
Rather than augment the network with attention modules, our CYBORG approach encourages the model to learn salient image features through a modified loss function. A natural benefit of the CYBORG approach is the simplicity associated with keeping the original network intact while only modifying the loss.
Furthermore, since CYBORG loss explicitly penalizes the model for straying from human-annotated regions of interest, networks trained with CYBORG show increased interpretability for humans.  This can be seen in Fig. \ref{fig:visualizations}(b), showing that CYBORG  encourages the network to focus on salient regions (the iris) as opposed to peripheral image features.

\subsection{Abnormality Detection in Chest X-Ray Images}

In the context of medical imaging, there exists a significant data scarcity due to (i) the inherently personal nature of the acquired data, and (ii) the time and cost required to collect said data. The COVID-19 pandemic has led to an increase in effort for timely anomaly detection ~\cite{zhang2020covid, ibrahim2021deep}, but most machine learning pipelines (especially for anomaly detection) typically ingest and learn from much larger datasets.

In order to remedy this data scarcity, there have been attempts to augment the limited raw image data with more informative auxiliary data. One such form of data is free-text labels that radiologists write down (or dictate) to describe the reasoning behind their diagnosis. Another form of data (also collected at time of diagnosis) is eye-tracking data that more implicitly highlights areas of importance as judged by the medical practitioners. The combination of raw chest x-ray (CXR) imaging, free-text labels, and eye-tracking data has led to impressive results in robust lung cancer detection ~\cite{chiu2022application}. In \cite{boecking2022making}, Boecking \etal focus primarily on text-based models to glean semantic value from free-text labels to improve their joint vision-language models, which are also the basis for work in \cite{iyer2022self, van2023probabilistic, sato2022anatomy}.

\subsection{Using Human Perception to Understand and Improve Computer Vision}

In \cite{o2012comparing}, O'Toole \etal show that current face recognition algorithms outperform humans, except  in challenging cases. RichardWebster \etal~\cite{richardwebster2018visual} demonstrated that observing the behaviour of humans completing a face recognition task can be used to explain  face recognition algorithms' decisions, allowing for increased model explainability. A recent paper by Fel \etal~\cite{thomasharmonizing} details a trade-off between neural network classification accuracy and alignment with human visual strategies for object recognition. They propose a general purpose training procedure that aligns neural network and human visual strategies while improving accuracy.

In the biometrics domain, it was found that human saliency and machine saliency provide complementary information, proving beneficial when combined~\cite{trokielewicz2019perception,moreira2019performance}. Human saliency assessed from eye tracking was collected by Czajka \etal~\cite{czajka2019domain} and used to derive filter kernels for iris recognition. This method outperformed non-human-driven approaches and was shown \cite{boyd2020post} to be the current state-of-the-art in post-mortem iris recognition. Boyd \etal~\cite{boyd2023pbm} collected human annotations on matching and non-matching features for post-mortem iris recognition, and showed how training models on the human saliency data led to a fully interpretable matching tool. Human saliency was later used in the iris PAD domain to augment the training data to emphasize regions defined by this saliency~\cite{boyd2022humanblur}. This approach resulted in methods generalizing exceptionally well to unknown attack types. Shen \etal~\cite{shen2021study} show that humans classify synthetically generated faces at no better than random chance. Boyd \etal~\cite{boyd2022value} then showed how supplying saliency information from deep learning models can boost human performance in the same task. Boyd \etal~\cite{boyd2021cyborg} incorporate human saliency in the form of explicit annotations into the loss function and demonstrate a significant improvement on open-set synthetic face detection. \textbf{Our work presented in this paper builds upon this preliminary efforts of Boyd \etal~\cite{boyd2021cyborg} to demonstrate the utility of human-guided training across various computer vision domains.}

More generally in machine learning, incorporation of psychophysics has aided in deep learning tasks such as image captioning for scene understanding~\cite{he2019human,huang2021specific}, handwriting analysis~\cite{grieggs2021measuring}, and natural language processing~\cite{zhang2020human}. Linsley \etal~\cite{linsley2018learning} proposed to incorporate human-sourced saliency into a self-attention mechanism, combining global and local attention in the ``GALA'' module. 
Bruckert \etal~\cite{bruckert2021deep} investigate  popular loss functions used in deep saliency models, showing the significance of loss function selection. It was found that linear combinations of several loss functions led to performance increases across datasets and architectures. \textbf{We build on this finding in this paper when exploring the optimal weighting of the loss components.}

\subsection{Salient Object Detection}

The goal of salient object detection (SOD) is to highlight regions of images humans deem salient ~\cite{borji2015salient, wang2021salient}. Although related, CYBORG and SOD differ in regards to the use of ground truth data. While SOD attempts to predict ground truth heatmaps, CYBORG uses \textit{subjective} heatmaps during training to guide the model towards salient image regions.

\section{Blending Human Perceptual Intelligence into Training: CYBORG Loss}
\label{sec:cyborg_loss}

The CYBORG loss function combines a \textit{human saliency loss component} with the traditional cross-entropy \textit{classification loss component}.  The human saliency loss component is created by comparing the human saliency map for an image to the model’s current class activation map for the image.
The relative weighting of the human saliency loss and the classification loss is explored thoroughly in Sec. \ref{sec:CYBORG_params}.  The human saliency loss component emphasizes ``where to look'' and the classification loss component emphasizes maximizing accuracy.
The intuition is that the human saliency loss guides the learning away from image features that are only accidentally correlated with class categories, and thereby improves the model’s generalization. 
In effect, CYBORG 
guides the model away from learning features that are ``right for the wrong reason'' \cite{pmlr-v139-zhou21g}.

The human saliency loss component steers activations in the feature maps from the last convolutional layer to align with human-derived saliency heatmaps by comparing them with model's salience. To accomplish this, a fully-differentiable version of the Class Activation Mapping (CAM) approach~\cite{zhou2016learning} is implemented,  enabling the generation of CAMs for all samples in each training batch. Formally, the CYBORG loss $\mathcal{L}_\text{CYBORG}$ is defined as:

\vskip-3mm
\begin{equation}
\begin{split}
\mathcal{L}_\text{CYBORG} = \frac{1}{K}\sum_{k=1}^K\sum_{c=1}^{C}\bm{1}_{y_k \in \mathcal{C}_c} \\
\Bigg[\underbrace{(1-\alpha)\mathcal{L}_s\big(\textbf{s}_k^{\text{(human)}},\textbf{s}_k^{\text{(model)}}\big)}_{\text{human salience loss component}} -\underbrace{\alpha\log p_{\text{model}}\big(y_k \in \mathcal{C}_c\big)}_{\text{classification loss component}}\Bigg]
\end{split}
\label{eqn:cyborg}
\end{equation}

\noindent
where $\mathcal{L}_s$ is a measure comparing model and human salience maps, $y_k$ is a class label for the $k$-th sample, $\bm{1}$ is a class indicator function equal to $1$ when $y_k \in \mathcal{C}_c$ (0 otherwise), $C$ is the total number of classes, $K$ is the number of samples in a batch, $\alpha$ is a trade-off parameter weighting human- and model-based saliencies, $\textbf{s}_k^{\text{(human)}}$ is the human saliency for the $k$-th sample, and 
\vspace{-0.45em}
$$
\textbf{s}_k^{(\text{model})} = \textbf{f}_1w_1^{(c)} + \textbf{f}_2w_2^{(c)} + \dots + \textbf{f}_Nw_N^{(c)}
$$
\noindent
is a class activation map-based model's saliency for the $k$-th sample, where $N$ is the number of feature maps $\textbf{f}$ in the last convolutional layer, and $w^{(c)}$ are the weights in the last classification layer belonging to predicted class $\mathcal{C}_c$. Both $\textbf{s}_k^{(\text{model})}$ and $\textbf{s}_k^{(\text{human})}$ are normalized to the range $\langle 0,1\rangle$, and additionally the human salience maps are downsized to the same size as the CAMs. This paper explores using $L_1$ norm, $L_2$ norm, Structural Similarity (SSIM) index, and combination of those, as measures in the salience loss  $\mathcal{L}_s$.\footnote{As mentioned in \cite{boyd2021cyborg}, the source code for CYBORG can be found here: \url{https://github.com/CVRL/CYBORG}}

\section{Experimental Setup}
\label{sec:experiments}

\begin{table}[!tb]
\centering
\caption{Details on the discovered final parameter sets based on the search conducted on the loss function and $\alpha$ parameter. Colors are matched with Fig. \ref{fig:cyb_settings}.}
\begin{tabular}{|c|c|c|c|}
\hline
 & \textbf{Setting Name} & \textbf{Human Sal. Loss} & \textbf{$\alpha$ Value} \\ \hline\hline
\cellcolor[HTML]{ECF9FF} CYBORG$_{gen}$ & \cellcolor[HTML]{ECF9FF}$S$        & SSIM                & 0.75            \\ \hline\hline
 
 \multirow{3}{*}{\cellcolor[HTML]{FFFBEB}} \multirow{3}{*}{CYBORG$_{arch}$} & \cellcolor[HTML]{FFFBEB}$S_d$         & SSIM+MSE                & 0.8            \\
\multirow{3}{*}{\cellcolor[HTML]{FFFBEB}}& \cellcolor[HTML]{FFFBEB}$S_r$         & L1                & 0.65            \\
\multirow{-3}{*}{\cellcolor[HTML]{FFFBEB}CYBORG$_{arch}$}& \cellcolor[HTML]{FFFBEB}$S_n$         & SSIM+L1                & 0.85            \\ \hline\hline
\multirow{9}{*}{\cellcolor[HTML]{FFE7CC}}\multirow{9}{*}{CYBORG$_{opt}$} & \cellcolor[HTML]{FFE7CC}$S_{d/f} $        & L1                & 0.25            \\
\multirow{9}{*}{\cellcolor[HTML]{FFE7CC}}& \cellcolor[HTML]{FFE7CC}$S_{d/i} $        & L1                & 0.55            \\
\multirow{9}{*}{\cellcolor[HTML]{FFE7CC}}& \cellcolor[HTML]{FFE7CC}$S_{d/c} $        & SSIM                & 0.7            \\
\multirow{9}{*}{\cellcolor[HTML]{FFE7CC}}& \cellcolor[HTML]{FFE7CC}$S_{r/f} $        & L1                & 0.35            \\
\multirow{9}{*}{\cellcolor[HTML]{FFE7CC}}& \cellcolor[HTML]{FFE7CC}$S_{r/i} $        & SSIM+L1                & 0.85            \\
\multirow{9}{*}{\cellcolor[HTML]{FFE7CC}}& \cellcolor[HTML]{FFE7CC}$S_{r/c} $        & SSIM+L1                & 0.75            \\
\multirow{9}{*}{\cellcolor[HTML]{FFE7CC}}& \cellcolor[HTML]{FFE7CC}$S_{n/f}  $       & L1                & 0.45            \\
\multirow{9}{*}{\cellcolor[HTML]{FFE7CC}}& \cellcolor[HTML]{FFE7CC}$S_{n/i}  $       & SSIM+L1                & 0.75            \\
\multirow{-9}{*}{\cellcolor[HTML]{FFE7CC}CYBORG$_{opt}$}& \cellcolor[HTML]{FFE7CC}$S_{n/c} $        & SSIM+L1                & 0.85           \\\hline

\end{tabular}
\label{tab:cyborg_settings}
\end{table}

\begin{figure}
    \centering
    \includegraphics[width=0.85\columnwidth]{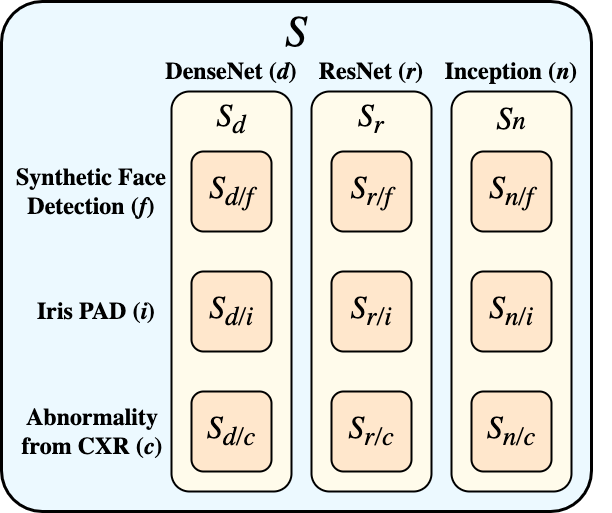}
    \vskip-1mm
    \caption{Explanation of parameter sets used in this work. }
    \label{fig:cyb_settings}
    \null\vskip-5mm
\end{figure}

\subsection{General Setup}

For all experimental runs in this work, model training parameters and procedures are kept constant. To ensure that observations are not architecture-specific, the base experiments are completed on three out-of-the-box architectures: DenseNet-121, ResNet50 and Inception v3.

The experimental setup for this work enables four specific improvements over previous CYBORG work\cite{boyd2021cyborg}.
\begin{enumerate}
    \item previous work studied only one domain, synthetic face detection, whereas this work studies three different domains in order to establish the generality of the CYBORG approach;
    \item previous work fixed the balance between the human saliency and  classification,  $\alpha$  in Eqn. \ref{eqn:cyborg}, at 0.5, whereas this work explores optimizing this parameter to achieve better performance;
    \item previous work uses mean squared error as the penalty for the human saliency component, without exploring other possibilities, whereas this work evaluates multiple alternatives;
    \item previous work uses only one type of human saliency data (annotations), whereas this work also uses saliency derived from eye-tracking data.
\end{enumerate}

\textbf{To address point 1}, three different domains are studied: 1) synthetically generated face detection \cite{marra2018detection,wang2019cnngenerated}, 2) iris presentation attack detection \cite{Boyd_PRL_2020,czajka2018presentation} and 3) abnormality detection from chest x-rays \cite{ccalli2021deep}. The goal is to outline the broad applicability of our CYBORG approach. 

\textbf{To address point 2}, the $\alpha$ values ranging from 0.05 to 1.0 in increments of 0.05 are used to determine the optimal value based on validation Area Under the ROC Curve (AUC) for all domains and network backbones. 

\textbf{To address point 3}, three loss penalties are employed in the human saliency component to determine the optimal based on the validation AUC. Loss penalties studied are mean squared error (MSE), mean absolute error (L1) and structural similarity index measure (SSIM). Additionally, inspired by \cite{bruckert2021deep}, pairs of these three losses are linearly combined to attain SSIM+MSE and SSIM+L1. Both L1 and MSE penalize the pixel-wise distance between the human saliency and the model saliency whereas SSIM measures the overall similarity \cite{brunet2011mathematical} between the human saliency and the model saliency. Thus, the combination of L1 or MSE with SSIM provides potentially complementary information. 

\textbf{Finally, to address point 4}, human saliency information  from eye tracking data is introduced in addition to annotation data to determine whether the proposed CYBORG loss can be used with various forms of explicit human saliency.

For all experiments, the Stochastic Gradient Descent (SGD) optimizer is used, with learning rate of $0.005$, modified by a factor of $0.1$ every 12 epochs. Training ran for maximum 50 epochs with a batch size of 20. The epoch with the highest validation accuracy was selected as the final model. These parameters are consistent with those proposed in \cite{Sharma_IJCB_2020,boyd2022humanblur,boyd2021cyborg}.
Within each individual domain, the validation set is constant for all experiments. 
All networks are initialized with pre-trained ImageNet weights~\cite{pytorchModelZoo}. Each model is independently trained 10 times, to generate error statistics on the test set.

\subsection{Effect of including human saliency in the loss}
\label{sec:scenarios}

To evaluate the effect of the human saliency loss component of CYBORG loss, models are trained in two scenarios: 1) with no human saliency information involved in the training  and 2) with human saliency information.
The first scenario represents the traditional approach to training deep CNN models.
Models are trained using a loss function that optimizes the classification accuracy, with the hopt that the resulting model can generalize well to unseen test data. 
Categorical cross-entropy is employed as the loss penalty for classification performance. Models trained in this scenario will be referred to as \textit{traditionally-trained} models.

The second scenario differs from the traditional scenario only in adding a human saliency component to the classification component, to create the CYBORG loss, as described in Sec. \ref{sec:cyborg_loss}. 
Because the addition of the human saliency component is the only difference between the two scenarios, performance differences can be directly attributed to the CYBORG approach.
Models trained in this scenario will be referred to as \textit{CYBORG trained} models.

\subsection{Selecting Optimal CYBORG Parameters}
\label{sec:CYBORG_params}

Two improvements over earlier CYBORG work that are introduced in this paper are (a) determining the optimal loss function for the human saliency component of CYBORG loss, and (b) determining the right balance between the human saliency and the classical loss components. In \cite{boyd2021cyborg}, the human saliency loss was arbitrarily selected as mean-squared-error loss and the balance of classification loss to human saliency loss was arbitrarily set as having the same importance ($\alpha$ = $0.5$). 

To determine a better solution for these two questions, a thorough parameter search is completed. For each of the  DenseNet, ResNet and Inception architectures, models are trained with $\alpha$ ranging from 0.05 to 1.0 in increments of 0.05, with a value of 1.0 resulting in using no human saliency in the training, \ie traditionally trained models. 

As explained previously, this parameter search for $\alpha$ is completed for five loss functions: mean squared error (MSE) as in \cite{boyd2021cyborg}, L1 loss, structural similarity loss (SSIM) and, taking inspiration from \cite{bruckert2021deep}, the combinations 
SSIM+L1 and SSIM+MSE losses. 
To determine the optimal combination of $\alpha$ and loss function for a given architecture and domain, the highest average AUC on the validation set across the 10 trained models is selected. 

The described approach of identifying the optimal combination of $\alpha$ and loss function will be henceforth referred to as CYBORG$_{opt}$. This is the most specialized approach as it is optimized to both the network architecture and the domain. In this work, as there are three studied architectures and three domains, there are nine individual CYBORG$_{opt}$ combinations, as seen in Fig. \ref{fig:cyb_settings} and Tab. \ref{tab:cyborg_settings}.

\subsection{Architecture-Specific CYBORG Parameters}
\label{sec:params-architecture-specific} 

The parameter combination defined by CYBORG$_{opt}$ is optimized for both the architecture  and the domain. However, if future researchers wish to use CYBORG on some different domain,  
a new set of recommended parameters is proposed. 
These are parameters specific to DenseNet, ResNet and Inception, but {\bf not} specific to a domain.
These will be referred to as CYBORG$_{arch}$, as seen in
Fig. \ref{fig:cyb_settings}.

A ranking system is used to determine the CYBORG$_{arch}$ parameter settings.
Across each domain, $\alpha$/loss combinations are ranked from best to worst based on  average validation AUC for the 10 trained models. The best combination is assigned a point value of 1, increasing by 1 for each subsequently well-performing combination. For each architecture, these point values are  summed across the three domains. The combination with the lowest overall point value 
performed most consistently over all three domains and is selected as the CYBORG$_{arch}$ set.

\subsection{Selecting Generic CYBORG Parameters}

A general parameter combination is also proposed. This is what the authors recommend future researchers employ if their domain and network architecture falls outside of those studied in this work. This parameter combination represents the consistently best-performing combination across architectures and domains. The one parameter combination here, represented by $S$ in Fig. \ref{fig:cyb_settings}, is denoted as CYBORG$_{gen}$. To calculate CYBORG$_{gen}$ parameters, the point values used for CYBORG$_{arch}$ are summed across the three architectures, and the ranking is repeated, as described in Sec. \ref{sec:params-architecture-specific}.

\subsection{Assessing The Value Of Human Annotations} 
\label{sec:face_segmentation}

In this work, human saliency maps have been utilized exclusively in the human saliency component of CYBORG loss.
However, what happens if we do not have human saliency data? 
This experiment answers whether deep learning-based segmentation masks can be substituted in place of human saliency maps, and still increase performance over classification loss alone.
This experiment scenario will be referred to as CYBORG-DL. For fairness, the same parameter search as for CYBORG$_{opt}$ is performed with the deep learning-based segmentation masks instead of human saliency maps. 

For face images, BiSeNet \cite{bisenet2019} is used to obtain a mask detailing facial regions excluding the hair and neck. For iris segmentation, a SegNet-based method \cite{Trokielewicz_IVC_2020} is used to extract the entire iris region excluding the pupil and occlusions from the eyelid and eyelashes. For chest region extraction from chest x-ray, a U-Net-based segmentation is employed to segment the lungs \cite{lanfredi2021comparing}. Using the segmented lungs, the convex hull was calculated to include the mediastinum and the bilateral hemidiaphragms. 
The average segmentation map across all training images for each of the three domains can be seen in Fig. \ref{fig:average_maps}. When compared to the average human saliency map across the same images, it is clear the human saliency is on average looking at more specific features than the deep learning-based segmentation output.

\subsection{How Much Training Data Does Traditional Training Need to Match CYBORG Performance?}
\label{sec:iterative}

One way to assess the importance of the increased accuracy achieved by CYBORG is to ask how much more data traditional training would need to achieve the same accuracy.
To investigate this, models are trained using only classification loss on increasing numbers of samples, in multiples of the size of the original training set. The crossover point between the AUC attained using CYBORG and the AUC for traditional training with increasing training set sizes tells us how much more powerful CYBORG learning is.

For fairness, as the training set increases in size, the proportions of the classes are kept constant. Also, the same validation set is used for all experiments in a given domain. 
For synthetic face detection, it was not possible to go past $10\times$ the original dataset size, as the real faces in the original sources were depleted and so maintaining the same proportions as the original dataset became impossible. For iris and chest X-ray samples, the associated datasets without human salience allowed a larger version of this experiment.

\subsection{Domains}
\label{sec:domains}

The three domains selected in this paper represent cases where data is inherently limited. This may be due to the lack of unknown attack types in the test set (synthetic face detection and iris presentation attack detection), or the cost associated with acquiring data (abnormality detection from chest x-rays).

\begin{table}[!tb]
\caption{Number of samples in the train, validation and test sets across the three studied domains, with the numbers of typical/atypical samples within each set.}
\centering
\begin{tabular}{c|ccc}
        & \multicolumn{3}{c}{\begin{tabular}[c]{@{}c@{}}Number of Samples \\ (typical/atypical)\end{tabular}} \\ \hline
Domains & \multicolumn{1}{c|}{Train}            & \multicolumn{1}{c|}{Validation}            & Test            \\ \hline\hline
\begin{tabular}[c]{@{}c@{}}Synthetic\\ Face Detection\end{tabular} &
  \multicolumn{1}{c|}{\begin{tabular}[c]{@{}c@{}}1,821\\ (919/902)\end{tabular}} &
  \multicolumn{1}{c|}{\begin{tabular}[c]{@{}c@{}}20,000\\ (10k/10k)\end{tabular}} &
  \begin{tabular}[c]{@{}c@{}}700,000\\ (100k/600k)\end{tabular} \\ \hline\hline
Iris PAD &
  \multicolumn{1}{c|}{\begin{tabular}[c]{@{}c@{}}765\\ (198/567)\end{tabular}} &
  \multicolumn{1}{c|}{\begin{tabular}[c]{@{}c@{}}23,312\\ (11,656/11,656)\end{tabular}} &
  \begin{tabular}[c]{@{}c@{}}12,432\\ (5,331/7,101)\end{tabular} \\ \hline\hline
\begin{tabular}[c]{@{}c@{}}Abnormality \\ from CXR\end{tabular} &
  \multicolumn{1}{c|}{\begin{tabular}[c]{@{}c@{}}1,988\\ (648/1,340)\end{tabular}} &
  \multicolumn{1}{c|}{\begin{tabular}[c]{@{}c@{}}1,508\\ (486/1,022)\end{tabular}} &
  \begin{tabular}[c]{@{}c@{}}3,802\\ (675/3,127)\end{tabular} \\ 
\end{tabular}
\label{tab:num_samples}
\end{table}

\subsubsection{\textbf{Synthetic Face Detection}}

\begin{figure}
    \centering
    \includegraphics[width=\columnwidth]{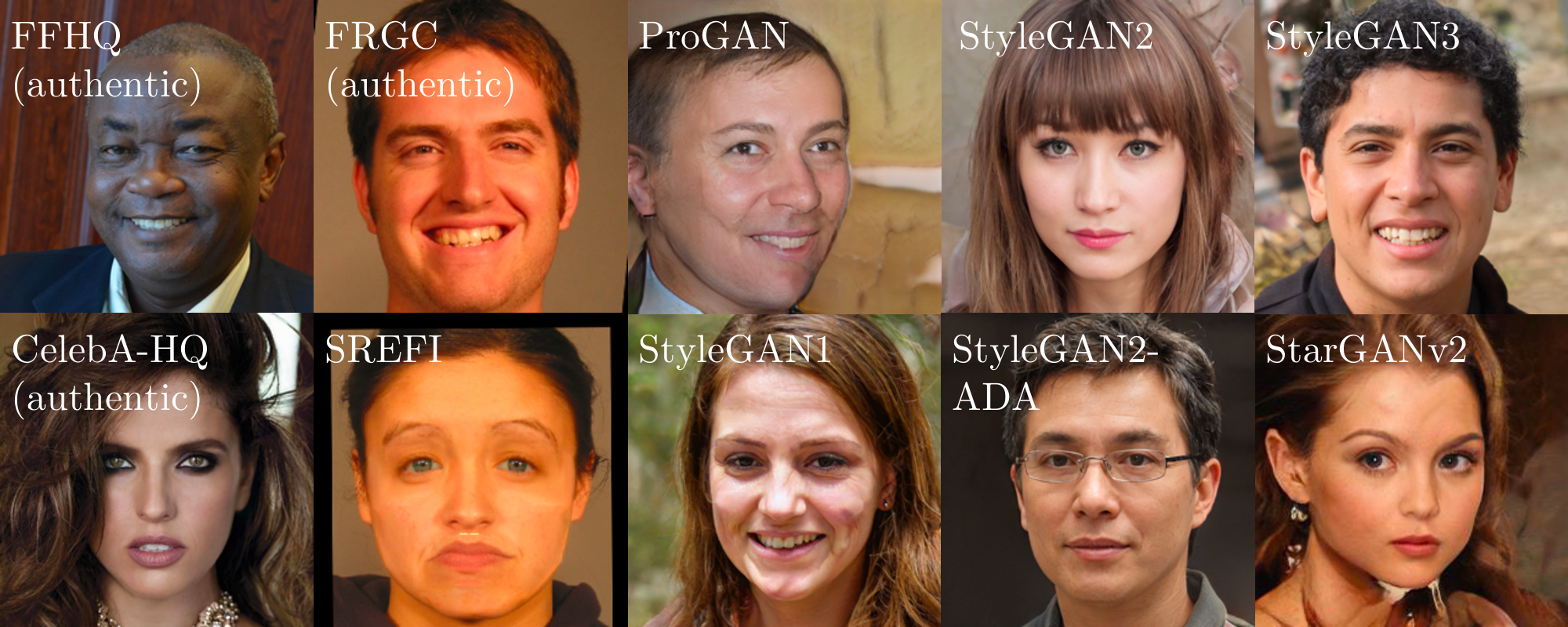}
    \vskip-1mm
    \caption{Example images from each data source for the task of synthetic face detection.}
    \label{fig:test_data_face}
\end{figure}

The task is to classify a face image as representing a real person or a synthetic (potentially non-existent) person.
Images of real persons are drawn from three datasets: CelebA-HQ~\cite{karras2017progressive}, Flickr-Faces-HQ (FFHQ)~\cite{karras2019style} and FRGC-Subset~\cite{Phillips_IVC_2017}.
Synthetic images of non-existent persons are drawn from seven generators (SREFI, ProGAN, StyleGAN, StyleGAN2, StyleGAN2-ADA, StyleGAN3, StarGANv2)~\cite{Banerjee_IJCB_2017,karras2017progressive,karras2019style,karras2020analyzing,Karras2020ada,Karras2021,choi2020stargan}).
The datasets are described in more detail below, and example images shown in Fig. \ref{fig:test_data_face}.

\vskip2mm
\noindent \textbf{Motivation for Selected Domain}

\vskip1mm\noindent{While} it is true that in this domain one could theoretically generate an infinite number of samples, because this dataset represents an open-set style evaluation (different image synthesizers in train and test), the generation of extra data from the same generator does not bring significant new information to the training process. Without new information, the ability of the proposed model to learn generalized features capable of distinguishing fake faces from newer generators is diminished. CYBORG addresses this by incorporating human defined regions of saliency into the training process. The nature of this domain is such that new synthesis methods are constantly being developed, so countermeasures need to be robust against as many types as possible. The use of this dataset helps to evaluate the ability of CYBORG to learn a more generalized feature representation that enables stronger classification performance on unseen generators in the test set.

\vskip2mm
\noindent \textbf{Image Data}

Authentic images are supplied by CelebA-HQ~\cite{karras2017progressive,liu2015faceattributes} provides 30,000 high-quality celebrity images, while Flickr-Faces-HQ (FFHQ)~\cite{karras2019style} contains 70,000 diverse faces from Flickr. The FRGC-Subset~\cite{Phillips_IVC_2017} includes 16,433 images from the Face Recognition Grand Challenge. For synthetically generated faces, SREFI~\cite{Banerjee_IJCB_2017} generates synthetic faces by blending regions of real images to create new identities. ProGAN~\cite{karras2017repo}, StyleGAN\cite{karras2019style}, and its successors (StyleGAN2, SG2-ADA, and StyleGAN3)~\cite{karras2020analyzing, Karras2020ada, Karras2021} produce 100,000 synthetic images, improving image quality and handling data-limited training. Lastly, StarGANv2~\cite{choi2020stargan} generates 100,000 high-quality mixed-style faces, using reference images for style transfer and filtering based on facial quality metrics.
Further extensive details about each of the image sources demonstrated in Fig. \ref{fig:test_data_face} can be found in the supplemental materials.

\vskip1mm\noindent{\textbf{\textit{Image Preprocessing}}} Face images from all data sources are aligned using {\it img2pose}~\cite{Albiero_CVPR_2021}, cropped, and resized to $224\times224$. Face bounding boxes are expanded 20\% in all directions before cropping, with an additional 30\% on the forehead to ensure the face is central and fully in view. Human saliency maps (described in the next section) are resized and cropped to the same specifications, to keep spatial correspondence. 

\vskip2mm
\noindent{\textbf{Human Saliency Data}}

\noindent 
The saliency data for the face images in the training is the same as used in \cite{boyd2021cyborg}, who replicated experiments similar to those of Shen \etal~\cite{shen2021study}.
In \cite{boyd2021cyborg}, subjects were shown a pair of face images, one real and one synthetic, and asked to judge which is real or synthetic, and also to annotate regions of the image that supported their decision. 
In \cite{shen2021study}, subjects were only asked the classification question, and not asked to annotate regions of the image.

Saliency data, consisting of image classifications and manual image annotations, were collected from 363 subjects recruited via Amazon Mechanical Turk.
On average, 29.6 image pairs were processed by each subject. 
Synthetic images consisted of even splits of (i) 500 images generated by the SREFI method with the FRGC-Subset dataset, and (ii) 500 images synthesized by StyleGAN2 (downloaded from \url{thispersondoesnotexist.com}). In total, 10,750 annotations were obtained.
(This matches the number of image pair samples in \cite{shen2021study}.)
In training our CYBORG models, only annotations for \textbf{correctly} classified pairs are used, so the number of images in Tab. \ref{tab:num_samples} is less than the total number of trials in \cite{boyd2021cyborg} and \cite{shen2021study}. To create a single human saliency heatmap for each image, we averaged all available binary annotations (generated by each annotator) into a heatmap with an intensity normalized to the $[0,1]$ range. (This averaging approach is also applied when generating human saliency heatmaps in two other  domains  in this paper: iris presentation attack detection and chest X-ray anomaly detection). This approach highlights features that were agreed on by most annotators, but also retains features where inter-rater agreement was low. The latter situation does not necessarily mean that the human saliency heatmap is of low significance or quality; it only means that annotators found different features useful in making their decision, which may still be useful for guiding the model's training.

\begin{figure}[!t]
    \centering
    \includegraphics[width=0.8\columnwidth]{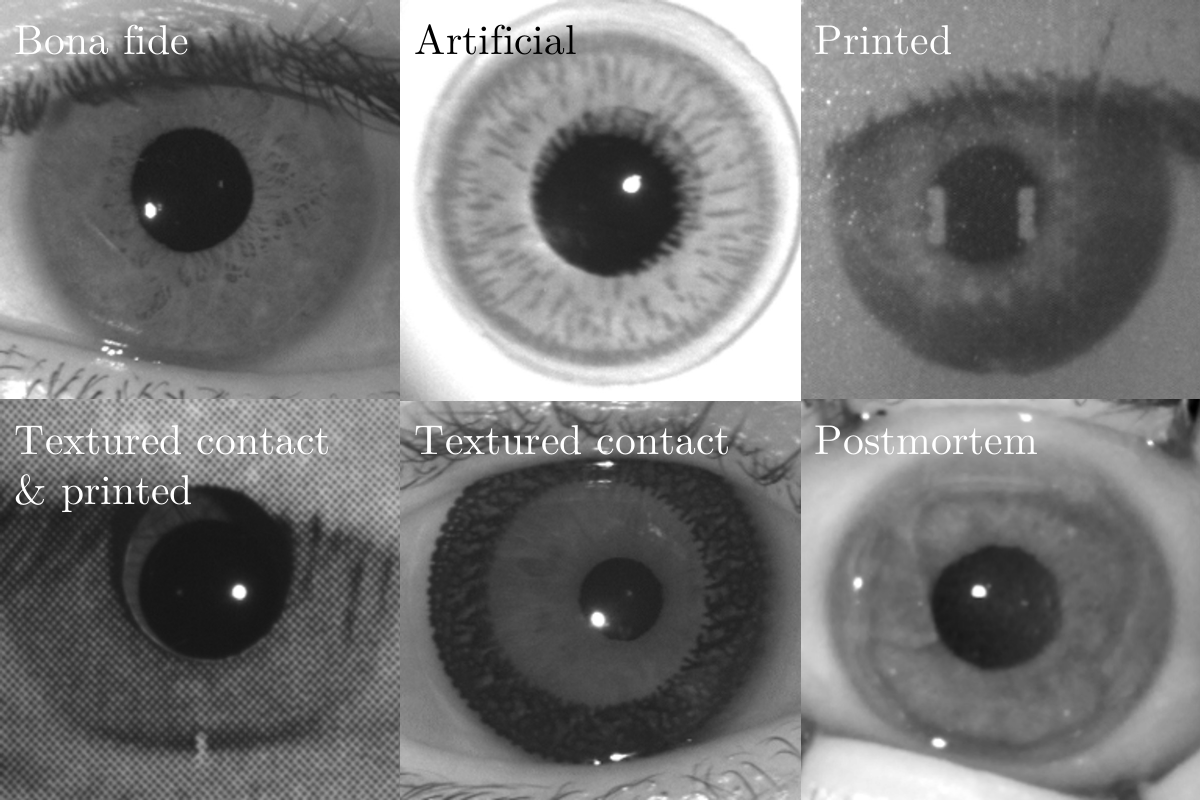}
    \vskip-1mm
    \caption{Example images from each data source for the task of iris presentation attack detection.}
    \label{fig:test_data_iris}
    \null\vskip-5mm
\end{figure}

\subsubsection{\textbf{Iris Presentation Attack Detection (PAD)}}

As discussed in the following section, iris images are classified as bona fide or presentation attack. (There are seven types of images in the attack class for the training and validation splits, and five types for the test split.) 
The training, validation and testing splits are in Tab. \ref{tab:num_samples}; typical refers to bona fide iris images and atypical to presentation attack images.

\vskip2mm\noindent \textbf{Motivation for Selected Domain}

\vskip1mm\noindent{Similarly} to synthetic face detection, the landscape for iris presentation attacks is constantly evolving as newer spoof scenarios are developed. Thus, given a training set of currently known attacks, we need to make sure that models are trained in a way to be robust to both known attacks (those seen during training) and unknown attacks (those not seen during training). Traditional training approaches show strong performance on known attacks but struggle to recognize unknown spoof examples, even when they are obvious to humans \cite{boyd_pad_assessment_tifs}. This domain represents an important open computer vision problem that supports and highlights the value of the CYBORG approach.

\vskip2mm
\noindent \textbf{Image Data}

\vskip1mm\noindent An effort was made an effort to acquire all publicly available iris PAD datasets~\cite{boyd_pad_assessment_tifs}. From the initial set of 800,000 iris images, duplicates and non-ISO-compliant~\cite{ISO_19794_6_2011} images were removed, resulting in 458,790 samples. This dataset was used to create training and validation sets. 
We also curated a sample-disjoint test set, which is identical to the most recent LivDet-Iris competition benchmark \cite{Das_IJCB_2020}. This LivDet-2020 test set contained 12,432 samples from 6 categories (live + 5 PAIs). This test dataset was excluded from all training and validation processes, and was held entirely for  final testing. This set-up allows for direct comparison with the results of the LivDet-Iris 2020 competition; it also allows us to assess the generalization capabilities of the proposed approach.
Example images are shown in Fig. \ref{fig:test_data_iris}, 
The term \textit{atypical} is assigned to the samples that differ from \textit{bona fide (live)} samples \ie presentation attacks.

Every image in the dataset was segmented using a SegNet-based method \cite{badrinarayanan2017segnet}. Images were then cropped and resized to $224\times224$ for input to the network.

\vskip2mm
\noindent{\bf Human Saliency Data}

\vskip1mm
\noindent The human saliency data integrated into CYBORG loss training for the task of iris PAD comes from 
\cite{boyd2022humanblur}. In \cite{boyd2022humanblur},
non-salient regions of iris images were blurred to deter models from learning distracting features, whereas this work highlights \textit{salient} regions to encourage models to learn features humans deem important. The salience data was collected via an internally developed online annotation tool. Participants in the study were presented 8 types of images: \textit{bona fide} and 7 \textit{abnormal} types, as presented in Fig. \ref{fig:test_data_iris}. Participants were not trained in iris PAD or iris recognition tasks, and were recruited from the University of Notre Dame students, staff and faculty at the time of data collection. Full details on the annotation collection process, as described in \cite{boyd2022humanblur}, can be found in the supplementary materials.

Only annotations from correctly classified samples are used in later training. Since PAD is a binary classification problem, decisions were \textit{correct} if the subject (i) correctly classified a bona fide sample as bona fide, or (ii) classified any of the 7 abnormal types as abnormal.

Note that merely collecting more labeled samples (bona-fide/abnormal) may be impossible in the context of biometric attacks since these may be sparsely represented in datasets of ample size. Additionally, increasing the number of labeled samples might not guide the network \textit{where} to look, opposite to the idea proposed throughout this work, \ie the network, by simply observing more data, would still need to figure out relevant features from irrelevant without further guidance provided by the loss function.

\vspace{1.5em}
\subsubsection{\textbf{Abnormality Detection from CXR}}
\begin{figure}[t]
    \centering
    \includegraphics[width=\columnwidth]{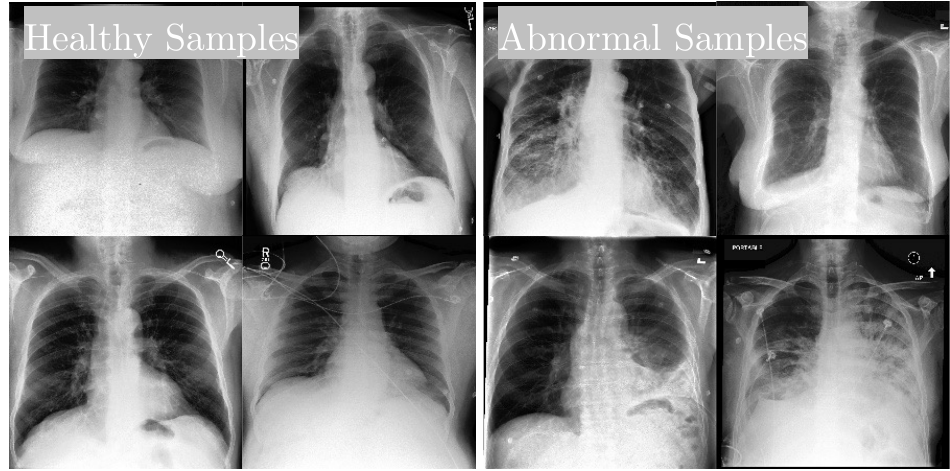}
    \vskip-1mm
    \caption{Examples of both healthy and abnormal chest x-rays for the task of abnormality detection from chest x-ray.}
    \label{fig:test_data_cxr}
    \null\vskip-5mm
\end{figure}

\begin{table*}
\centering
\caption{Overall \textbf{Area Under ROC Curve (AUC)} results for all experimentation. In all cases, CYBORG outperforms traditionally trained models. The N/A columns refer to cases when the experiment was not possible to perform. The $*$ refers to when the same configuration appears as best in two scenarios.}
\begin{tabular}{|c|c||c||c|c||c|c|c|}
\hline
  Application & Network & Traditional & CYBORG-DL & CYBORG-DL-Fine & CYBORG$_{gen}$ & CYBORG$_{arch}$ & CYBORG$_{opt}$ \\ \hline\hline
 \multirow{3}{*}{Synthetic Face}       & DenseNet  & 0.528 $\pm$ 0.050 & 0.615 $\pm$ 0.051 & 0.670 $\pm$ 0.042 & 0.619 $\pm$ 0.032 & 0.645 $\pm$ 0.020 & \textbf{0.714 $\pm$ 0.013} \\ \cline{2-8} 
                                      & ResNet & 0.526 $\pm$ 0.057 & 0.565 $\pm$ 0.063 & 0.639 $\pm$ 0.036 & 0.617 $\pm$ 0.046 & \textbf{0.675 $\pm$ 0.040} & 0.669 $\pm$ 0.024 \\ \cline{2-8} 
                                      & Inception & 0.555 $\pm$ 0.033 & 0.581 $\pm$ 0.037 & 0.675 $\pm$ 0.039 & 0.628 $\pm$ 0.047 & 0.651 $\pm$ 0.022 & \textbf{0.704 $\pm$ 0.024} \\ \hline\hline
\multirow{3}{*}{Iris PAD}             & DenseNet  & 0.881 $\pm$ 0.022 & 0.900 $\pm$ 0.012 & N/A & 0.911 $\pm$ 0.014 & 0.912 $\pm$ 0.015 & \textbf{0.929 $\pm$ 0.009} \\ \cline{2-8} 
                                      & ResNet    & 0.885 $\pm$ 0.024 & 0.897 $\pm$ 0.018 & N/A & 0.916 $\pm$ 0.008 & 0.904 $\pm$ 0.013 & \textbf{0.921 $\pm$ 0.019} \\ \cline{2-8} 
                                      & Inception & 0.877 $\pm$ 0.023 & 0.894 $\pm$ 0.022 & N/A & 0.905 $\pm$ 0.011 & 0.909 $\pm$ 0.011 & \textbf{0.917 $\pm$ 0.017} \\ \hline\hline
\multirow{3}{*}{\begin{tabular}[c]{@{}c@{}}Abnormality \\ from CXR\end{tabular}} & DenseNet  & 0.734 $\pm$ 0.024 & 0.739 $\pm$ 0.010 & N/A & 0.756 $\pm$ 0.004 & 0.757 $\pm$ 0.003 & \textbf{0.762 $\pm$ 0.004} \\ \cline{2-8} 
                                      & ResNet    & 0.733 $\pm$ 0.005 & 0.741 $\pm$ 0.007 & N/A & 0.750 $\pm$ 0.007 & 0.748 $\pm$ 0.003 & \textbf{0.755 $\pm$ 0.004} \\ \cline{2-8} 
                                      & Inception & 0.737 $\pm$ 0.007 & 0.744 $\pm$ 0.011 & N/A & 0.749 $\pm$ 0.009 & \textbf{0.753 $\pm$ 0.008*} & \textbf{0.753 $\pm$ 0.008*} \\ \hline
\end{tabular}
\label{tab:results}
\end{table*}

In order to apply CYBORG loss to the third domain of X-ray abnormality detection, we converted a multi-class dataset (originally 13 classes) into a binary abnormality present / no abnormality clssification.
Training, validation and testing splits can be seen in Tab. \ref{tab:num_samples}, as defined by the authors of the MIMIC Chest X-ray JPG (MIMIC-CXR-JPG) Database. ``Typical'' samples in this case refer to no abnormality present and atypical refers to scans showing an abnormality. 

\vskip2mm\noindent \textbf{Motivation for Selected Domain}

\vskip1mm\noindent{Acquisition} of data in the medical imaging domain is laborious and expensive. Labeled data requires expert annotation. Additionally, in many cases the acquisition of more data is impossible due to the rarity of some medical conditions, privacy issues, or the lack of the capture equipment. These limitations make it critical that we maximize the value of the data we do have. The eye-tracking data used in this work was collected using a non-intrusive device during routine report writing, meaning it required no additional effort from the radiologists. Results on this domain detail how small amounts of data can be enhanced using human saliency, increasing the value of each sample.

\vskip2mm
\noindent \textbf{Image Data} 

\vskip1mm
\noindent The MIMIC Chest X-ray JPG (MIMIC-CXR-JPG) Database v2.0.0 \cite{johnsonmimic,johnson2019mimic} is a publicly available dataset of chest radiographs with labels derived from 227,827 free-text radiology reports. 
This JPG version of the MIMIC-CXR dataset is derived from the original MIMIC-CXR dataset, which provided DICOM images and the corresponding free-text labels from the reports. The aim of MIMIC-CXR-JPG data was to provide a convenient processed version of MIMIC-CXR data, as well as to standardize reference for data splits and image labels. In total, the dataset contains 377,110 JPG format images and corresponding labels. 

As noted earlier, the original labels correspond to either healthy (no abnormality) or twelve possible abnormalities. In order to reduce this task from 13-class to binary classification, we grouped the 12 ``abnormal'' classifications under 1 class, simply labeled as abnormal. Future work includes extending CYBORG to multi-class classification. The dataset is de-identified in accordance with the Safe Harbor requirements of the US Health Insurance Portability and Accountability Act of 1996 (HIPAA). Protected health information (PHI) has also been removed. Given the breadth of the labeled, de-identified images, the data is intended to support a wide body of research  including image understanding, natural language processing, and decision support.

The training data was limited to images filtered as follows: images without classification labels were discarded; only frontal CXRs were kept, \ie images with ``ViewPosition'' equals to ``AP'' (anterior-posterior) or ``PA'' (posterior-anterior). Furthermore, studies with more than one frontal image were excluded.

\vskip2mm
\noindent \textbf{Human Saliency Data}

\vskip1mm
\noindent The human saliency data for CXR anomaly detection-based experiments came from the REFLACX dataset, ~\cite{lanfredi2021reflacx}, which builds upon the existing MIMIC-CXR dataset ~\cite{johnson2019mimic}. REFLACX offers annotations in the form of eye tracking data from radiologist sessions with a timestamped transcription of the dictated report. There are 3,032 labeled samples in the dataset from five radiologists; 109 of these samples have labels from all five radiologists for assessing inter-rater reliability. In addition to an image-level label, each scan was further labeled with ellipses that localized abnormalities and bounding boxes around the heart and lungs.

Eye-tracking saliency maps were generated by placing Gaussian distributions centered on each fixation point and combining them using a sum weighted by the fixation duration. Fixation points with a fixation duration of less than 150ms were discarded as this was determined to be the minimum time required for humans to process visual information \cite{Thorpe1996}. Following Le Meur \& Baccino \cite{le2013methods}, the Gaussian distributions had a standard deviation of 1 degree of visual angle in each axis to represent location uncertainties.

\section{Evaluation}
\label{sec:evaluation}

\begin{figure*}[t]
    \begin{subfigure}[b]{1\textwidth}
      \begin{subfigure}[b]{0.32\textwidth}
          \centering
            \includegraphics[width=1\columnwidth]{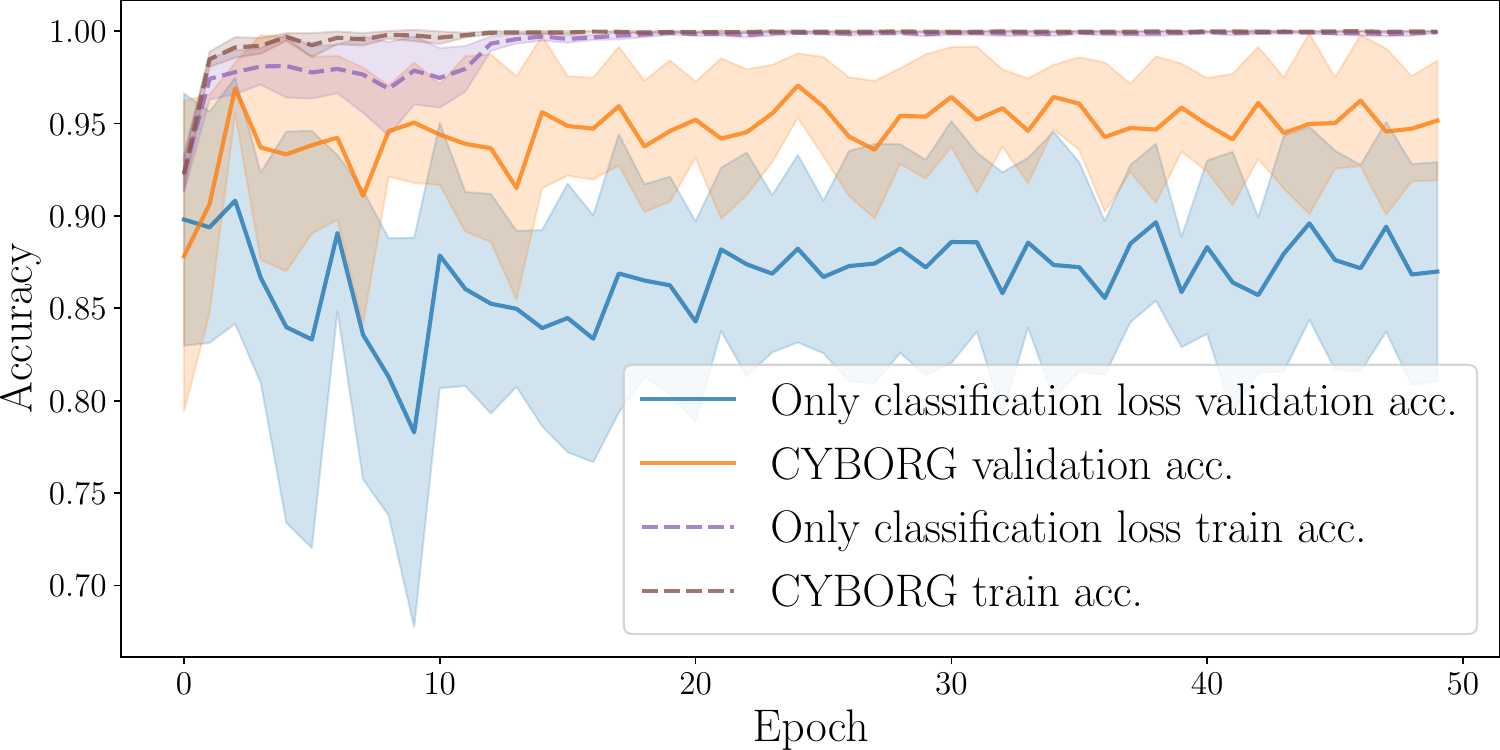}
            \caption{Synthetic Face Detection}
      \end{subfigure}
      \hfill
      \begin{subfigure}[b]{0.32\textwidth}
          \centering
          \includegraphics[width=1\columnwidth]{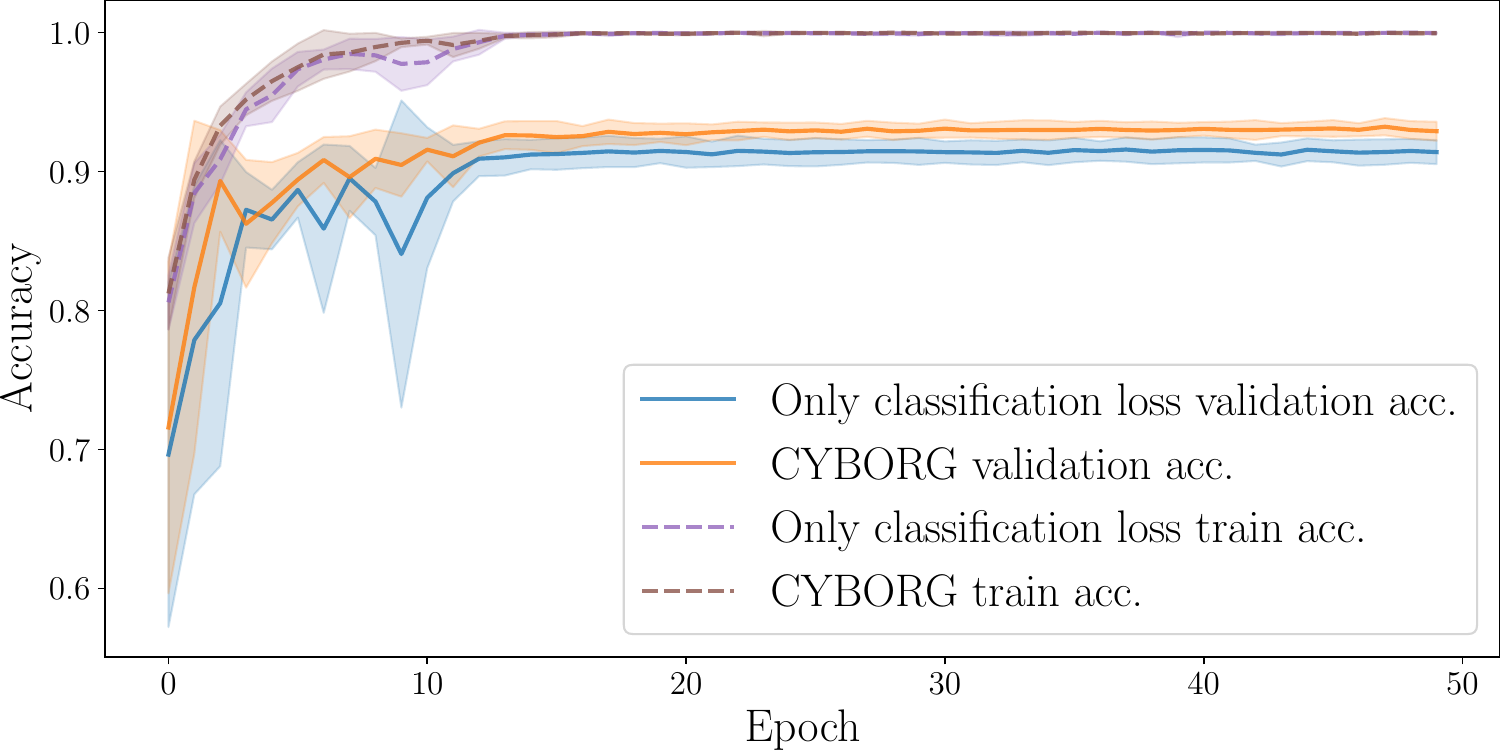}
          \caption{Iris PAD}
      \end{subfigure}
      \hfill
      \begin{subfigure}[b]{0.32\textwidth}
          \centering
          \includegraphics[width=1\columnwidth]{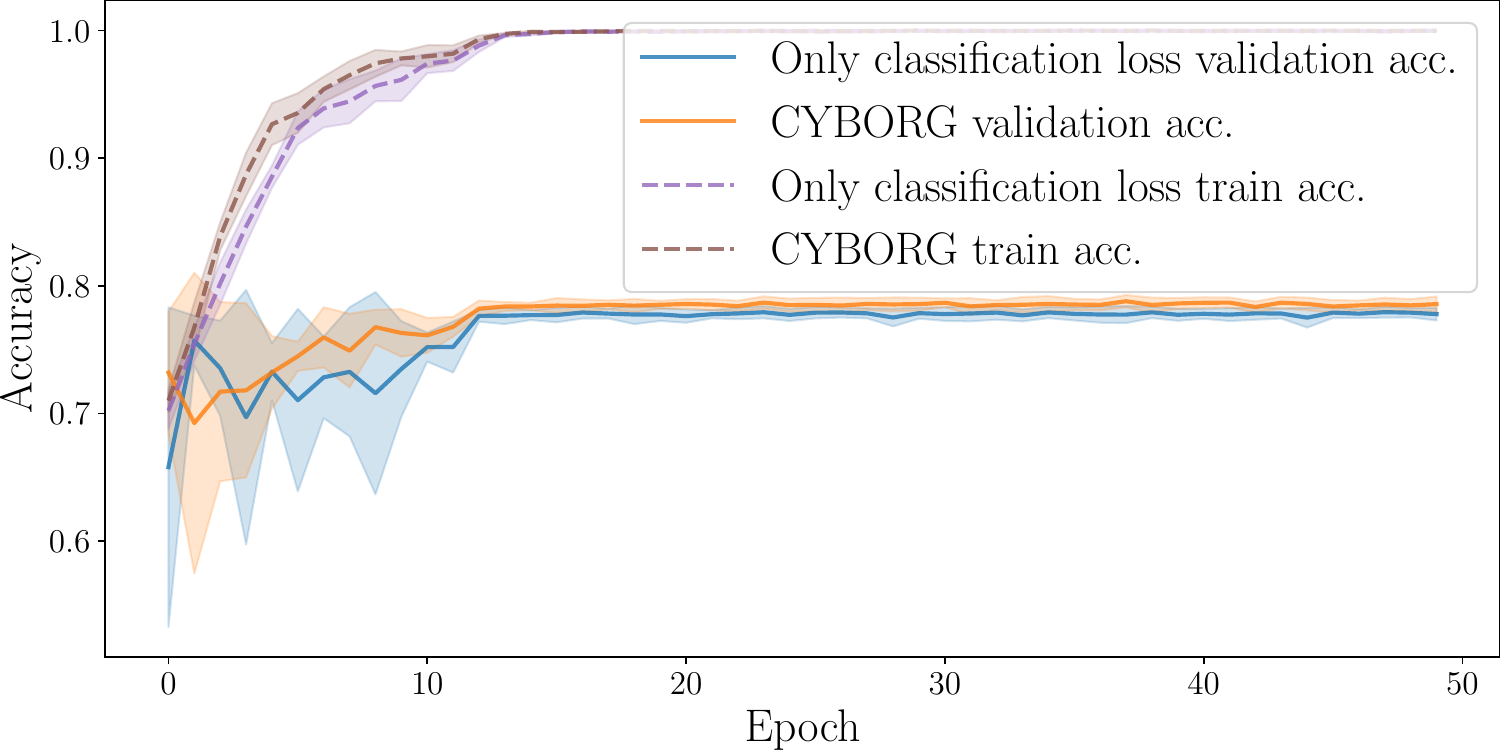}
          \caption{Abnormality from CXR}
      \end{subfigure}
    
  \end{subfigure} \vskip3mm
  \caption{Comparison of ResNet50 training and validation accuracy for CYBORG versus traditional training, across the three domains. 
  The CYBORG training approach achieves higher validation accuracy, indicating more effective, generalizable feature learning.
  Shaded area represents $\pm1$ standard deviation of the accuracy by epoch. }
  \label{fig:train_acc_plots}
\end{figure*}

An important note regarding the presented results is that the goal of this work was not to beat the state-of-the-art performance for any specific domain in a presence of ample amount of training data. Instead, the goal is to comprehensively demonstrate that the incorporation of human saliency into the loss function results in a significant improvement when the training data is {\it limited}. In other words, this framework allows for much better use of the existing training data, if human perceptual data is available. The baseline in this work is when the training procedure uses only traditional classification loss without any human saliency component, \ie the traditionally trained models. Because the training parameters (optimizer, learning rates, best model criteria, etc.) are kept constant for all experiments in this work, and the only variation is the human saliency component in the loss, this is a fair comparison. 
Future work may include the optimization of the training procedure such that performance in the individual domains can be increased. Additionally, for future work, CYBORG could be incorporated into current state-of-the-art methods for a specific domain to increase performance.

As mentioned in Sec. \ref{sec:domains}, the synthetic face detection domain represents an open-set style evaluation. Thus, the results in this domain will be comparably worse relative to the other domains. This is to be expected, as similar observations were made in \cite{boyd_pad_assessment_tifs}, which compared closed-set experimentation to open-set performance. Increases in performance in this domain represent a boost in generalization capabilities to unseen data sources.

Two metrics are used to evaluate the performance in this paper: Area Under the ROC Curve (AUC) and Average Precision (AP). 
We believe AUC to be more appropriate in this instance as it details the separation between the typical and atypical samples in a threshold-free way. This is important for open-set evaluation.
Average precision represents the area under the PR curve and indicates whether the model correctly identifies all positive samples without incorrectly classifying many negative samples as positive. 

The main results are in Tab. \ref{tab:results} for AUC and Tab. A in supplementary materials for AP. 
As the trends for AUC and AP are identical, all discussion will focus on the AUC. Tab. \ref{tab:results} table contains the average AUC $\pm 1\sigma$ across the 10 trained models for each of the three architectures in each of the three domains. \textit{Traditional} corresponds to the experiment setting without any human saliency included. CYBORG$_{gen}$, CYBORG$_{arch}$ and CYBORG$_{opt}$ all represent different parameter combination approaches for the CYBORG loss as detailed in Tab. \ref{tab:cyborg_settings}. 

\subsection{Does human-saliency-guided training produce a model with improved generalization? (RQ1)} 

Results in Tab. \ref{tab:results} show that in all cases the CYBORG-trained models achieve greater performance than traditionally-trained models. 
These results span three popular network architectures, each used to generate models in three different problem domains.
This demonstrates that \textbf{CYBORG training improves accuracy in a way that is not dependent on a particular network architecture or specific to a particular problem domain.}

For the CYBORG$_{opt}$ models, the performance difference for all nine results is greater than the standard deviation intervals.
The largest increase in performance is for the synthetic face detection problem, where CYBORG$_{opt}$ results in relative performance increases over the traditionally trained models of 35.23\%, 27.19\% and 26.85\% for DenseNet121, ResNet50 and Inception v3, respectively.

\subsection{Does human-salience-guided training improve robustness against overfitting? (RQ2)}

\begin{figure}[t]
    \centering
  \begin{subfigure}[b]{1\columnwidth}
      \begin{subfigure}[b]{0.82\textwidth}
          \centering
            \includegraphics[width=1\columnwidth]{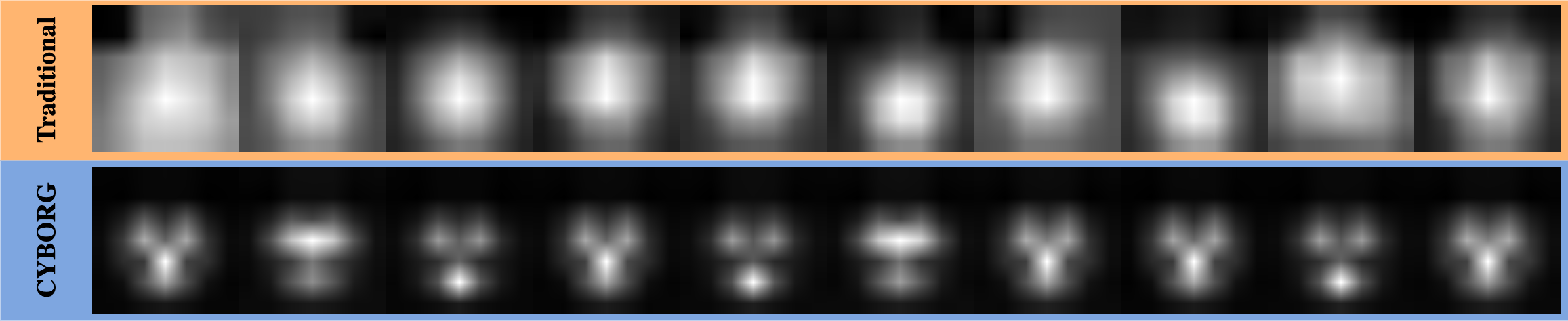}
      \end{subfigure}
      \hfill
      \begin{subfigure}[b]{0.169\textwidth}
          \centering
            \includegraphics[width=1\columnwidth]{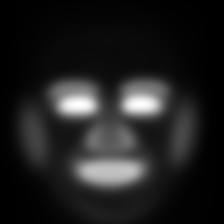}
      \end{subfigure}
    \caption{Synthetic Face Detection}
    \end{subfigure}
    \begin{subfigure}[b]{1\columnwidth}
      \begin{subfigure}[b]{0.82\textwidth}
          \centering
            \includegraphics[width=1\columnwidth]{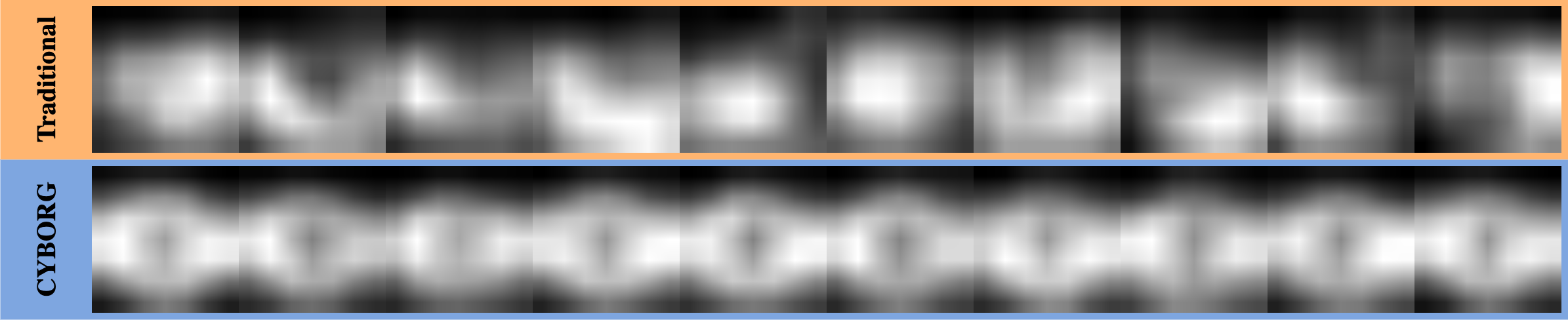}
      \end{subfigure}
      \hfill
      \begin{subfigure}[b]{0.169\textwidth}
          \centering
            \includegraphics[width=1\columnwidth]{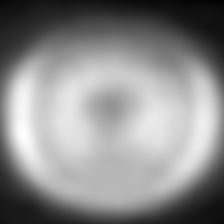}
      \end{subfigure}
    \caption{Iris PAD}
    \end{subfigure}

    \begin{subfigure}[b]{1\columnwidth}
      \begin{subfigure}[b]{0.82\textwidth}
          \centering
            \includegraphics[width=1\columnwidth]{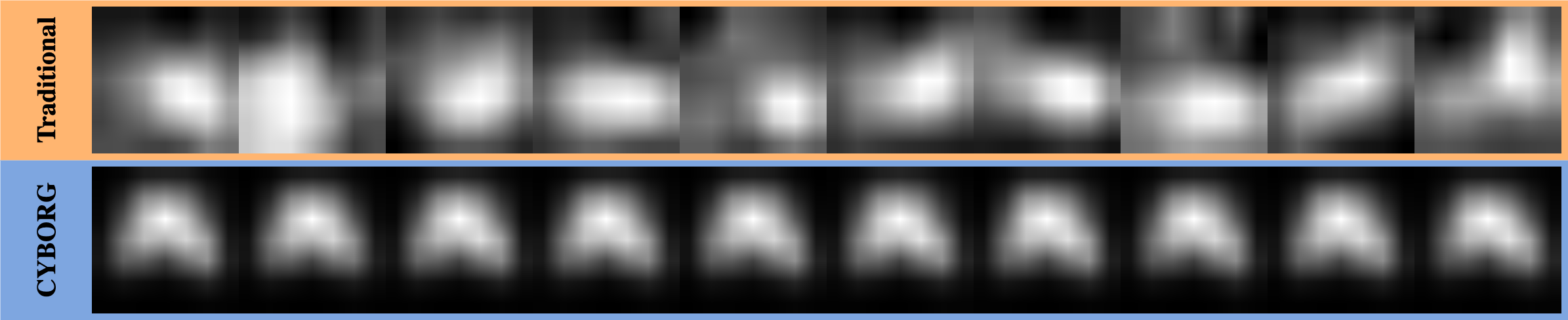}
            \caption*{Average model saliency on the test set.}

      \end{subfigure}
      \hfill
      \begin{subfigure}[b]{0.169\textwidth}
          \centering
            \includegraphics[width=1\columnwidth]{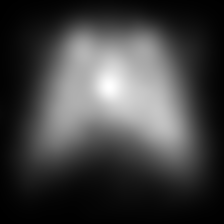}
            \caption*{Humans}
      \end{subfigure}
      \caption{Abnormality from CXR}
  \end{subfigure}
  \caption{Visualizations on the test set. The left image shows the model visualizations for the 10 trained models on the test set. The right image shows the average human saliency collected on the training set. The CYBORG trained models (blue box) use features more similar to the human saliency than traditionally trained models (orange box). Additionally, CYBORG models show much higher consistency across runs.}
  \label{fig:visualizations}
\end{figure}

The training and validation accuracy during the ResNet50 training for each of the domains are shown in Fig. \ref{fig:train_acc_plots}. (Plots for the other networks are similar and are included into supplementary materials.)  Training accuracy quickly approaches 100\% for both CYBORG and traditional training in all three cases.   However, CYBORG achieves higher validation accuracy throughout, indicating more effective learning.   CYBORG’s improvement in validation accuracy is largest for the problem of detecting synthetic face images, but there is also consistent improvement for iris PAD and for abnormality detection from CXR.  Clearly, CYBORG approach guides the training process to learn features that enable higher validation accuracy.  The CYBORG-learned features that achieve higher validation accuracy then also achieve higher test accuracy, showing that they are simply more effective.  CYBORG training also reaches its peak validation accuracy in fewer epochs than traditional training, suggesting that it enables models to converge at a faster rate.  Additionally, CYBORG validation accuracy appears less prone to sharp drops in accuracy between epochs, suggesting that it is overall more stable.  Overall, these results show that \textbf{CYBORG does reduce the tendency of the training process to overfit on the training data}.

\subsection{Does human-salience-guided training produce models that focus on human-salient regions? (RQ2)}

An underlying assumption of the CYBORG approach is that traditional training allows the model to form features using any element of the training images, resulting in features that can be based on incidental properties of the training data, whereas CYBORG training guides the model to learn features based on image regions judged salient by humans.
To show that this is true, CAM visualizations on the test set for the three domains can be seen in Fig. \ref{fig:visualizations}.  These visualizations are the average CAM generated on all samples in the test set, using the same mechanism as during training, for both traditional and CYBORG models, for each of the 10 independent trainings.  

The contrast between the CAMs for traditional and CYBORG trained models is striking.
The CAMs for traditionally-trained model uniformly lack a coherent focus on any particular region of the image.
Also, the variation in the CAM visualizations across the ten trials of traditional training is much larger than for CYBORG training.
Even though CYBORG uses human saliency maps with training images during training, the model that is learned keeps a similar focus when processing the images in the test set.
For all three of the domains, not one of the ten independent trials of traditional training came close to learning a model with the same coherence as one of the CYBORG models.
These visualizations show that \textbf{CYBORG training results in models that have a more coherent focus on the human-salient regions of the image, and multiple independent trials of CYBORG training on the same training data result in more consistent models than traditional training}.

\subsection{How useful is it to optimize CYBORG to architecture and problem domain? (RQ3)}

\begin{figure*}[t]
  \begin{subfigure}[b]{1\textwidth}
      \begin{subfigure}[b]{0.32\textwidth}
          \centering
            \includegraphics[width=1\columnwidth]{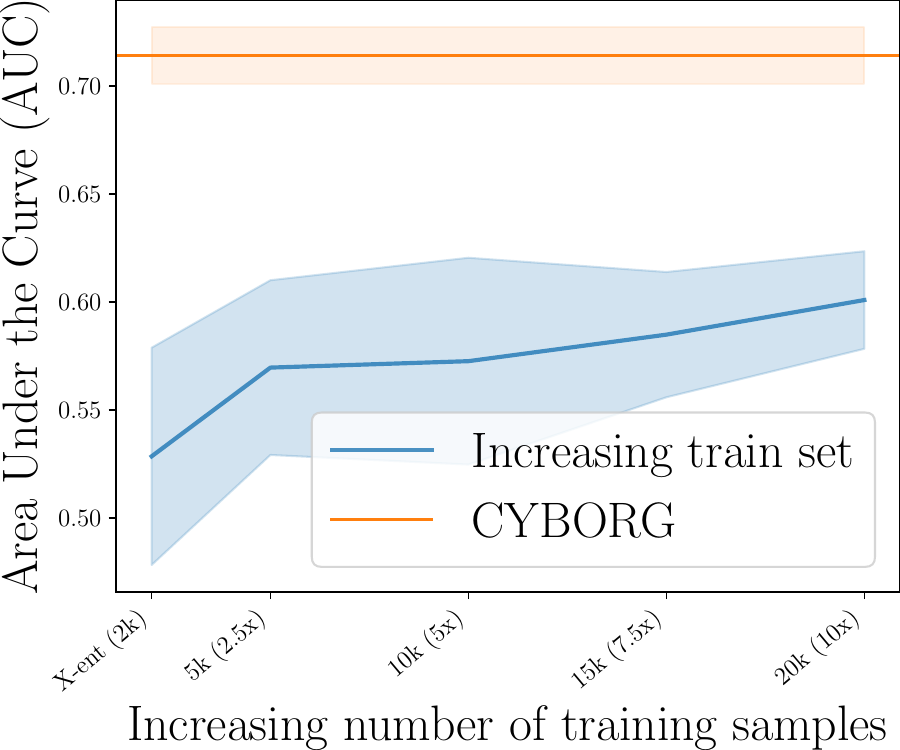}
      \end{subfigure}
      \hfill
      \begin{subfigure}[b]{0.32\textwidth}
          \centering
          \includegraphics[width=1\columnwidth]{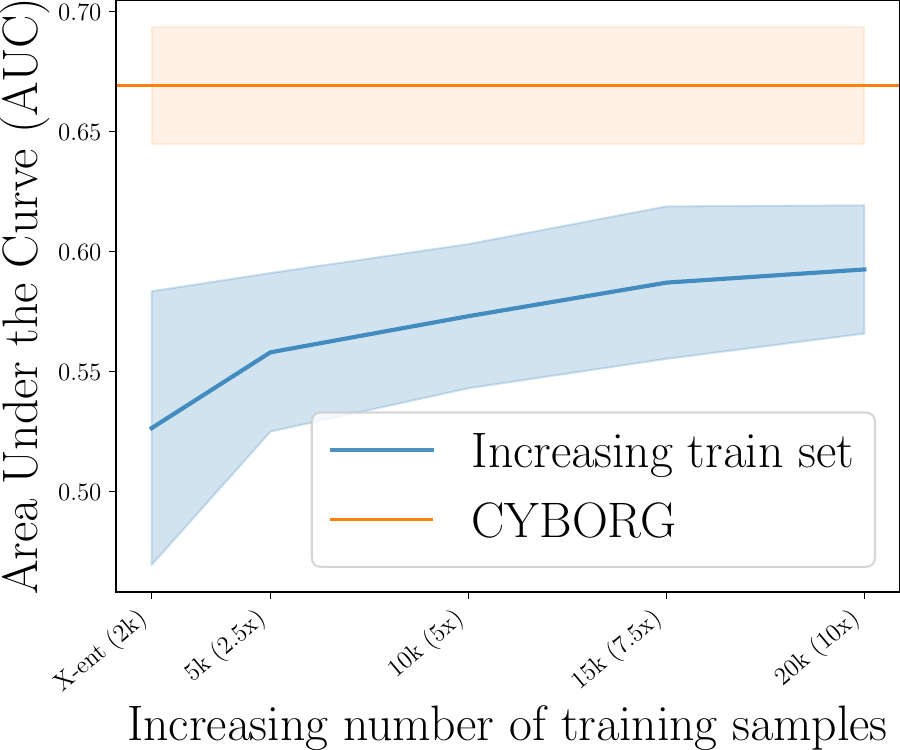}
      \end{subfigure}
      \hfill
      \begin{subfigure}[b]{0.32\textwidth}
          \centering
          \includegraphics[width=1\columnwidth]{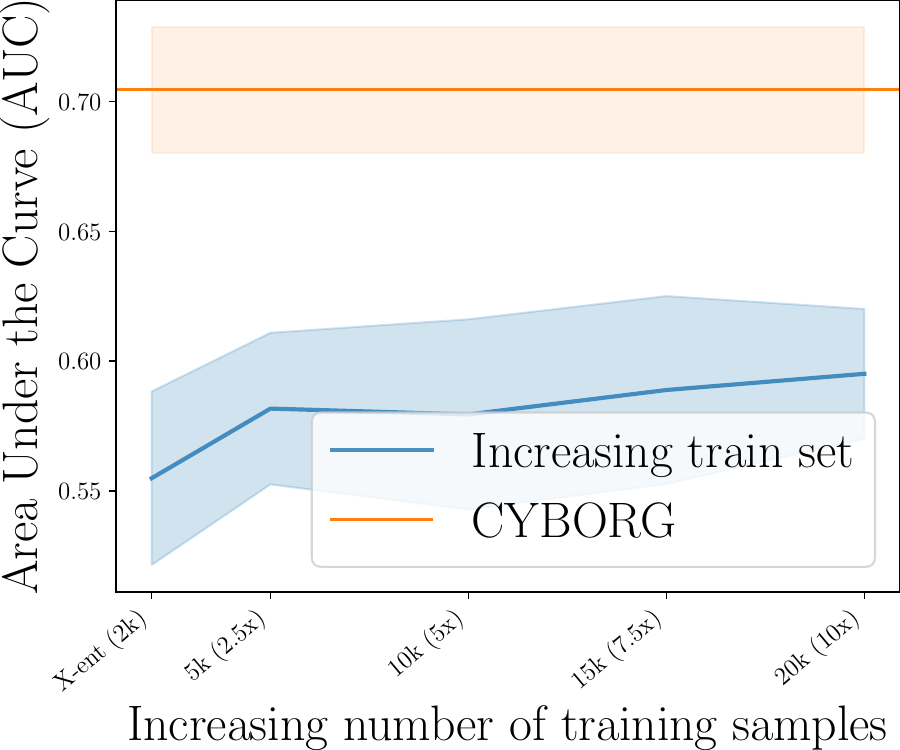}
      \end{subfigure}
\caption{Synthetic Face Detection}
  \end{subfigure}

    \begin{subfigure}[b]{1\textwidth}
      \begin{subfigure}[b]{0.32\textwidth}
          \centering
            \includegraphics[width=1\columnwidth]{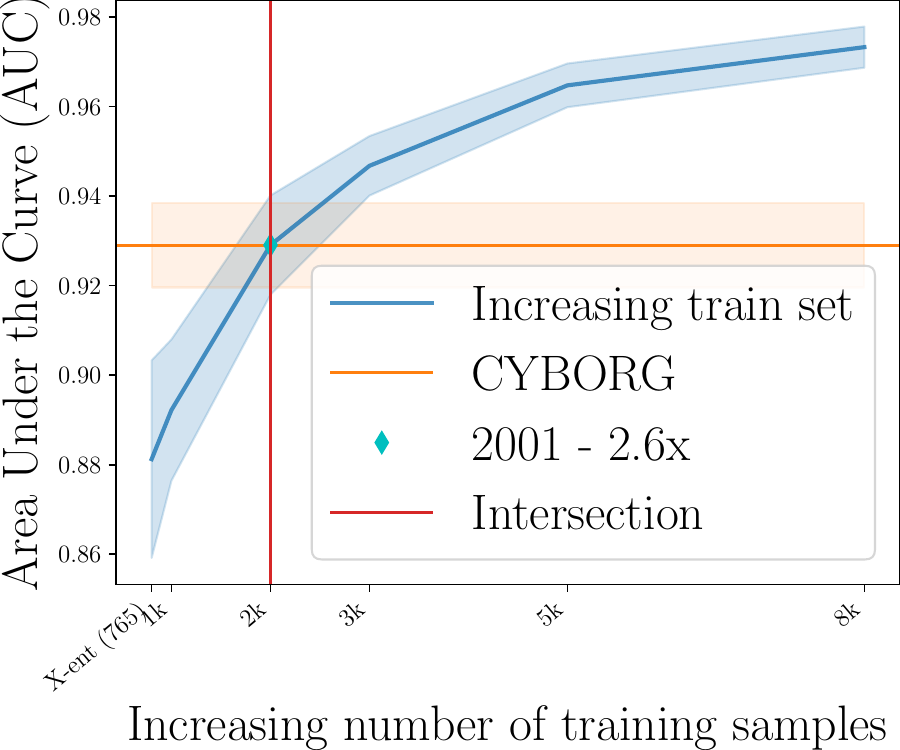}
      \end{subfigure}
      \hfill
      \begin{subfigure}[b]{0.32\textwidth}
          \centering
          \includegraphics[width=1\columnwidth]{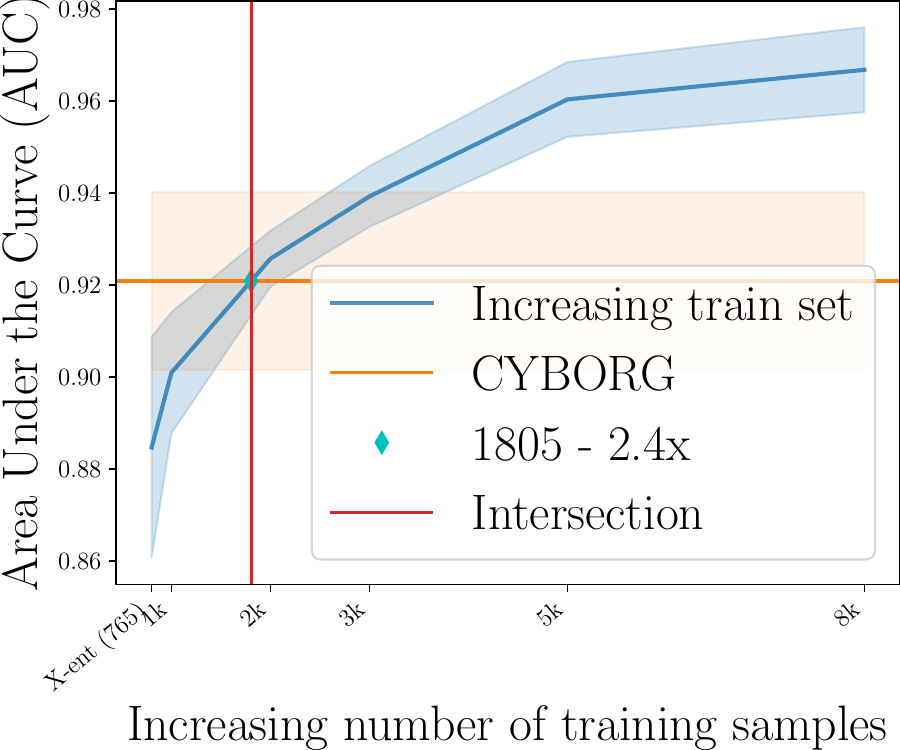}
      \end{subfigure}
      \hfill
      \begin{subfigure}[b]{0.32\textwidth}
          \centering
          \includegraphics[width=1\columnwidth]{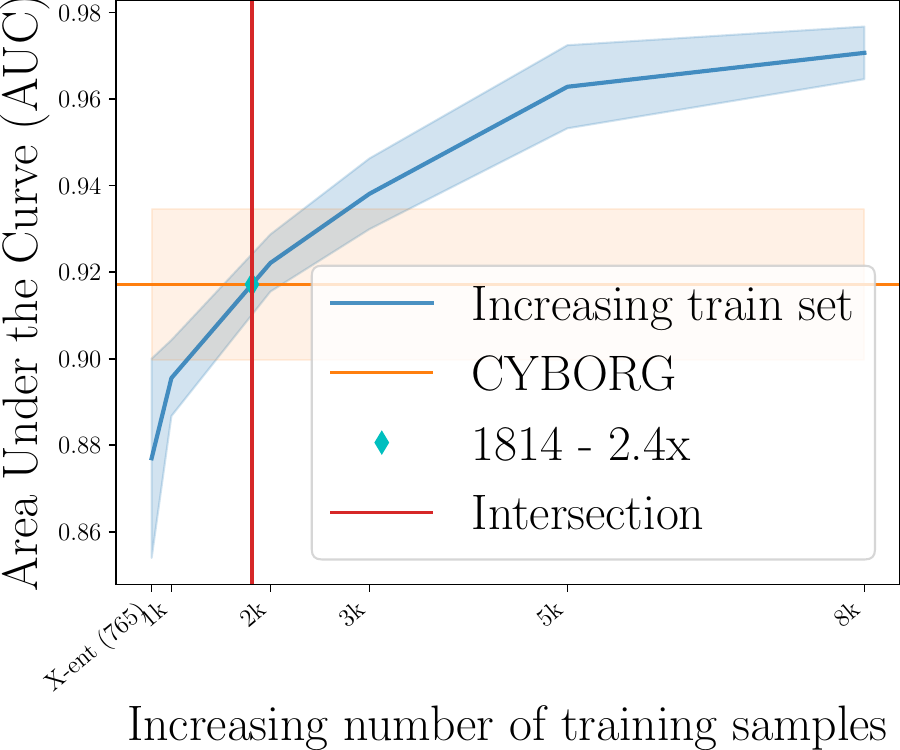}
      \end{subfigure}
\caption{Iris PAD}
  \end{subfigure}\vskip3mm

    \begin{subfigure}[b]{1\textwidth}
      \begin{subfigure}[b]{0.32\textwidth}
          \centering
            \includegraphics[width=1\columnwidth]{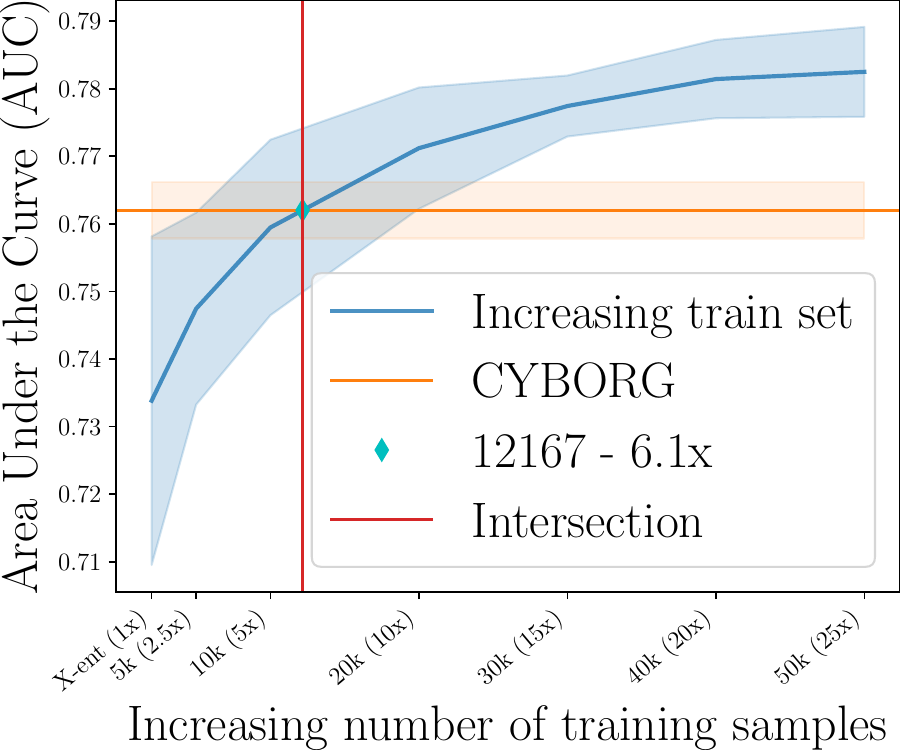}
          \caption*{DenseNet-121}
      \end{subfigure}
      \hfill
      \begin{subfigure}[b]{0.32\textwidth}
          \centering
          \includegraphics[width=1\columnwidth]{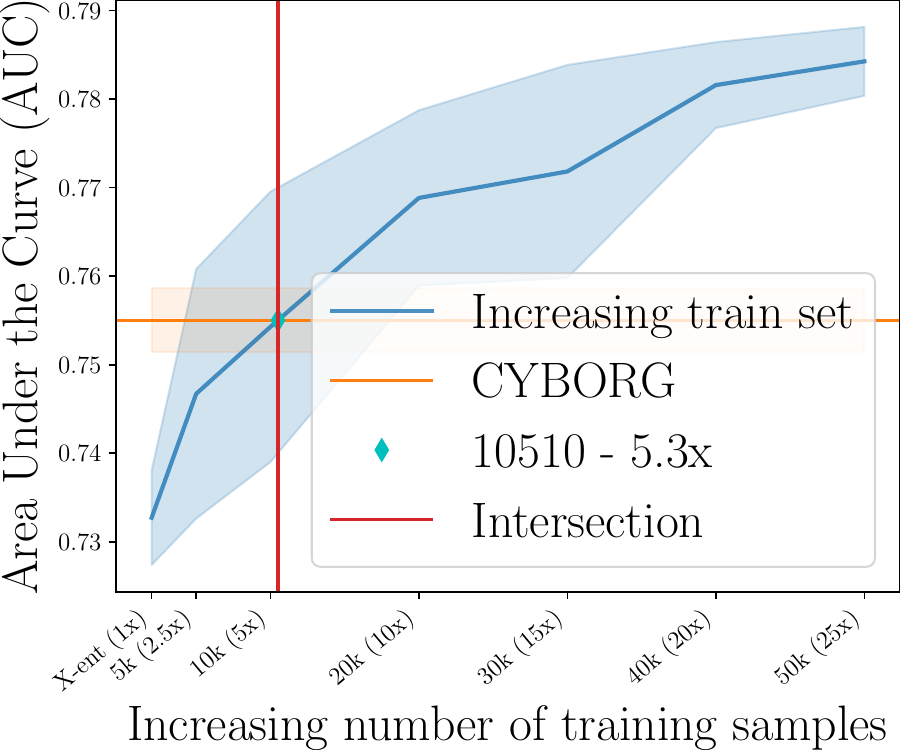}
          \caption*{ResNet50}
      \end{subfigure}
      \hfill
      \begin{subfigure}[b]{0.32\textwidth}
          \centering
          \includegraphics[width=1\columnwidth]{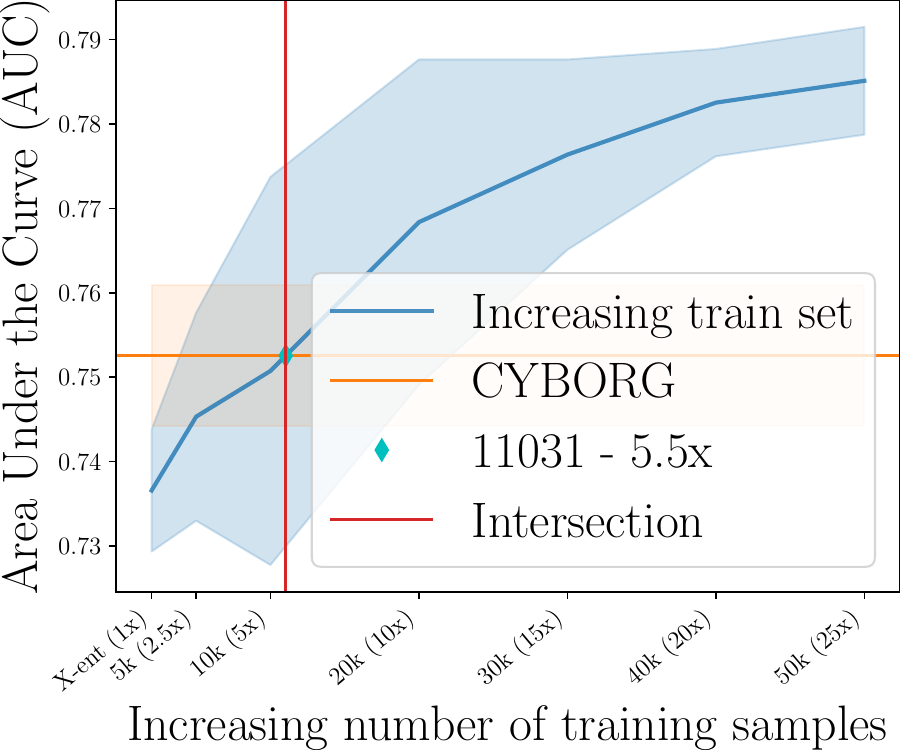}
          \caption*{Inception-v3}
      \end{subfigure}
\caption{Abnormality from CXR}
  \end{subfigure}
  \caption{Plots showing the intersection point of CYBORG and the addition of more data to the training set for traditionally trained models. The diamond outlines the number of samples required to match the performance and is also given as a multiple in size to the original set (the one used to train CYBORG).}
  \label{fig:iterative_all}
  \null\vskip-5mm
\end{figure*}

As seen in Tab. \ref{tab:results}, CYBORG$_{gen}$ shows large accuracy gains over traditionally trained models, across all three architectures and all three problem domains.
The CYBORG$_{gen}$ parameters (SSIM in loss term, $\alpha$ = 0.75 for blending saliency and cross-entropy) are good recommended parameter settings for initial experiments with a new architecture or problem domain.
The CYBORG$_{arch}$ parameter sets outperform CYBORG$_{gen}$ in 7 of 9 instances, which is remarkably good given that they are optimized only for network architecture and are problem domain invariant.
CYBORG$_{arch}$ achieves the largest gains over CYBORG$_{gen}$ for synthetic face detection, and achieves smaller gains and mixed results for the other two domains.
CYBORG$_{opt}$ achieves the highest accuracy in 8 of the 9 instances, with CYBORG$_{arch}$ having marginally higher accuracy in the remaining instance.
Thus, \textbf{even though the generic parameter settings for CYBORG result in accuracy improvement over traditional training for all three architectures and problem domains, it can still be worthwhile to optimize the two CYBORG parameters to the combination of architecture and problem domain}.

\subsection{How much extra training data does the traditional training require to achieve CYBORG accuracy? (RQ4)}
\label{sec:more_data}

Fig. \ref{fig:iterative_all} asks how much extra training data is required for traditional training to achieve the same accuracy as CYBORG training.
To have a common reference point across problem domains, results are described as a multiple of the original training set size.
For synthetic face detection, the authors ran out of training samples after expanding the set to $10\times$ the size of the original training data while maintaining identical class proportions as the original.
None of the three architectures achieved CYBORG level accuracy even with $10\times$ the size of the original training data.
For abnormality detection from CXR, DenseNet, ResNet and Inception required $6.1\times$, $5.3\times$ and $5.5\times$ the size of the original training data, respectively, to achieve CYBORG level accuracy.
For iris PAD, DenseNet, ResNet and Inception required $2.6\times$, $2.4\times$ and $2.4\times$ the size of the original training data, respectively, to achieve CYBORG level accuracy.
These results demonstrate that CYBORG training simply makes more effective use of the training data, to a level that traditional training cannot match in any of the nine instances with twice the size of training data and, for the synthetic face detection problem, even with $10\times$ the size of the training data.

An important point about the synthetic face detection problem is that {\it the test data is composed of GAN image sources not present in the training data}.
Thus this problem is evaluated in a more ``open set’’ manner. Traditionally trained models cannot effectively learn features from the training data that generalize well to the test set, whereas CYBORG is able to learn features from the training data that transfer remarkably to the test data.

For iris PAD, each sample with human annotations is worth roughly $2.5$ samples in traditional training.  In security applications such as this, the attack landscape is always changing.  New attacks and variations on known attacks are regular occurrences. Additionally, because of the unpredictable nature of presentation attacks, it may not be possible to collect substantial numbers of samples of new attacks. The ability to train models on fewer samples while still achieving greater generalization is paramount.  CYBORG excels at learning models that achieve the highest possibly accuracy from limited amounts of training data.

The result for CXR abnormality detection is significant because each one chest x-ray with eye-tracking data can be equated to six without.  In the field of medical imaging, acquiring additional HIPAA-compliant data is expensive, laborious and sometimes just not practical.  So being able to extract all the value possible from the available data is of utmost importance.  Also, the results in this problem domain show that saliency data can be effectively acquired through eye-tracking, so the data is collected passively during radiologists' normal work.  This shows the utility human saliency can have in a problem domain where such data may initially seem difficult to acquire.

The results in this section show that the accuracy gains achieved by {\bf CYBORG can equate to more than traditional training can achieve with $2\times$, $5\times$ or even more than $10\times$ as much training data. It can be more effective to collect additional information,  in the form of human saliency,  with a smaller number of training samples than it is to collect much larger amounts of training data.}

\begin{table*}[t]
\centering
\caption{Overall \textbf{Area Under ROC Curve (AUC)} results for all experimentation in which random noise, inverted saliency and a Gaussian kernel is used in place of human saliency.}
\begin{tabular}{|c|c||c||c|c|c||c|}
\hline
  Application & Network & Traditional & Random Noise & Inverted Saliency & Gaussian Kernel & CYBORG$_{opt}$ \\ \hline\hline
 \multirow{3}{*}{Synthetic Face}      & DenseNet  & 0.528 $\pm$ 0.050 & 0.559 $\pm$ 0.048 & 0.460 $\pm$ 0.046 & \textbf{0.754 $\pm$ 0.024} & 0.714 $\pm$ 0.013 \\ \cline{2-7} 
                                      & ResNet    & 0.526 $\pm$ 0.057 & 0.598 $\pm$ 0.049 & 0.554 $\pm$ 0.049 & \textbf{0.695 $\pm$ 0.017} & 0.669 $\pm$ 0.024 \\ \cline{2-7} 
                                      & Inception & 0.555 $\pm$ 0.033 & 0.675 $\pm$ 0.032 & 0.662 $\pm$ 0.055 & \textbf{0.738 $\pm$ 0.036} & 0.704 $\pm$ 0.024 \\ \hline\hline

\multirow{3}{*}{Iris PAD}             & DenseNet  & 0.881 $\pm$ 0.022 & 0.889 $\pm$ 0.013 & 0.823 $\pm$ 0.035 & 0.848 $\pm$ 0.015 & \textbf{0.929 $\pm$ 0.009} \\ \cline{2-7} 
                                      & ResNet    & 0.885 $\pm$ 0.024 & 0.897 $\pm$ 0.018 & 0.830 $\pm$ 0.025 & 0.875 $\pm$ 0.014 & \textbf{0.921 $\pm$ 0.019} \\ \cline{2-7} 
                                      & Inception & 0.877 $\pm$ 0.023 & 0.879 $\pm$ 0.017 & 0.823 $\pm$ 0.025 & 0.867 $\pm$ 0.021 & \textbf{0.917 $\pm$ 0.017} \\ \hline\hline

\multirow{3}{*}{\begin{tabular}[c]{@{}c@{}}Abnormality \\ from CXR\end{tabular}} 
                                      & DenseNet  & 0.734 $\pm$ 0.024 & 0.699 $\pm$ 0.016 & 0.444 $\pm$ 0.064 & 0.726 $\pm$ 0.030 & \textbf{0.762 $\pm$ 0.004} \\ \cline{2-7} 
                                      & ResNet    & 0.733 $\pm$ 0.005 & 0.725 $\pm$ 0.007 & 0.581 $\pm$ 0.075 & 0.734 $\pm$ 0.003 & \textbf{0.755 $\pm$ 0.004} \\ \cline{2-7} 
                                      & Inception & 0.737 $\pm$ 0.007 & 0.725 $\pm$ 0.025 & 0.470 $\pm$ 0.057 & 0.743 $\pm$ 0.009 & \textbf{0.753 $\pm$ 0.008*} \\ \hline
\end{tabular}
\label{tab:extra_results}
\end{table*}
\subsection{How does salience from selected segmentation algorithms compare to human salience? (RQ4)}

\subsubsection{Coarse segmentation masks}

Up to this point, our CYBORG experiments have used saliency maps derived from human input on each training image.
For the chest x-ray domain, the saliency maps were derived from eye-tracking data rather than as explicit manual annotations.
In this section, we ask if useful salience data can be obtained from an automated segmentation algorithm selected with some knowledge about the problem domain.

For all three domains, as described in Sec. \ref{sec:face_segmentation}, deep learning models are selected to segment the overall important regions. For synthetic face detection, the model extracts the facial region excluding hair and neck. For iris PAD, the iris is localized and eyelids/eyelashes are excluded. For abnormality from chest x-ray, the mediastinum and the bilateral hemidiaphragms are extracted.  The automatically segmented image regions are used in place of human annotation of the salient regions.  This experimental setting is referred to as CYBORG-DL.  CYBORG-DL is optimized in the same way as CYBORG$_{opt}$ (see Sec. \ref{sec:CYBORG_params}), \ie fully optimized to both the architecture and domain, and so can be compared to CYBORG$_{opt}$ results in Tab. \ref{tab:results}.

Interestingly, in all nine instances, CYBORG-DL outperforms traditionally trained models. This shows the value of the CYBORG approach, even when using automated region segmentations in place of human saliency annotations.  CYBORG learning still guides the learning to defined regions, resulting in features that generalize better than those learned with traditional training.  

However, in all nine instances, CYBORG$_{opt}$ significantly outperforms CYBORG-DL. This shows that \textbf{while CYBORG improves on traditional training even if automated segmentations are used for saliency, the highest accuracy is achieved using human saliency maps.}
The per-sample detail of the human saliency maps means the it guides models to more useful features on a per-image basis.  General region segmentation algorithms could perhaps be adapted to provide segmentations more specifically related to salience.

\begin{figure}[t]
\centering
    \begin{subfigure}[b]{1\columnwidth}
      \begin{subfigure}[b]{0.4\textwidth}
          \centering
            \includegraphics[width=1\columnwidth]{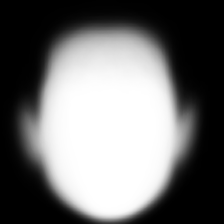}
            \caption{Face}
      \end{subfigure}
      \centering
      \begin{subfigure}[b]{0.4\textwidth}
          \centering
          \includegraphics[width=1\columnwidth]{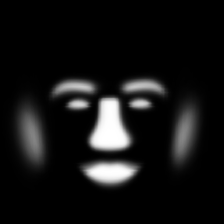}
        \caption{Face - Human-inspired}

      \end{subfigure}
      \hfill
  \end{subfigure}\vskip3mm
  \begin{subfigure}[b]{1\columnwidth}
      \begin{subfigure}[b]{0.4\textwidth}
          \centering
            \includegraphics[width=1\columnwidth]{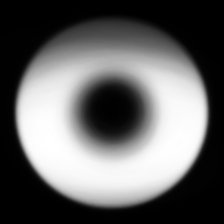}
          \caption{Iris}
      \end{subfigure}
      \centering
      \begin{subfigure}[b]{0.4\textwidth}
          \centering
          \includegraphics[width=1\columnwidth]{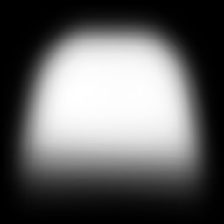}
          \caption{Chest X-Ray}
      \end{subfigure}
  \end{subfigure}\vskip3mm
  \caption{Average Deep Learning-Based Segmentation Maps on the training data. These figures are calculated in the same way as the average human saliency maps in Fig. \ref{fig:visualizations}. }
  \label{fig:average_maps}
  \null\vskip-5mm
\end{figure}

\subsubsection{Human-inspired fine segmentation masks}

When comparing the average human saliency map for synthetic face detection to both iris PAD and abnormality from CXR (Fig. \ref{fig:average_maps}), it is clear that the features are much more specific and constant. Annotators seem to primarily focus on the eyes, eyebrows, nose, mouth and ears. This begs the question: can we use a more fine-grained segmentation to extract these regions to use with CYBORG? Using the same BiSeNet model \cite{bisenet2019} as for the CYBORG-DL experiment, we can extract these specific features on a per-sample basis. These segmentation masks will be referred to as human-inspired fine masks, since the human saliency maps suggested the finer-detail features to extract. Fig. \ref{fig:average_maps} also shows the average human-inspired fine mask on the same data. This experiment will be referred to as CYBORG-DL-Fine.

Similar to the CYBORG-DL experiment, optimization of the CYBORG approach is done in the same way as for CYBORG$_{opt}$, but with the human-inspired fine masks. Results can be seen in Tab. \ref{tab:results}. CYBORG-DL-Fine shows a clear improvement of CYBORG-DL, showing that more specific and domain-aware segmentation masks can boost performance. However, CYBORG-DL-Fine still does not surpass the performance of CYBORG$_{opt}$.  Even though the human-inspired finer-detail masks can improve performance over the more coarse segmentations, they do not capture the complexity of the human annotations. Future work could include generating the human-inspired fine masks on the larger dataset described in Sec. \ref{sec:more_data}.  Currently, there are no human-inspired finer-detail mask equivalents for iris PAD or abnormality detection from CXR.

\subsubsection{Alternative Human Saliency Replacements and Modifications}

To further investigate the value of human saliency in the CYBORG approach, experiments were conducted with models that replaced the human saliency maps with random uniform noise, inverted human saliency, and a 2D Gaussian kernel. Using the random noise in place of actual saliency explores whether CYBORG is truly guiding the models to human-salient image regions, or if it is simply performing network regularization. Inverted saliency guides the model towards the opposite of what humans deem important. The Gaussian kernel focuses the model tightly on the center of the image.

Results for these three experiments are demonstrated in Tab. \ref{tab:extra_results}. As expected, using random noise used instead of actual saliency does not achieve the performance anywhere close to when human perception information is utilized. It does, however, serve as a regularizer in some cases, and thus improves the accuracy for iris PAD and synthetic face detection compared to cross-entropy-only training. Oppositely, performance degrades when using random noise for anomaly detection from CXR scans.

For all tasks, models trained on inverted model saliency saw decreased performance, with the largest decreases in synthetic face detection and abnormality detection from CXR. Even worse: using inverted saliency produces results inferior to using classification loss alone. This result validates the correctness and utility of human-sourced saliency maps in all domains.

Interestingly, the performance of synthetic face detection does  increase when training with Gaussian kernels instead of human saliency. However, we found that this performance increase stemmed from the preprocessing of face images (alignment and cropping). As such, the Gaussian kernel, tightly focused on inner face features, matched pretty well an average human saliency region. This suggests that for highly pre-processed and aligned data (such as center-cropped and normalized face images), using a Gaussian kernel can substitute human saliency in the task of synthetic sample detection. For iris PAD and detection of abnormality from CXR, hence two domains, in which spatial location of salient features is unpredictable, using a Gaussian kernel did not show significant performance increases over using classification alone. In these two domains humans provided strong salient regions that increase the models' focus.

\subsection{Salience-modified training data or salience-aware loss function? (RQ5)}

In earlier work Boyd \etal \cite{boyd2022humanblur} used human saliency information to directly modify the training data. Regions of a training image were blurred in an amount inversely proportional to the human saliency maps. Densely annotated regions were left un-blurred and un-annotated regions are blurred to a maximal strength. This effectively removed information deemed by human annotators to be not salient to the problem.  Conversely, CYBORG uses human saliency maps as additional information during training, without modifying the original images. This section compare the two approaches to determine which approach to using human saliency is most effective.

The data and training procedure for this work described in Sec. \ref{sec:experiments} is the  same as in \cite{boyd2022humanblur}, so a direct comparison of results is possible. However, only DenseNet is studied in \cite{boyd2022humanblur}.  For this evaluation, we run the same experiments on the two additional network architectures.  Results attained running the experiments were as follows: 0.890 $\pm$ 0.009, 0.891 $\pm$ 0.011 and 0.883 $\pm$ 0.014 for DenseNet, ResNet and Inception respectively. That is, CYBORG achieves much better accuracy than our previous approach based on information removal. An important note is that the large difference between the traditionally trained models in this work and \cite{boyd2022humanblur} is because in that work, Gaussian blur augmentations are incorporated into the training procedure, which actually degraded performance, but was a fairer comparison to the proposed method. Thus, we conclude that \textbf{CYBORG is a more effective incorporation strategy for human saliency information}. Due to the large difference in performance between \cite{boyd2022humanblur} and CYBORG, it was decided that it was not worth studying the previous method on all domains.

\section{Conclusion}
\label{sec:conclusions}

We have shown how human judgement about the salient regions of an image can be incorporated into the loss function to train better-performing deep CNNs. Through the guidance incorporated into the loss function, models learn with a preference for extracting information from regions deemed salient by humans. Our approach is compared to traditional deep CNN training through extensive experiments, using three popular CNN backbones to solve tasks in three different domains. CAM visualizations confirm that CYBORG-trained models do in fact focus on image regions judged as salient by humans, in contrast to traditionally-trained models, which show a fundamentally less coherent focus (Fig. \ref{fig:visualizations}). Performance results demonstrate the advantages of CYBORG training, and that it can be applied across different CNN backbones and different problem domains.  CYBORG models  generalize better, as seen in Fig. \ref{fig:train_acc_plots}. And CYBORG models achieve equivalent or better accuracy while requiring a smaller amount of training data (Tab. \ref{tab:results}).
 
It is natural to ask whether it is advantageous to have human perception applied on a per-image basis, or whether a human-inspired problem-relevant automatic segmentation masks could be used. Results show that the latter does result in an improvement relative to traditionally-trained models. However, automatic segmentation of image regions does not achieve the accuracy that per-image human-derived perception does. a more effective approach to reduce the effort required to obtain human-derived salience information is the use of eye-tracking while humans perform the task in the normal manner.
 
In previous work, we incorporated human saliency information by blurring less salient regions from the training images. Our CYBORG approach of incorporating human saliency  into the loss function improves upon our prior approach. The modified loss function encourages the model to focus on human-salient regions while still using all available information in the training images.
 
The ability to compare CAMs for the CYBORG-trained model to heat maps for human salience is also an important element of explainable and reliable AI. Substantial deviations between CAMs and human-salience heatmaps would indicate that learned models are less explainable and may have incorporated an accidental relationship in the training data.


\section*{Acknowledgments}

This work was supported by the U.S. Department of Defense (Contract No. W52P1J2093009). 
The views and conclusions contained in this document are those of the authors and should not be interpreted as representing the official policies, either expressed or implied, of the U.S. Department of Defense or the U.S. Government. Dr. Czajka was also partially supported by the National Science Foundation under grant No. 2237880. Any opinions, findings, conclusions, or recommendations expressed in this material are those of the author(s) and do not necessarily reflect the views of the National Science Foundation.

\ifCLASSOPTIONcaptionsoff
  \newpage
\fi

\bibliographystyle{IEEEtran}
\bibliography{main}

\begin{IEEEbiography}[{\includegraphics[height=1.25in,clip,keepaspectratio]{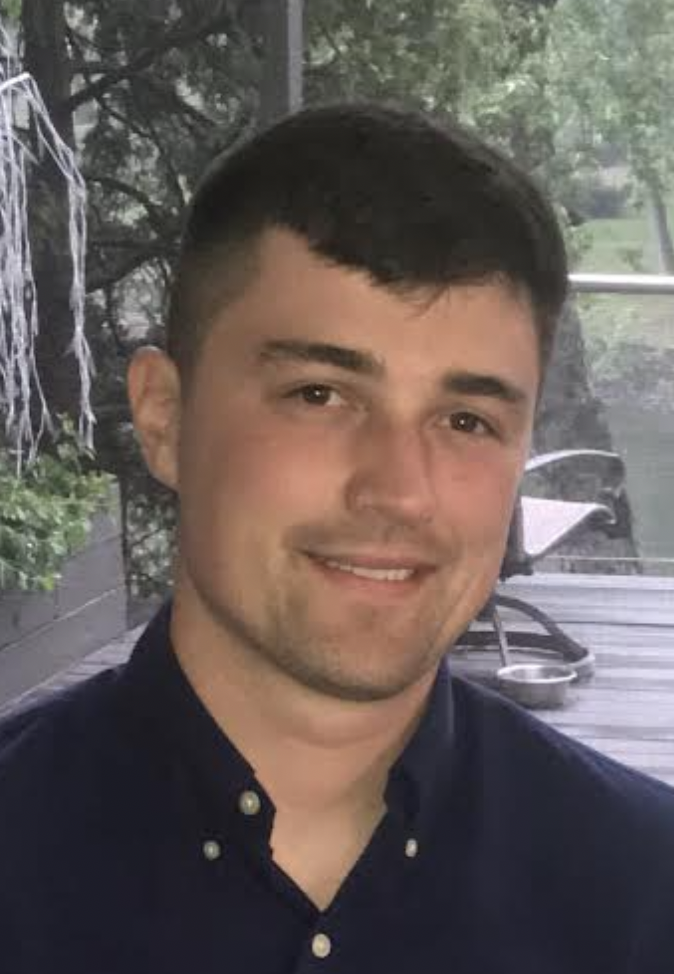}}]{Aidan Boyd} received his Ph.D from the University of Notre Dame in 2023 where he worked under advisors Dr. Adam Czajka and Dr. Kevin Bowyer. He graduated with a First Class Honours degree in Electronic and Computer Engineering from the National University of Ireland, Galway. He attained a Masters degree in Computer Science and Engineering from the University of Notre Dame in 2021. Aidan is now a Computer Vision Researcher at Nokia Bell Labs. His interests include computer vision, deep learning, biometrics and the incorporation of human intelligence into machine learning algorithms.
\end{IEEEbiography}
\vspace{-3em}
\begin{IEEEbiography}
[{\includegraphics[height=1.25in,clip,keepaspectratio]{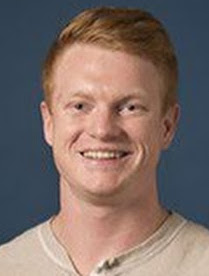}}]
{Patrick Tinsley} attained his PhD at the University of Notre Dame in 2024, advised by Dr. Adam Czajka and Dr. Patrick Flynn. He also attended Notre Dame for his Bachelors and Masters degrees (2017 and 2018). His undergraduate studies were in Applied and Computational Mathematics and Statistics with a specialization in predictive analytics. His interests include facial recognition, synthetic biometrics, and generative adversarial networks. Patrick now works as a Manager of Medical Technology at AngelEye Health.
\end{IEEEbiography}
\vspace{-3em}
\begin{IEEEbiography}[{\includegraphics[height=1.25in,clip,keepaspectratio]{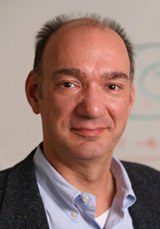}}]{Kevin W. Bowyer}
is the Schubmehl-Prein Family Professor of Computer Science and Engineering at the University of Notre Dame, and also serves as Director of International Summer Engineering Programs for the Notre Dame College of Engineering. In 2019, Professor Bowyer was elected as a Fellow of the American Association for the Advancement of Science. Professor Bowyer is also a Fellow of the IEEE and of the IAPR, and received a Technical Achievement Award from the IEEE Computer Society, with the citation ``for pioneering contributions to the science and engineering of biometrics.'' Professor Bowyer currently serves as the Editor-in-Chief of the {\it IEEE Transactions on Biometrics, Behavior and Identity Science}, and previously served as Editor-in-Chief of the {\it IEEE Transactions on Pattern Analysis and Machine Intelligence}.
\end{IEEEbiography}
\vspace{-3em}
\begin{IEEEbiography}[{\includegraphics[height=1.25in,clip,keepaspectratio]{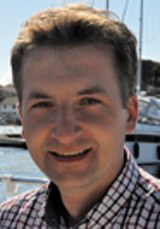}}]{Adam Czajka} (M'02--SM'12) is an Associate Professor in the Department of Computer Science and Engineering in the College of Engineering at the University of Notre Dame. He is a Senior Member of the Institute of Electrical and Electronics Engineers, Inc. (IEEE) and VP for Finance of the IEEE Biometrics Council. His research focuses on computer vision, biometrics and security, with a special interest in methods increasing reliability of biometric identification in adverse scenarios such as detection of unknown presentation attacks. He is the recipient of the NSF CAREER award. Dr Czajka's research has been funded by the US Department of Defense, US National Institute of Justice, FBI Biometric Center of Excellence, NIST, IARPA, US Army, US National Science Foundation, European Commission, Polish Ministry of Higher Education, and numerous companies.
\end{IEEEbiography}

\newpage

\renewcommand\thetable{\Alph{table}}

\appendix

\begin{table*}[!tb]
\centering
\caption{Overall \textbf{Average Precision (AP)} results for all experimentation. In all cases, CYBORG outperforms traditionally trained models. The N/A columns refer to cases when the experiment was not possible to perform. The $*$ refers to when the same configuration appears as best in two scenarios.}
\begin{tabular}{|c|c||c||c|c||c|c|c|}
\hline
  Application & Network & Traditional & CYBORG-DL & CYBORG-DL-Fine & CYBORG$_{gen}$ & CYBORG$_{arch}$ & CYBORG$_{opt}$ \\ \hline\hline
 \multirow{3}{*}{Synthetic Face}       & DenseNet  & 0.867 $\pm$ 0.021 & 0.902 $\pm$ 0.011 & 0.915 $\pm$ 0.016 & 0.899 $\pm$ 0.013 & 0.91 $\pm$ 0.007 & \textbf{0.93 $\pm$ 0.004} \\ \cline{2-8} 
                                       & ResNet    & 0.865 $\pm$ 0.024 & 0.892 $\pm$ 0.018 & 0.903 $\pm$ 0.014 & 0.898 $\pm$ 0.017 & \textbf{0.918 $\pm$ 0.014} & 0.915 $\pm$ 0.011 \\ \cline{2-8} 
                                       & Inception & 0.875 $\pm$ 0.013 & 0.878 $\pm$ 0.018 & 0.918 $\pm$ 0.015 & 0.898 $\pm$ 0.022 & 0.906 $\pm$ 0.011 & \textbf{0.926 $\pm$ 0.012} \\ \hline\hline
\multirow{3}{*}{Iris PAD}              & DenseNet  & 0.91 $\pm$ 0.016 & 0.925 $\pm$ 0.01 & N/A & 0.931 $\pm$ 0.01 & 0.933 $\pm$ 0.011 & \textbf{0.943 $\pm$ 0.008} \\ \cline{2-8} 
                                       & ResNet    & 0.911 $\pm$ 0.015 & 0.919 $\pm$ 0.017 & N/A & 0.935 $\pm$ 0.006 & 0.928 $\pm$ 0.009 & \textbf{0.939 $\pm$ 0.013} \\ \cline{2-8} 
                                       & Inception & 0.906 $\pm$ 0.019 & 0.921 $\pm$ 0.016 & N/A & 0.927 $\pm$ 0.009 & 0.928 $\pm$ 0.01 & \textbf{0.935 $\pm$ 0.016} \\ \hline\hline
\multirow{3}{*}{\begin{tabular}[c]{@{}c@{}}Abnormality \\ from CXR\end{tabular}} 
                                      & DenseNet  & 0.915 $\pm$ 0.008 & 0.915 $\pm$ 0.005 & N/A & 0.921 $\pm$ 0.003 & 0.921 $\pm$ 0.002 & \textbf{0.922 $\pm$ 0.002} \\ \cline{2-8} 
                                      & ResNet    & 0.911 $\pm$ 0.002 & 0.915 $\pm$ 0.004 & N/A & 0.918 $\pm$ 0.003 & 0.918 $\pm$ 0.002 & \textbf{0.92 $\pm$ 0.002} \\ \cline{2-8} 
                                      & Inception & 0.914 $\pm$ 0.004 & 0.917 $\pm$ 0.004 & N/A & 0.919 $\pm$ 0.004 & \textbf{0.921 $\pm$ 0.003*} & \textbf{0.921 $\pm$ 0.003*} \\ \hline
\end{tabular}
\label{tab:results_ap}
\end{table*}

\section*{Iris PAD Human Saliency Collection Details}

Upon presentation of an image, participants were asked to select the type of image they believed it to be (one of eight types as above or \textit{unsure}). Participants were then asked to highlight at least five regions of the image that support their decision about the type of image. The regions highlighted were not constrained on size or location within the image. The objective was to collect data on which regions of interest in an ISO-compliant iris image led humans (non-experts) to a correct \textit{bona fide/abnormal} classification decision. There are two reasons for using non-experts: (i) there are no experts formally trained in iris image examination (such experts do exist in, \eg, fingerprint analysis); (ii) to investigate whether or not human saliency from non-experts can boost model generalization for a given domain. 

Data collection was done for 150 participants, with each participant rating 30 image pairs, and annotating an image in 27 pairs, with an average of 3 pairs rated as unable to decide.
Images were assigned to users randomly such that an average of five subjects would annotate each image. 
Thus, not all images have the same number of salience annotations, and our proposed approach accounts for this in the averaging of individual annotations into an overall salience map for an image.

\section*{Synthetic Face Detection Image Source Details}

\vskip1mm\noindent{\textit{\textbf{CelebA-HQ}}}
~\cite{karras2017progressive} contains $1024\times1024$ versions of 30,000 celebrity images from the CelebA dataset~\cite{liu2015faceattributes}. 

\vskip1mm\noindent{\textbf{\textit{Flickr-Faces-HQ (FFHQ)}}} is a collection of 70,000 ($1024\times1024$) images from Flickr. Images show faces varying in age, ethnicity, gender, hairstyle, glasses, jewelry, etc.~\cite{karras2019style}. 

\vskip1mm\noindent{\textbf{\textit{FRGC-Subset}}} contains 16,433 faces, compiled from collections for the Face Recognition Grand Challenge etc~\cite{Phillips_IVC_2017}. Images show frontal faces varying in expression, ethnicity, gender, and age.

\vskip1mm\noindent{\textbf{\textit{SREFI}}} is an image dataset generated by the ``synthesis of realistic face images'' (SREFI)~\cite{Banerjee_IJCB_2017} method. The SREFI method matches similar \textit{real} face images based on VGG-Face features, splits them into region-specific triangles, and combines areas from donor images to create a blended identity. To ensure consistency, identity-salient facial features (such as the mouth and eyes) on the generated image are required to come from the same donor.

\vskip1mm\noindent{\textbf{\textit{ProGAN}}} features 100,000 images downloaded from~\cite{karras2017repo}. Unlike its successors (StyleGAN), ProGAN  was trained on the CelebA-HQ dataset described above~\cite{karras2017progressive}.

\vskip1mm\noindent{\textbf{\textit{StyleGAN}}} is the backbone for the next four synthetic datasets used in this work ~\cite{karras2019style, karras2020analyzing, Karras2020ada, Karras2021}. The original StyleGAN was trained in a similar fashion to its predecessor (ProGAN) \cite{karras2017progressive}, but with the added feature of mixable disentangled layers for style transfer. StyleGAN2 \cite{karras2020analyzing} removed artifacts found in original StyleGAN images and improved image reconstruction via path length regularization. StyleGAN2 with adaptive discriminator augmentation (SG2-ADA) \cite{Karras2020ada} solves for training GANs in data-limited scenarios. Finally, StyleGAN3 ~\cite{Karras2021} mitigates aliasing in rotation- and translation-invariant generator networks.

\vskip1mm\noindent For StyleGAN and StyleGAN2, sets of 100,000 fake face images were downloaded from their GitHub repositories. For StyleGAN2-ADA and StyleGAN3, sets of 100,000 images were generated using default settings, including the recommended truncation value ($\psi$) of 0.5. 

\vskip1mm\noindent{\textbf{\textit{StarGANv2}}} is a collection of mixed-style face images, as generated by StarGANv2 ~\cite{choi2020stargan}. The generated images show source identities ``dressed'' in the style of supplied reference images. In order to ensure high quality of the generated images, 250,000 images were initially synthesized using the supplied network (pre-trained on CelebA-HQ). The synthetic samples were then scored and sorted according to facial quality using FaceQNet~\cite{hernandez2020biometric}, which evaluates input images' suitability for face recognition tasks. The final dataset consisted of the top-ranked 100,000 images.

\section*{CYBORG Parameter Search}

\begin{figure*}[t]
  \begin{subfigure}[b]{1\textwidth}
      \begin{subfigure}[b]{0.19\textwidth}
          \centering
          \includegraphics[width=1\columnwidth]{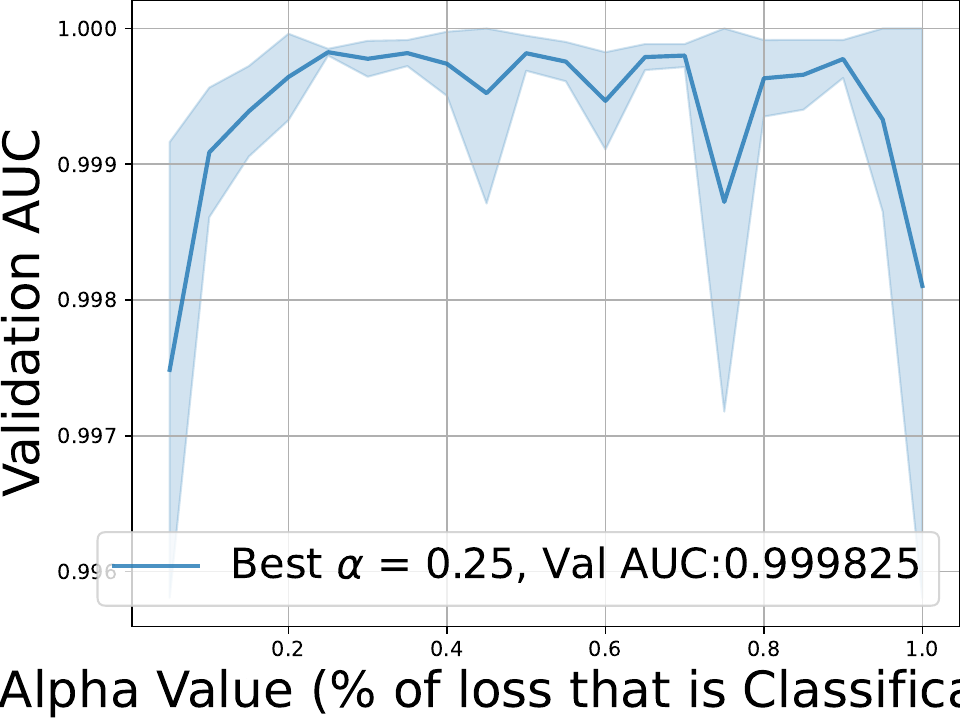}
          \caption{L1}
      \end{subfigure}
      \hfill
      \begin{subfigure}[b]{0.19\textwidth}
          \centering
          \includegraphics[width=1\columnwidth]{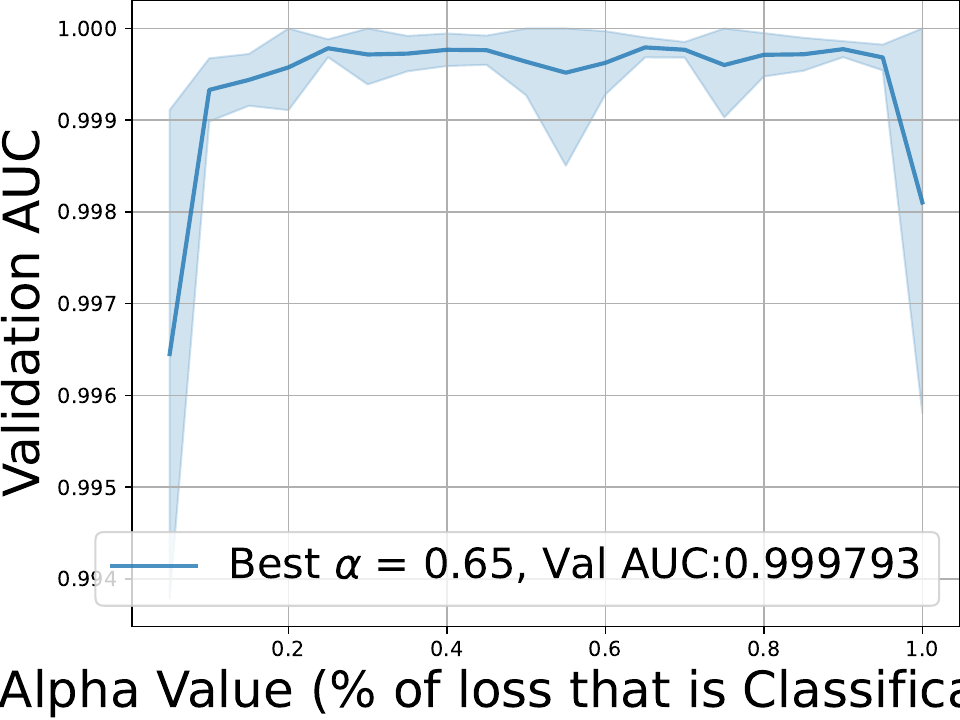}
          \caption{MSE}
      \end{subfigure}
      \hfill
      \begin{subfigure}[b]{0.19\textwidth}
          \centering
          \includegraphics[width=1\columnwidth]{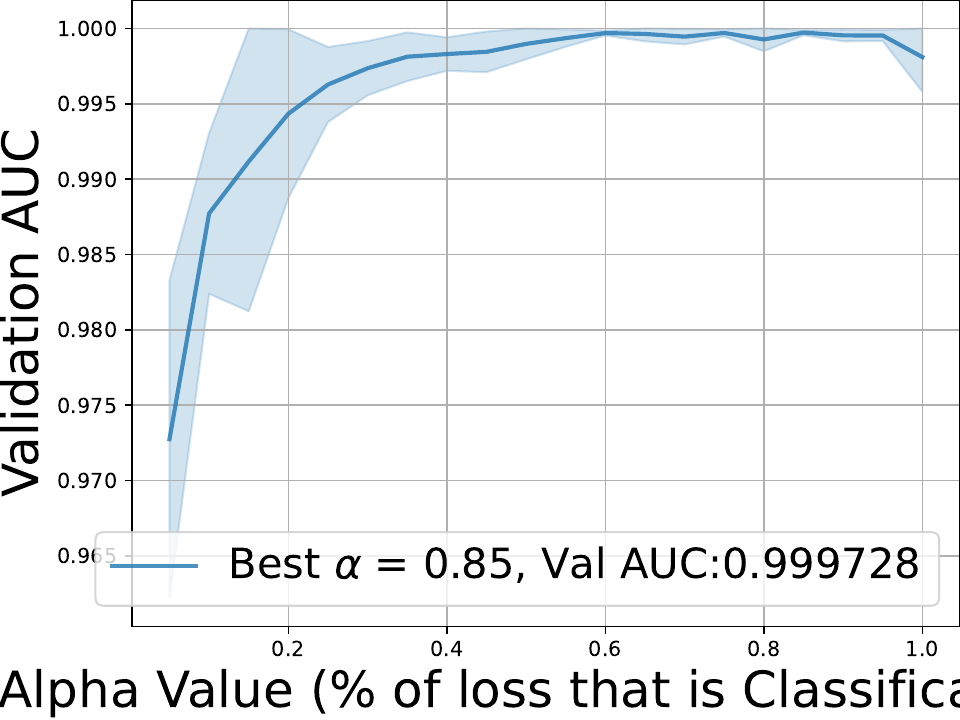}
          \caption{SSIM}
      \end{subfigure}
      \hfill
      \begin{subfigure}[b]{0.19\textwidth}
          \centering
          \includegraphics[width=1\columnwidth]{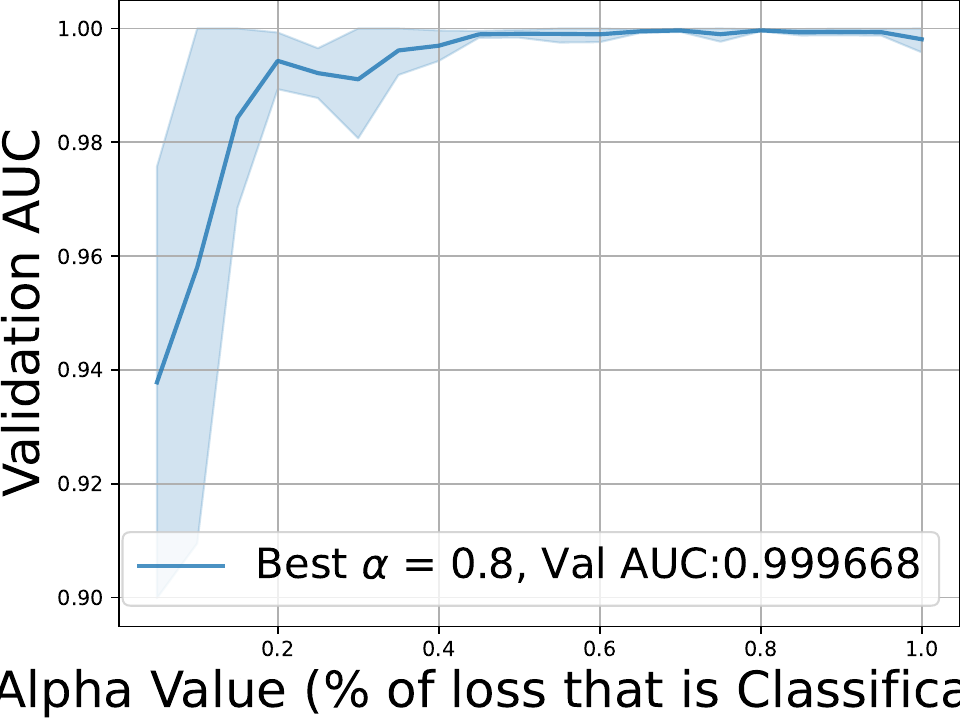}
          \caption{SSIM\&L1}
      \end{subfigure}
      \hfill
      \begin{subfigure}[b]{0.19\textwidth}
          \centering
          \includegraphics[width=1\columnwidth]{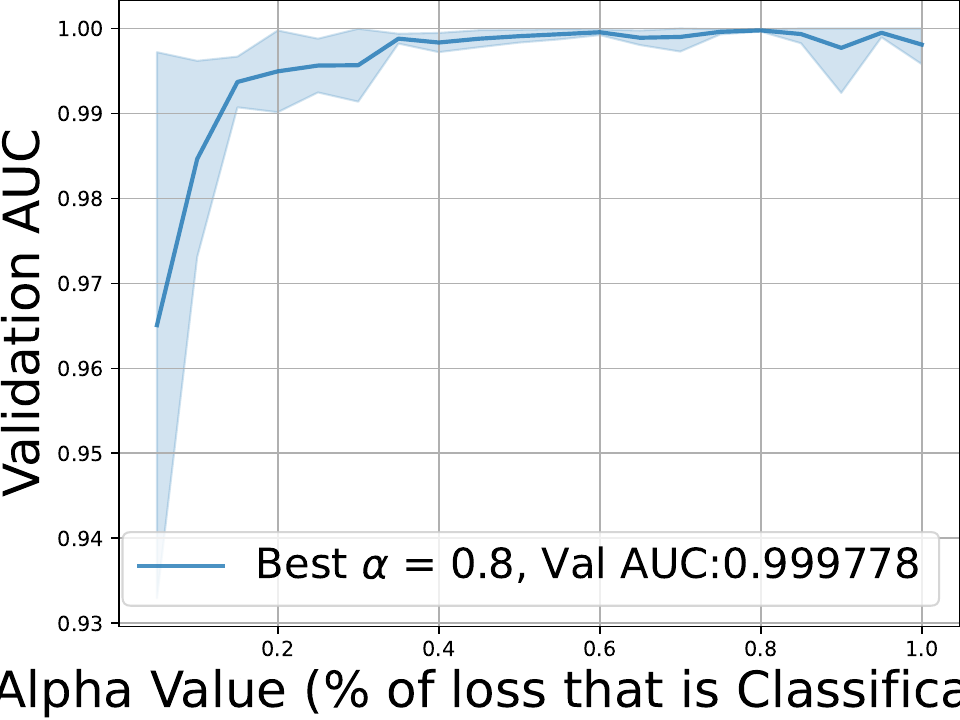}
          \caption{SSIM\&MSE}
      \end{subfigure}
      \caption{DenseNet}
  \end{subfigure} 
  \begin{subfigure}[b]{1\textwidth}
      \begin{subfigure}[b]{0.19\textwidth}
          \centering
          \includegraphics[width=1\columnwidth]{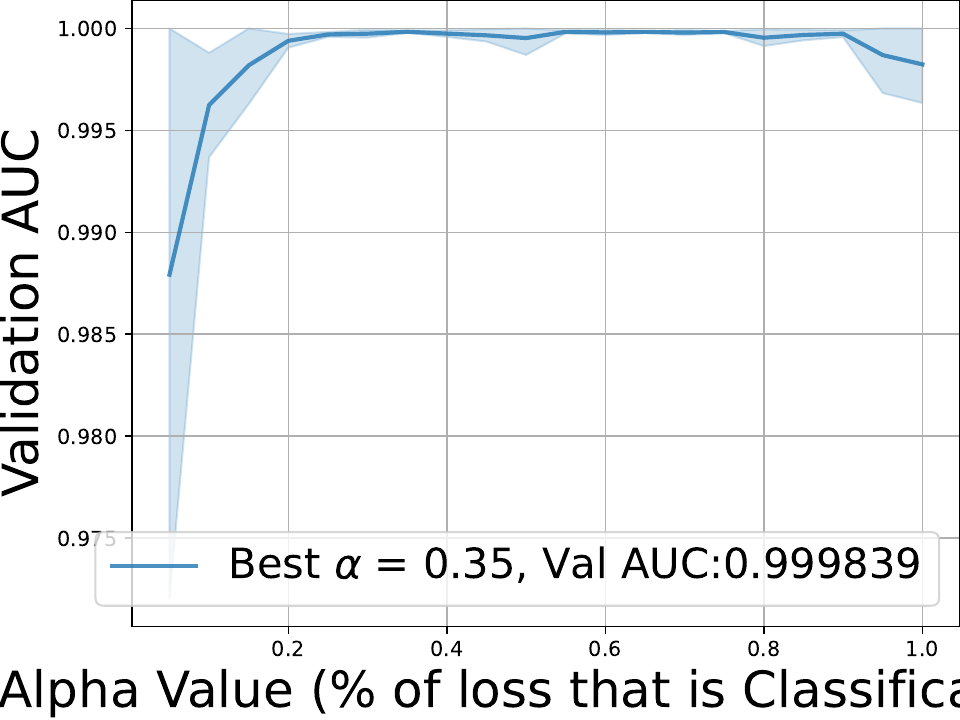}
          \caption{L1}
      \end{subfigure}
      \hfill
      \begin{subfigure}[b]{0.19\textwidth}
          \centering
          \includegraphics[width=1\columnwidth]{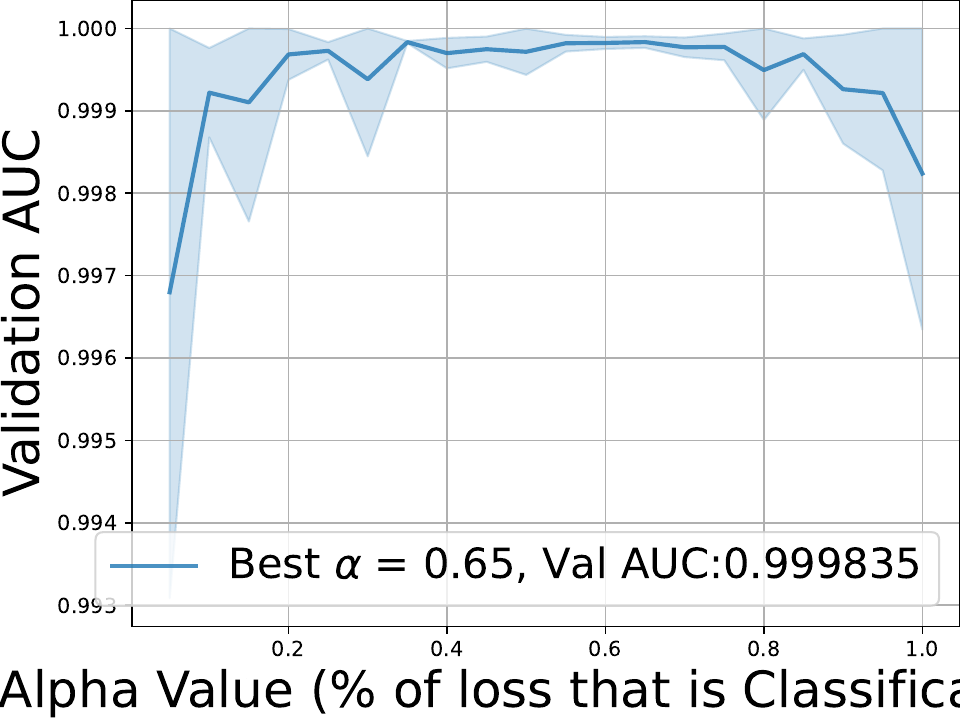}
          \caption{MSE}
      \end{subfigure}
      \hfill
      \begin{subfigure}[b]{0.19\textwidth}
          \centering
          \includegraphics[width=1\columnwidth]{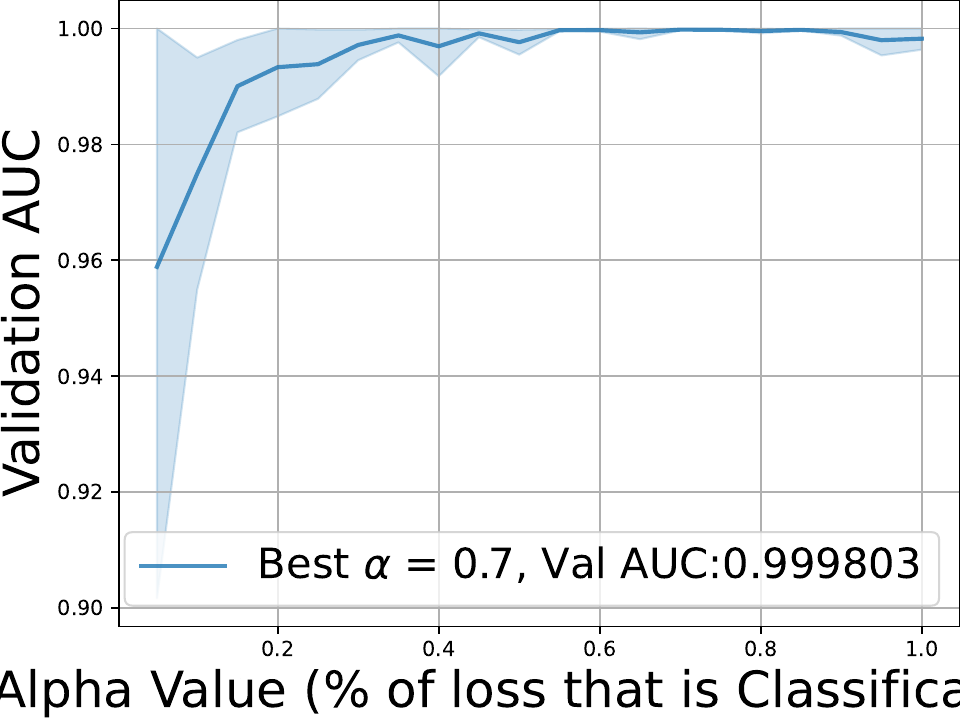}
          \caption{SSIM}
      \end{subfigure}
      \hfill
      \begin{subfigure}[b]{0.19\textwidth}
          \centering
          \includegraphics[width=1\columnwidth]{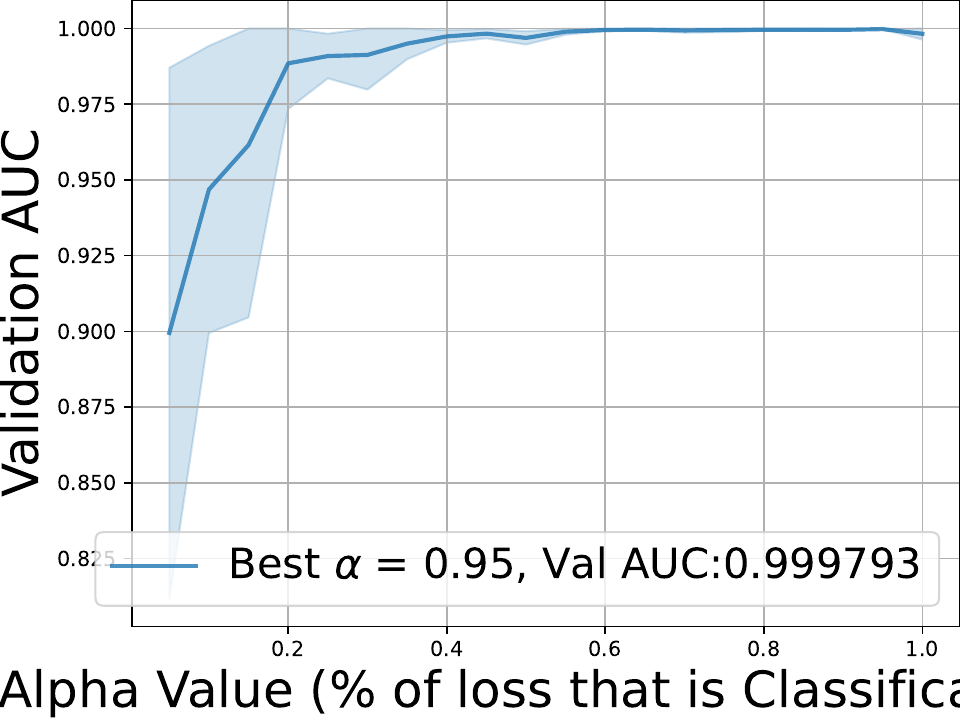}
          \caption{SSIM\&L1}
      \end{subfigure}
      \hfill
      \begin{subfigure}[b]{0.19\textwidth}
          \centering
          \includegraphics[width=1\columnwidth]{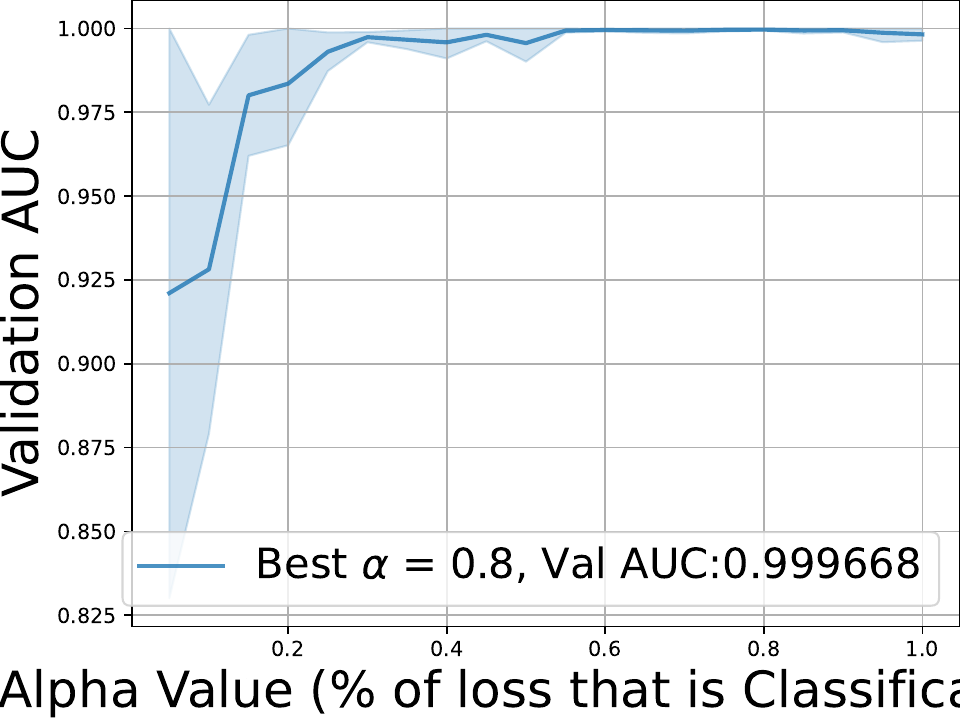}
          \caption{SSIM\&MSE}
      \end{subfigure}
      \caption{Resnet}
  \end{subfigure} 
\begin{subfigure}[b]{1\textwidth}
      \begin{subfigure}[b]{0.19\textwidth}
          \centering
          \includegraphics[width=1\columnwidth]{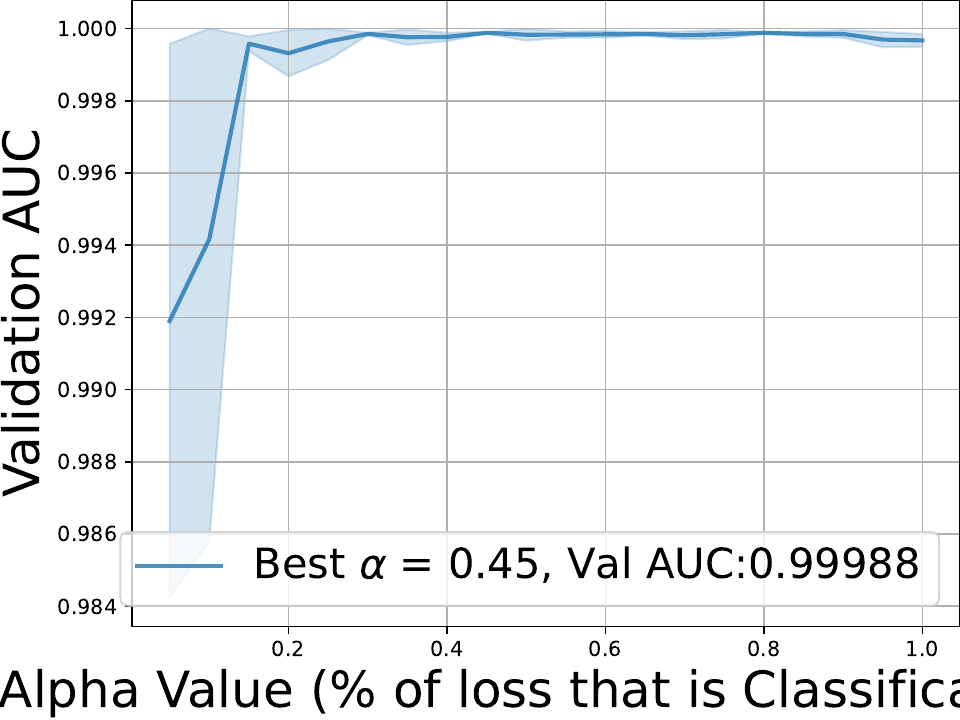}
          \caption{L1}
      \end{subfigure}
      \hfill
      \begin{subfigure}[b]{0.19\textwidth}
          \centering
          \includegraphics[width=1\columnwidth]{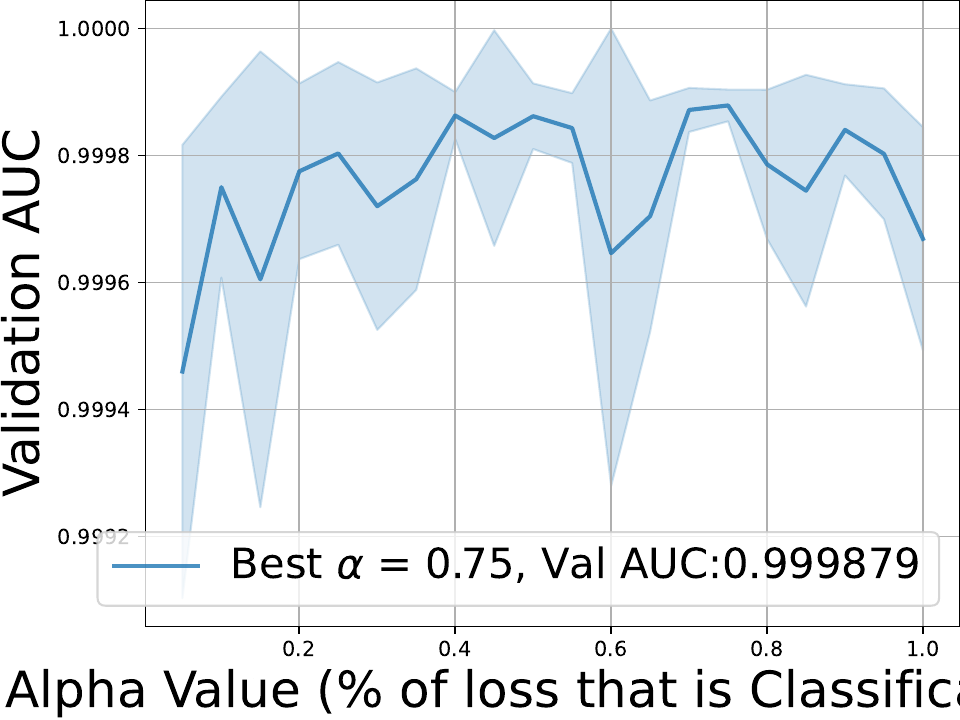}
          \caption{MSE}
      \end{subfigure}
      \hfill
      \begin{subfigure}[b]{0.19\textwidth}
          \centering
          \includegraphics[width=1\columnwidth]{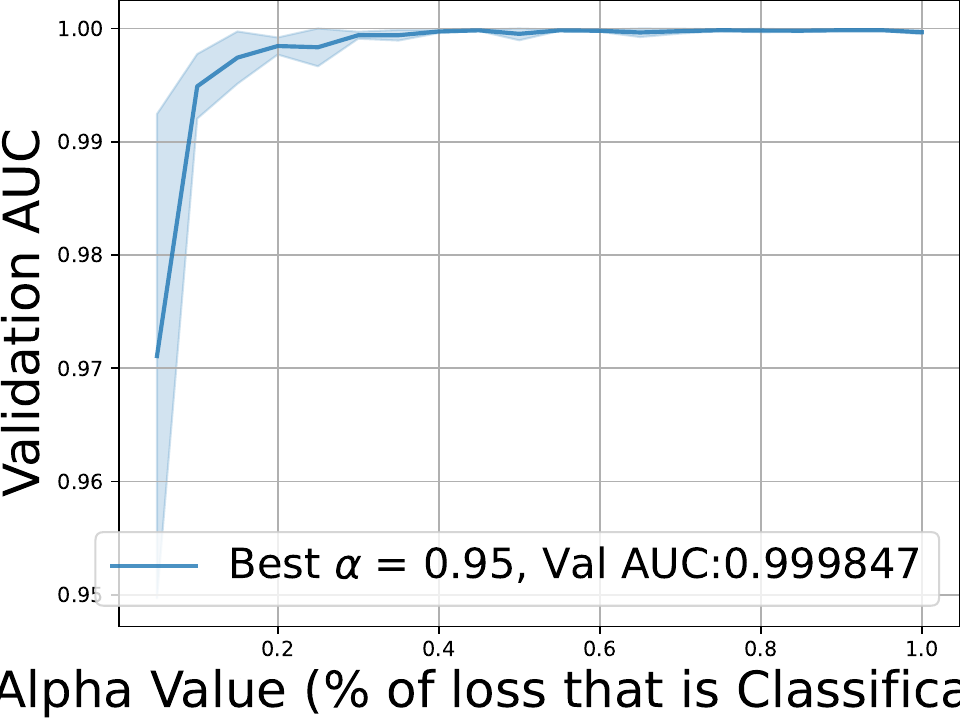}
          \caption{SSIM}
      \end{subfigure}
      \hfill
      \begin{subfigure}[b]{0.19\textwidth}
          \centering
          \includegraphics[width=1\columnwidth]{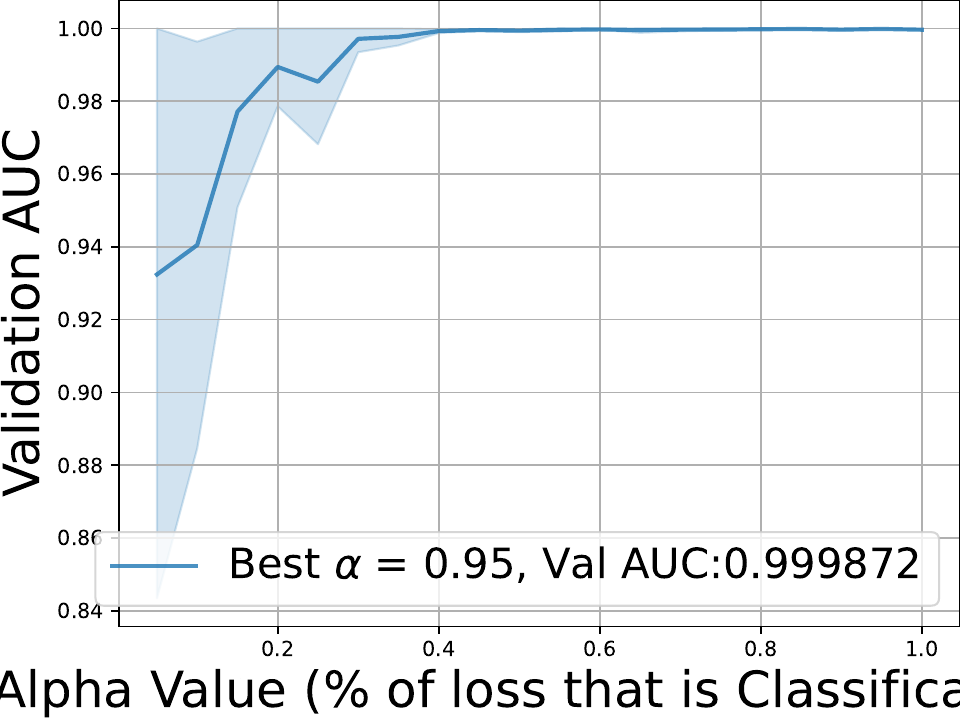}
          \caption{SSIM\&L1}
      \end{subfigure}
      \hfill
      \begin{subfigure}[b]{0.19\textwidth}
          \centering
          \includegraphics[width=1\columnwidth]{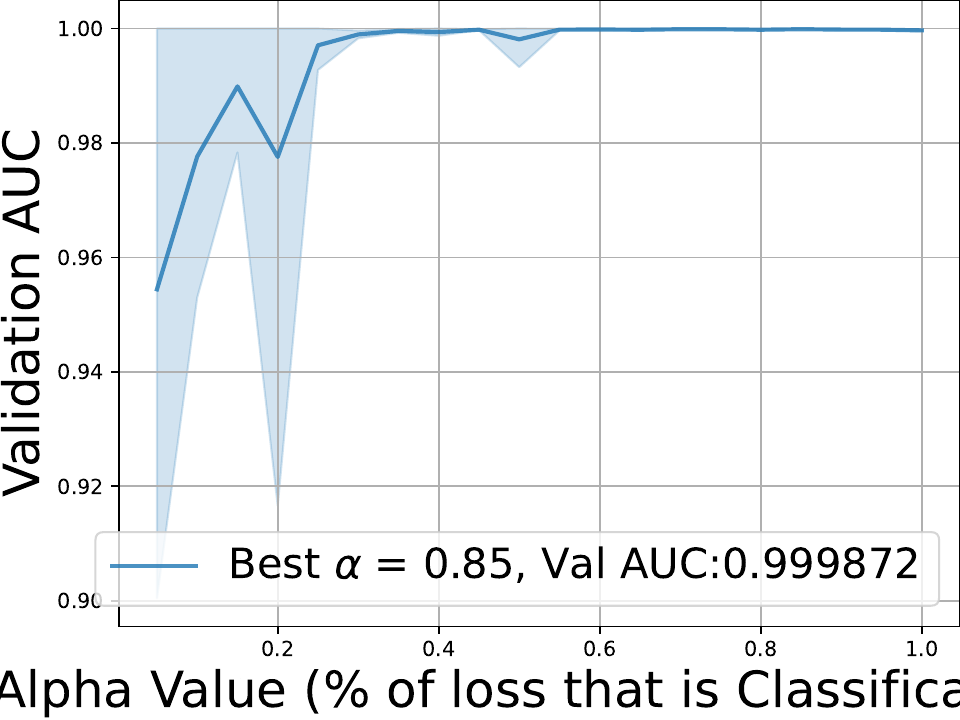}
          \caption{SSIM\&MSE}
      \end{subfigure}
      \caption{Inception}
  \end{subfigure} 
  \caption{Synthetic Face.}
  \label{fig:face-ablation}
  \null\vskip-5mm
\end{figure*}

\begin{figure*}[t]
  \begin{subfigure}[b]{1\textwidth}
      \begin{subfigure}[b]{0.19\textwidth}
          \centering
          \includegraphics[width=1\columnwidth]{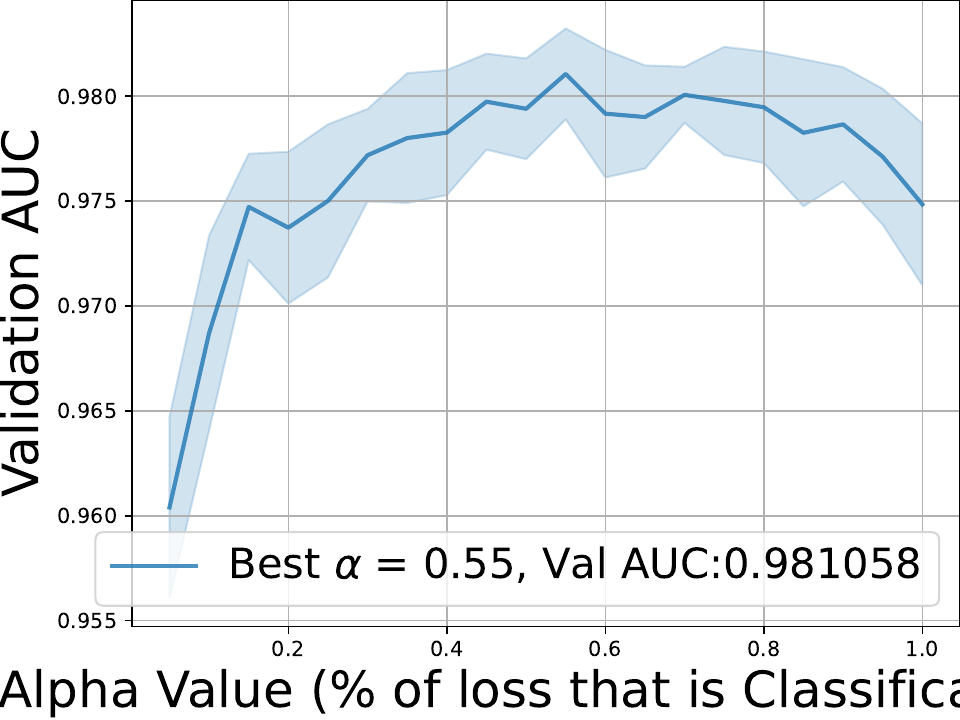}
          \caption{L1}
      \end{subfigure}
      \hfill
      \begin{subfigure}[b]{0.19\textwidth}
          \centering
          \includegraphics[width=1\columnwidth]{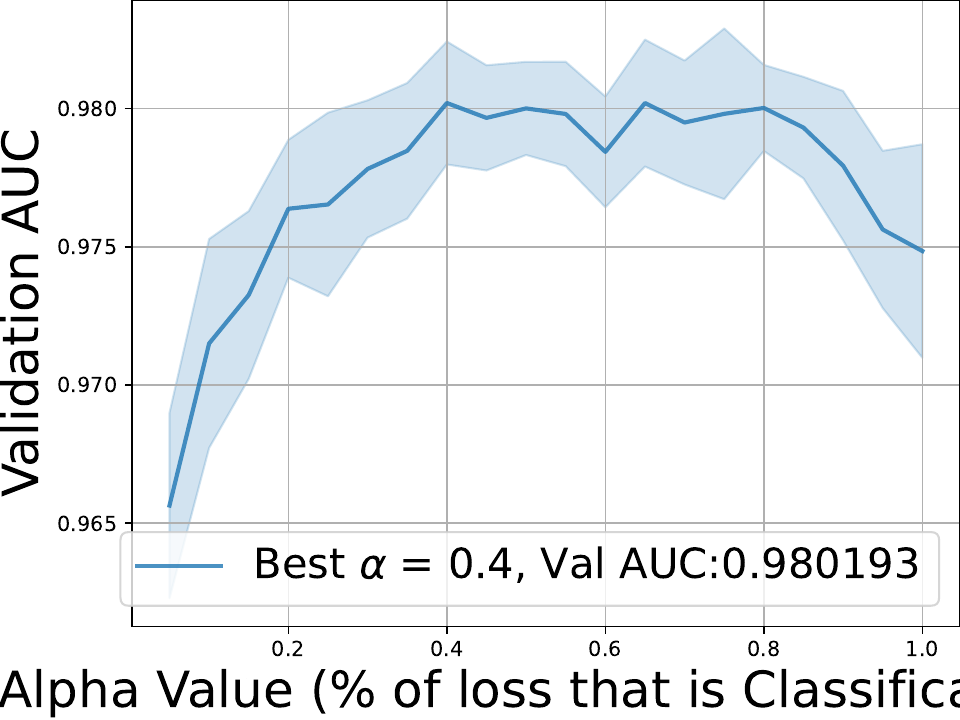}
          \caption{MSE}
      \end{subfigure}
      \hfill
      \begin{subfigure}[b]{0.19\textwidth}
          \centering
          \includegraphics[width=1\columnwidth]{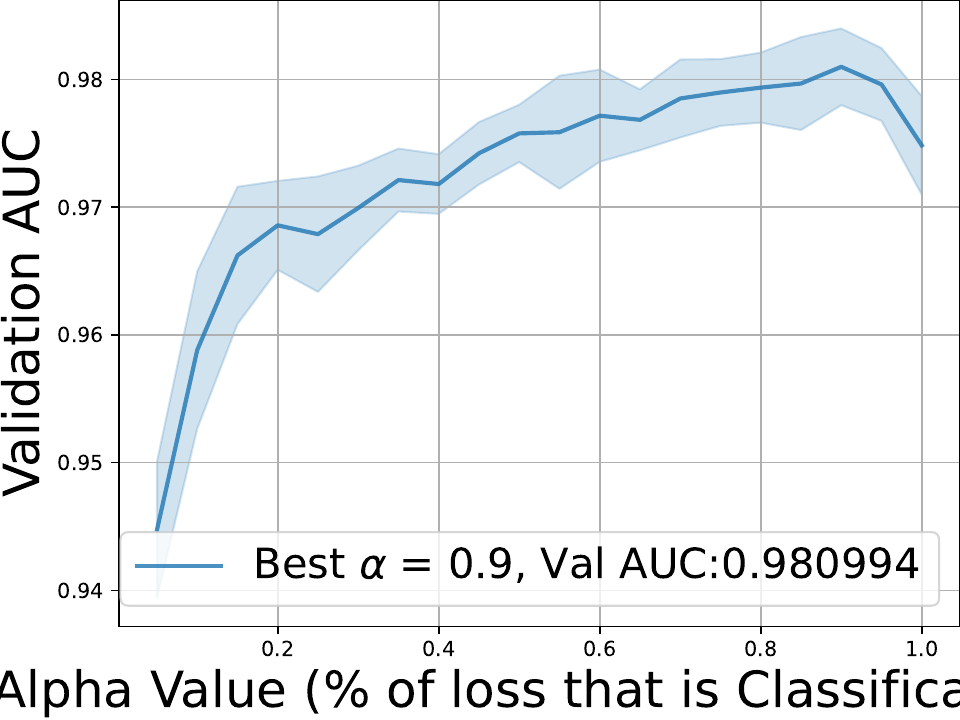}
          \caption{SSIM}
      \end{subfigure}
      \hfill
      \begin{subfigure}[b]{0.19\textwidth}
          \centering
          \includegraphics[width=1\columnwidth]{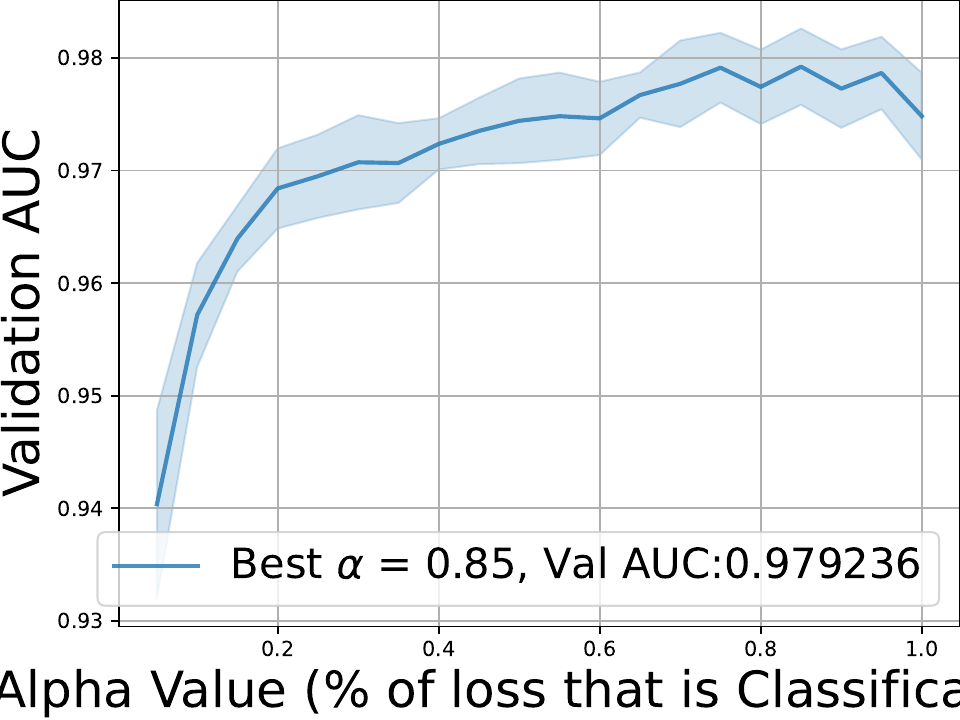}
          \caption{SSIM\&L1}
      \end{subfigure}
      \hfill
      \begin{subfigure}[b]{0.19\textwidth}
          \centering
          \includegraphics[width=1\columnwidth]{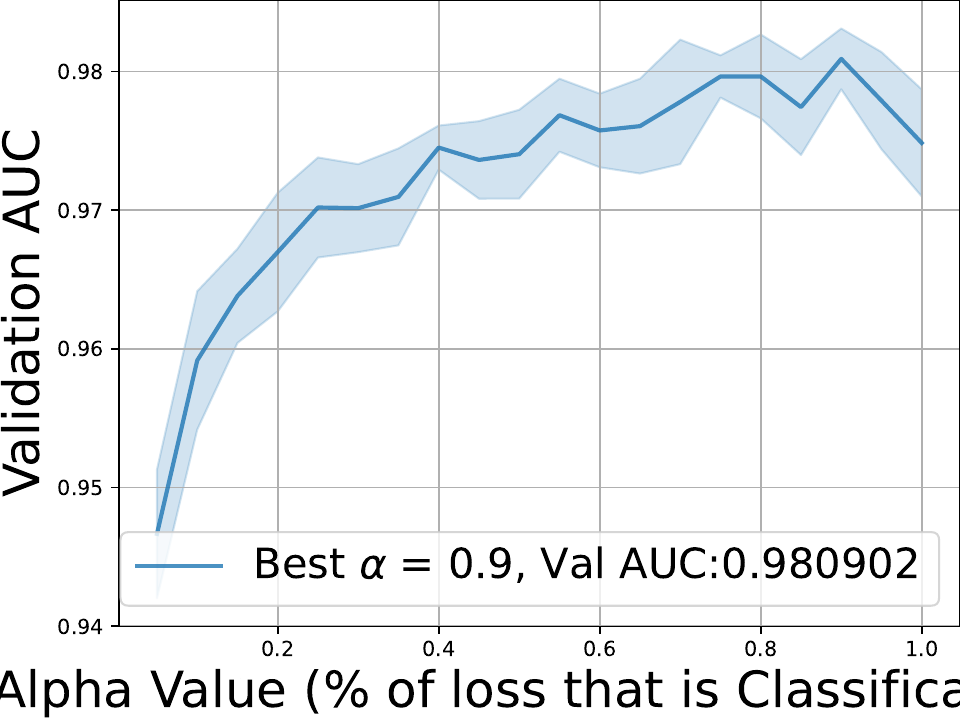}
          \caption{SSIM\&MSE}
      \end{subfigure}
      \caption{DenseNet}
  \end{subfigure} 
  \begin{subfigure}[b]{1\textwidth}
      \begin{subfigure}[b]{0.19\textwidth}
          \centering
          \includegraphics[width=1\columnwidth]{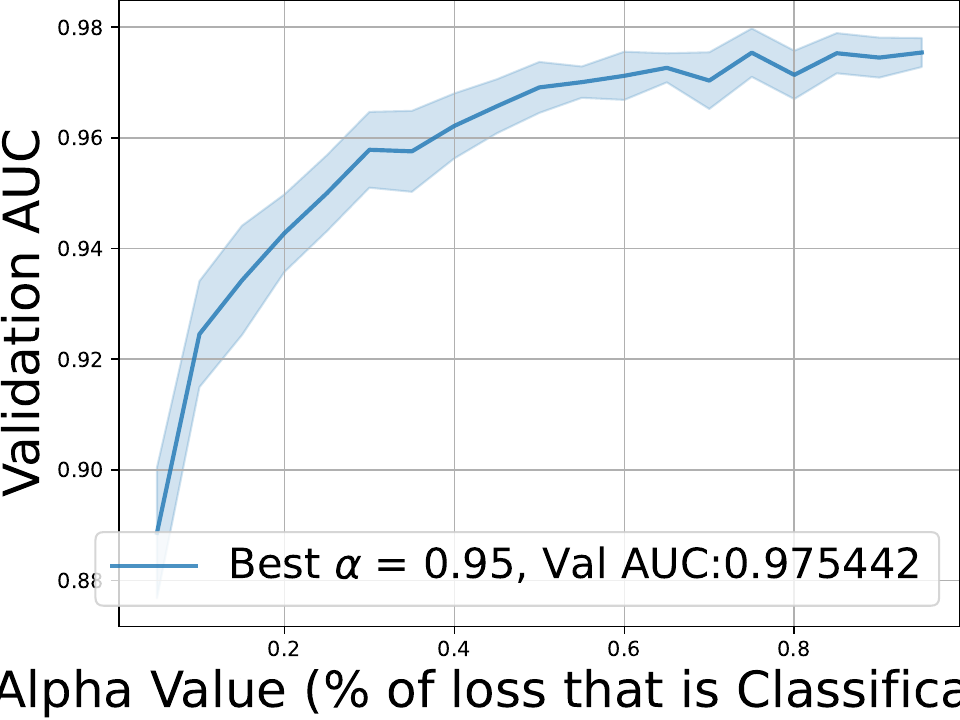}
          \caption{L1}
      \end{subfigure}
      \hfill
      \begin{subfigure}[b]{0.19\textwidth}
          \centering
          \includegraphics[width=1\columnwidth]{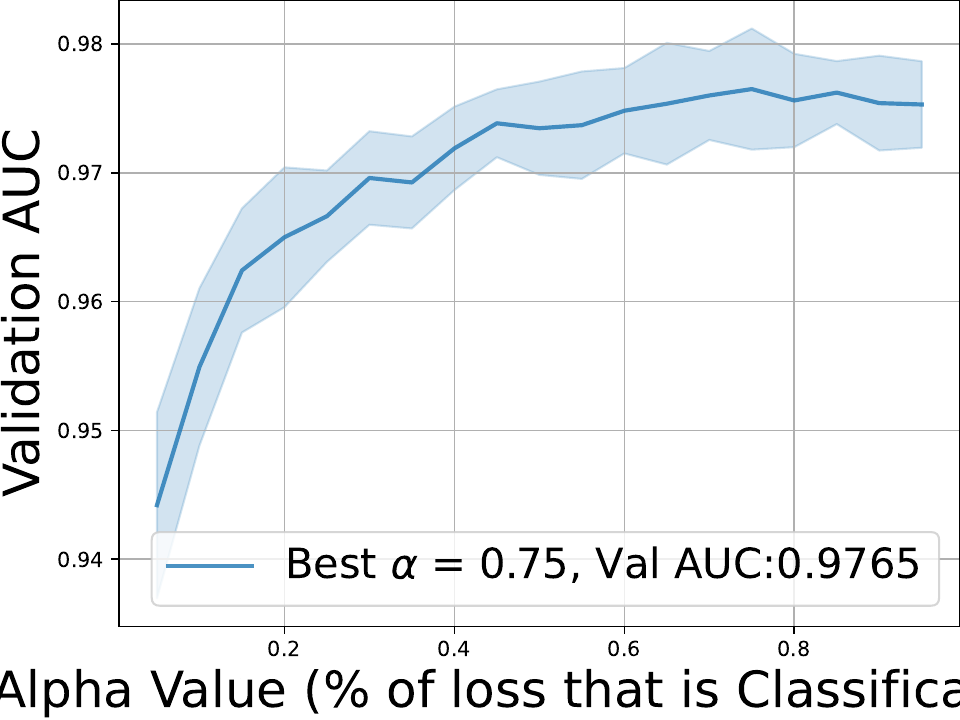}
          \caption{MSE}
      \end{subfigure}
      \hfill
      \begin{subfigure}[b]{0.19\textwidth}
          \centering
          \includegraphics[width=1\columnwidth]{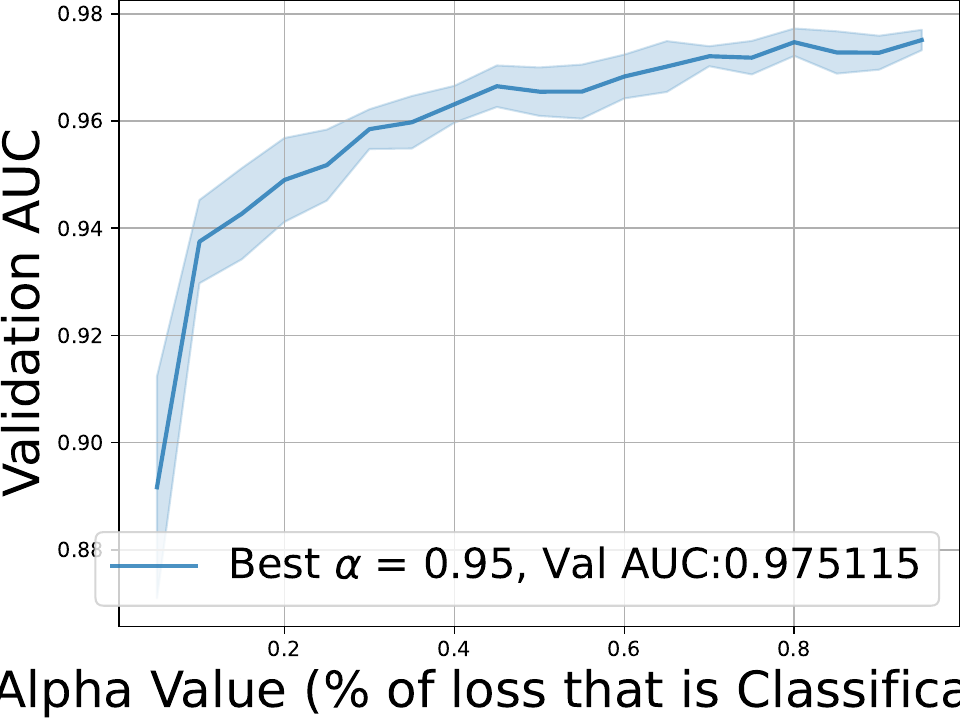}
          \caption{SSIM}
      \end{subfigure}
      \hfill
      \begin{subfigure}[b]{0.19\textwidth}
          \centering
          \includegraphics[width=1\columnwidth]{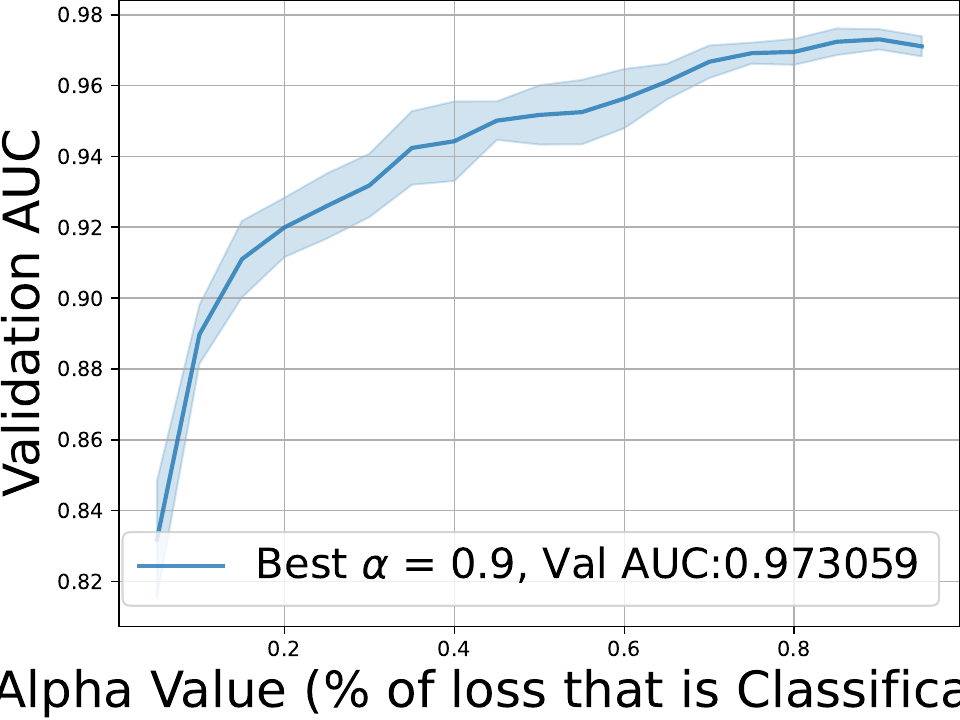}
          \caption{SSIM\&L1}
      \end{subfigure}
      \hfill
      \begin{subfigure}[b]{0.19\textwidth}
          \centering
          \includegraphics[width=1\columnwidth]{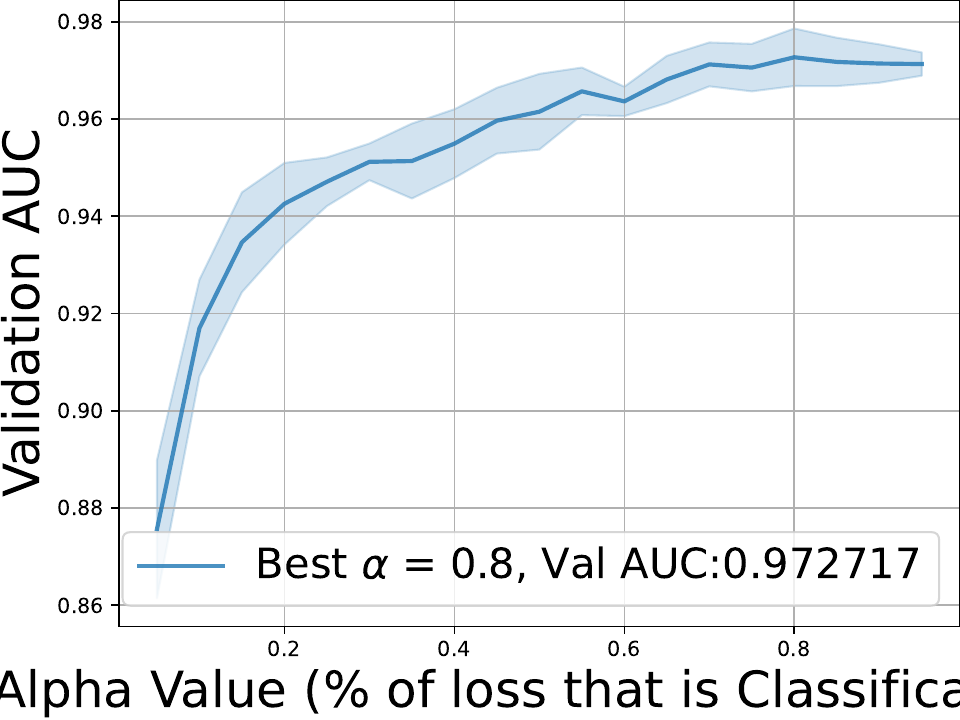}
          \caption{SSIM\&MSE}
      \end{subfigure}
      \caption{Resnet}
  \end{subfigure} 
\begin{subfigure}[b]{1\textwidth}
      \begin{subfigure}[b]{0.19\textwidth}
          \centering
          \includegraphics[width=1\columnwidth]{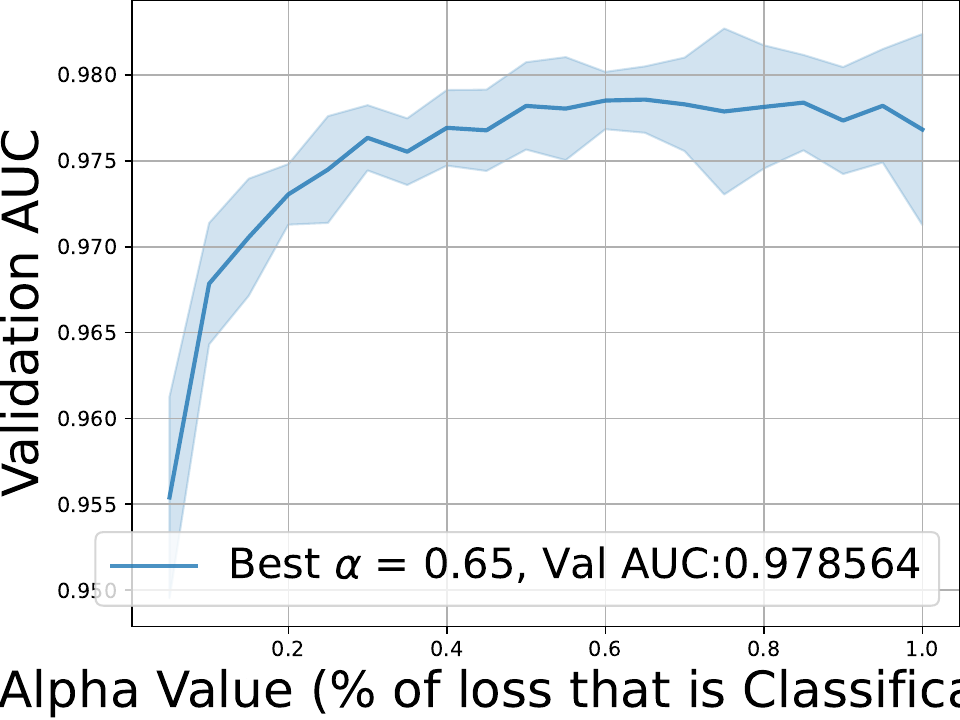}
          \caption{L1}
      \end{subfigure}
      \hfill
      \begin{subfigure}[b]{0.19\textwidth}
          \centering
          \includegraphics[width=1\columnwidth]{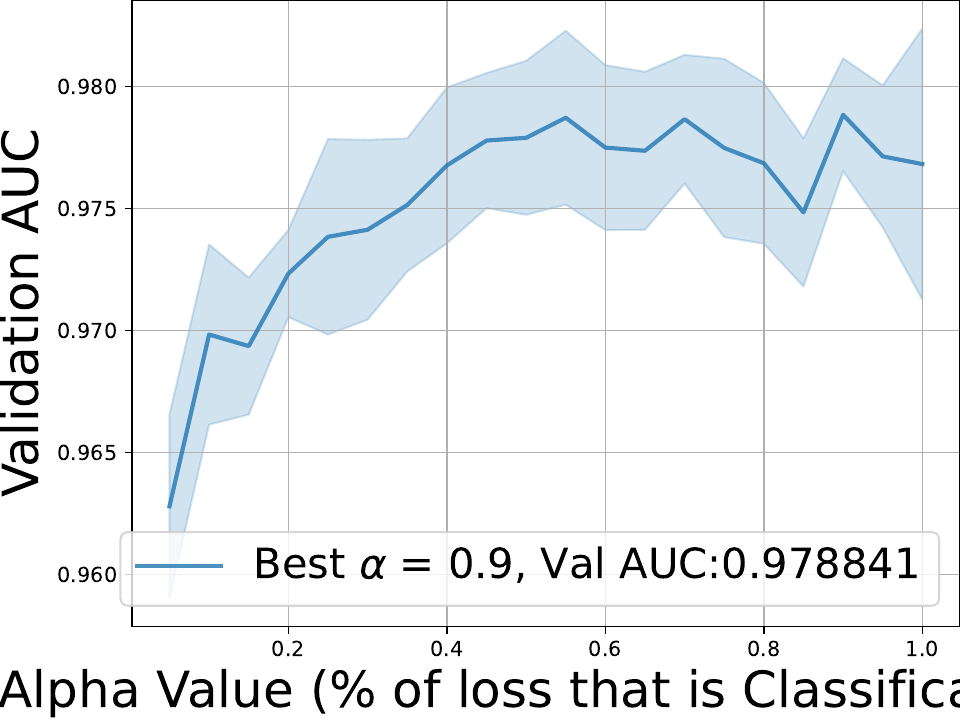}
          \caption{MSE}
      \end{subfigure}
      \hfill
      \begin{subfigure}[b]{0.19\textwidth}
          \centering
          \includegraphics[width=1\columnwidth]{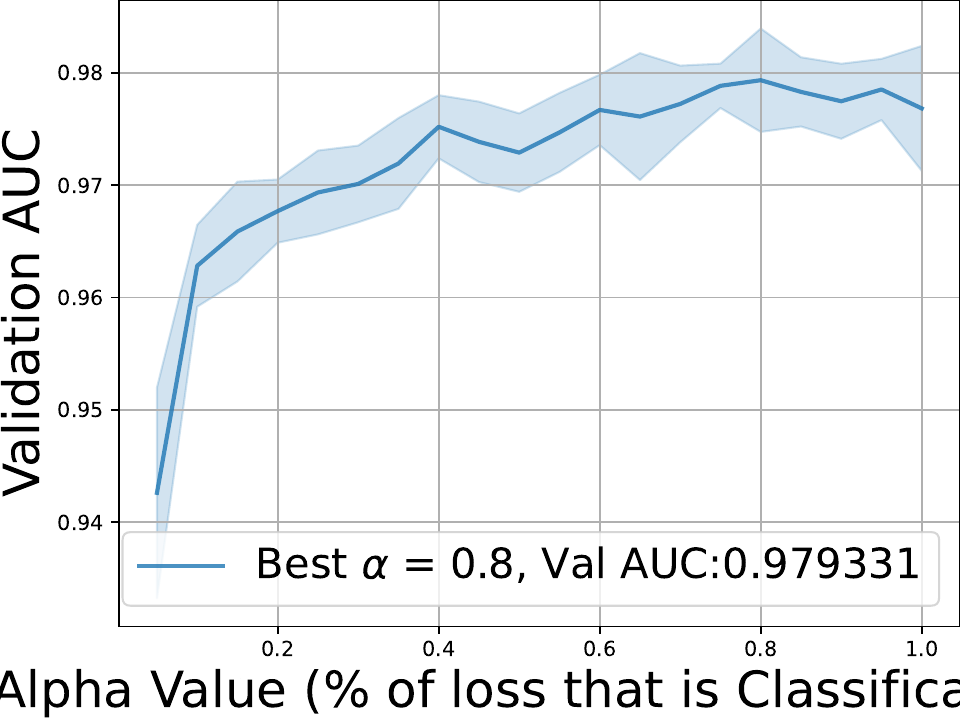}
          \caption{SSIM}
      \end{subfigure}
      \hfill
      \begin{subfigure}[b]{0.19\textwidth}
          \centering
          \includegraphics[width=1\columnwidth]{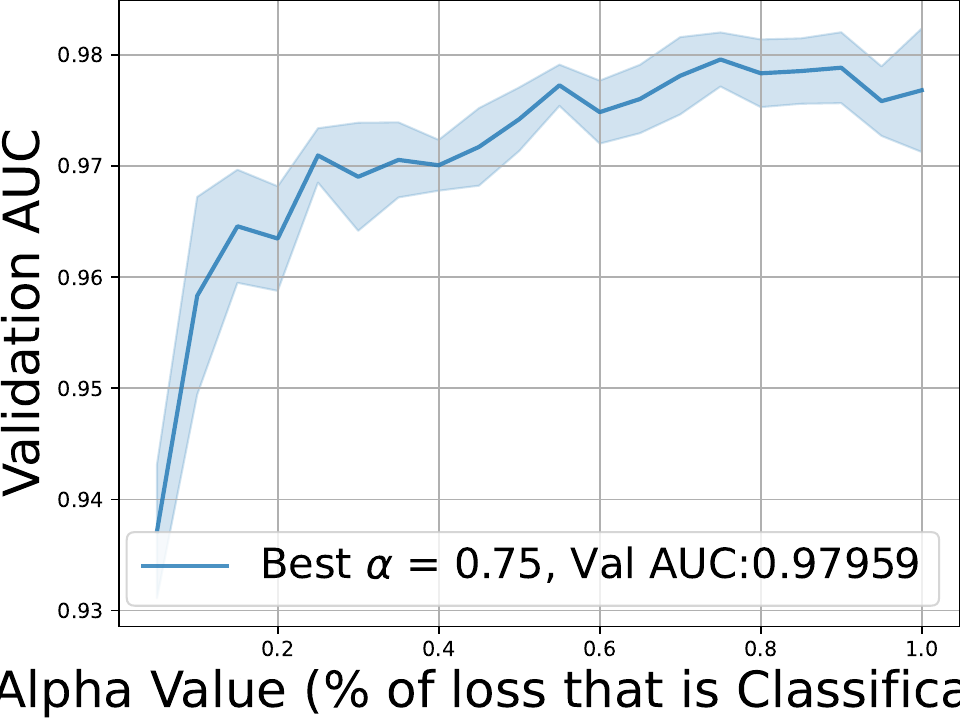}
          \caption{SSIM\&L1}
      \end{subfigure}
      \hfill
      \begin{subfigure}[b]{0.19\textwidth}
          \centering
          \includegraphics[width=1\columnwidth]{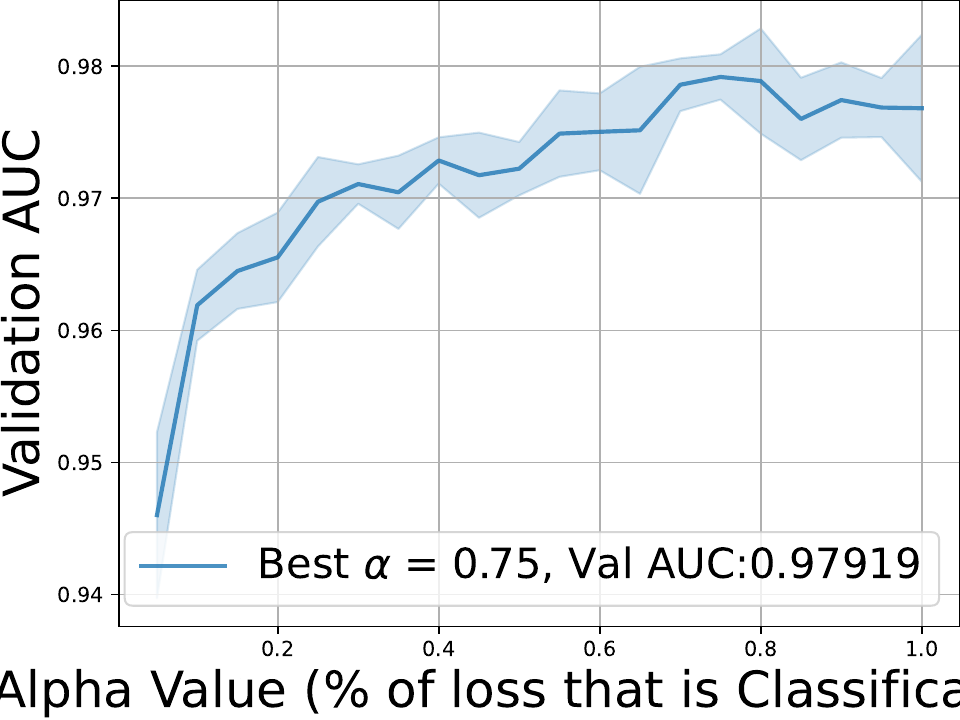}
          \caption{SSIM\&MSE}
      \end{subfigure}
      \caption{Inception}
  \end{subfigure} 
  \caption{Iris.}
  \label{fig:iris-ablation}
  \null\vskip-5mm
\end{figure*}

\begin{figure*}[t]
  \begin{subfigure}[b]{1\textwidth}
      \begin{subfigure}[b]{0.19\textwidth}
          \centering
          \includegraphics[width=1\columnwidth]{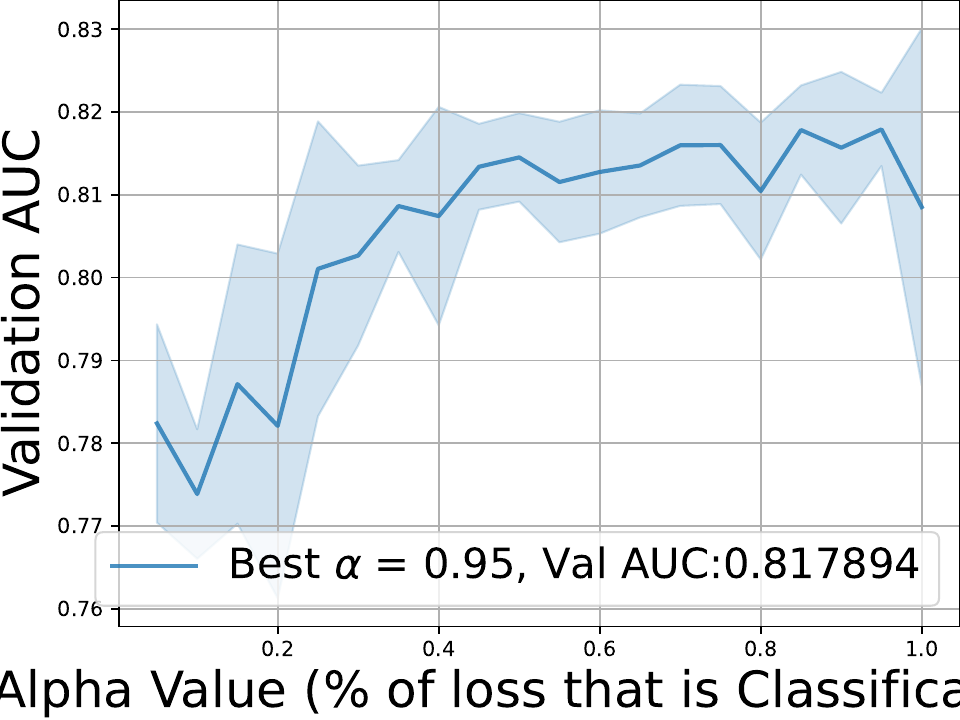}
          \caption{L1}
      \end{subfigure}
      \hfill
      \begin{subfigure}[b]{0.19\textwidth}
          \centering
          \includegraphics[width=1\columnwidth]{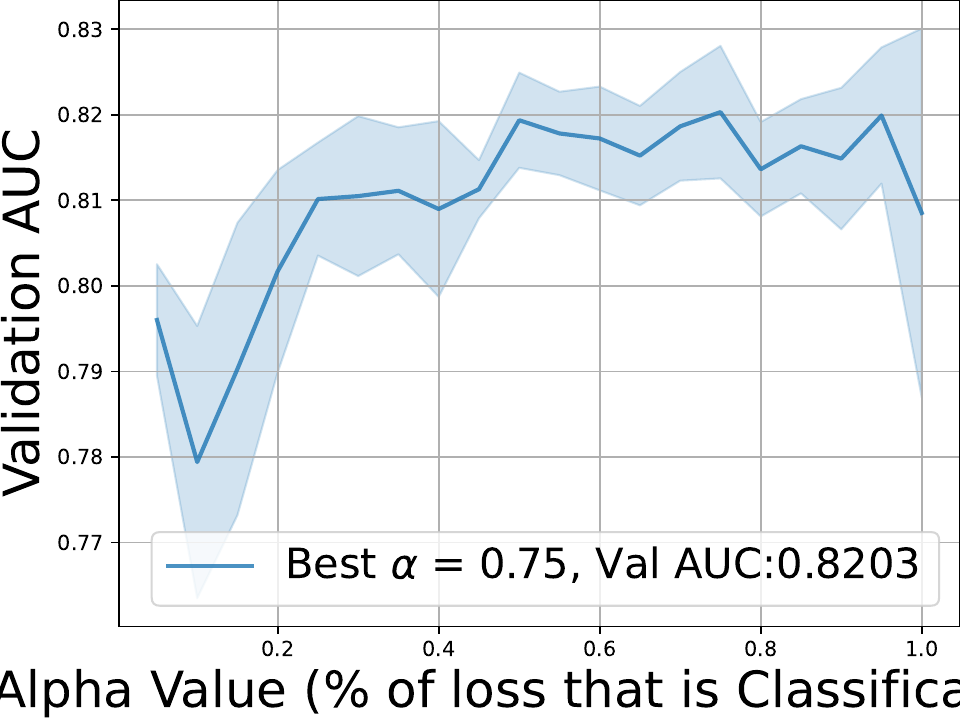}
          \caption{MSE}
      \end{subfigure}
      \hfill
      \begin{subfigure}[b]{0.19\textwidth}
          \centering
          \includegraphics[width=1\columnwidth]{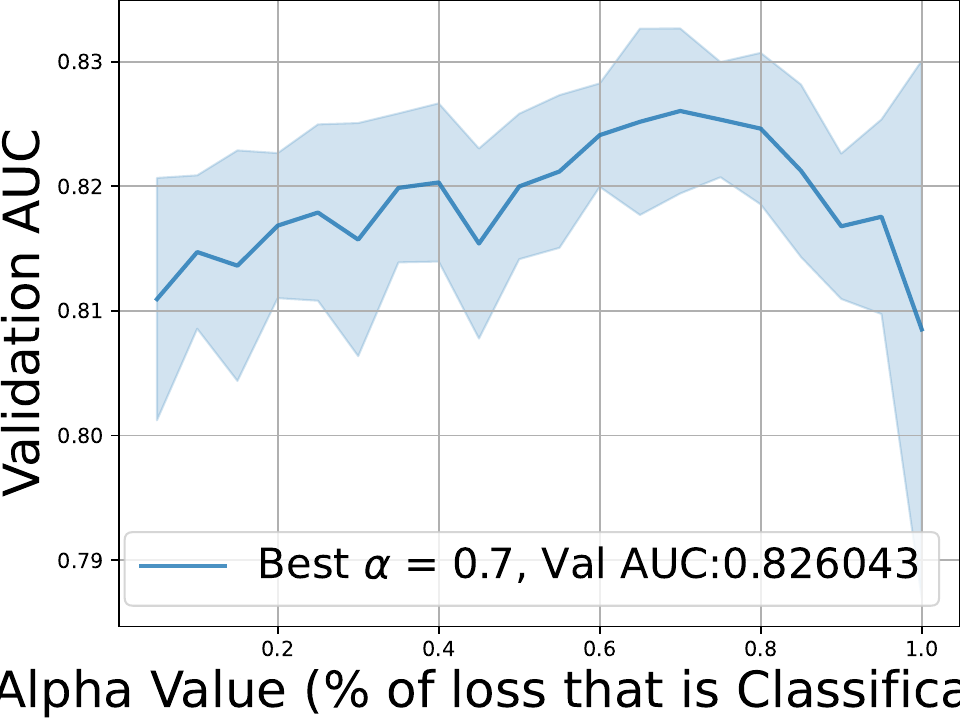}
          \caption{SSIM}
      \end{subfigure}
      \hfill
      \begin{subfigure}[b]{0.19\textwidth}
          \centering
          \includegraphics[width=1\columnwidth]{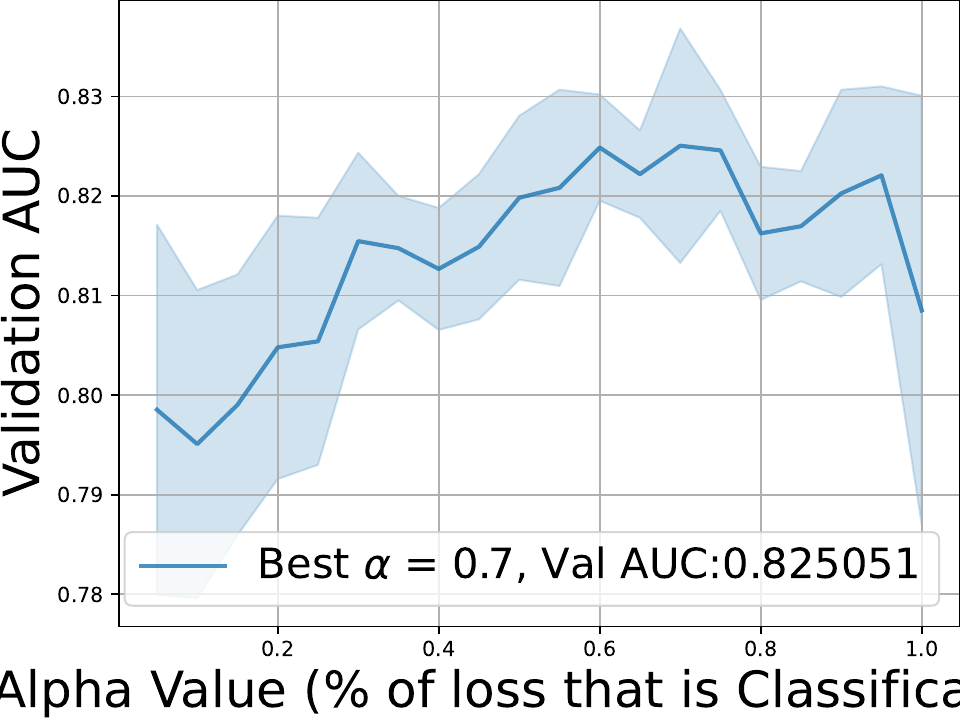}
          \caption{SSIM\&L1}
      \end{subfigure}
      \hfill
      \begin{subfigure}[b]{0.19\textwidth}
          \centering
          \includegraphics[width=1\columnwidth]{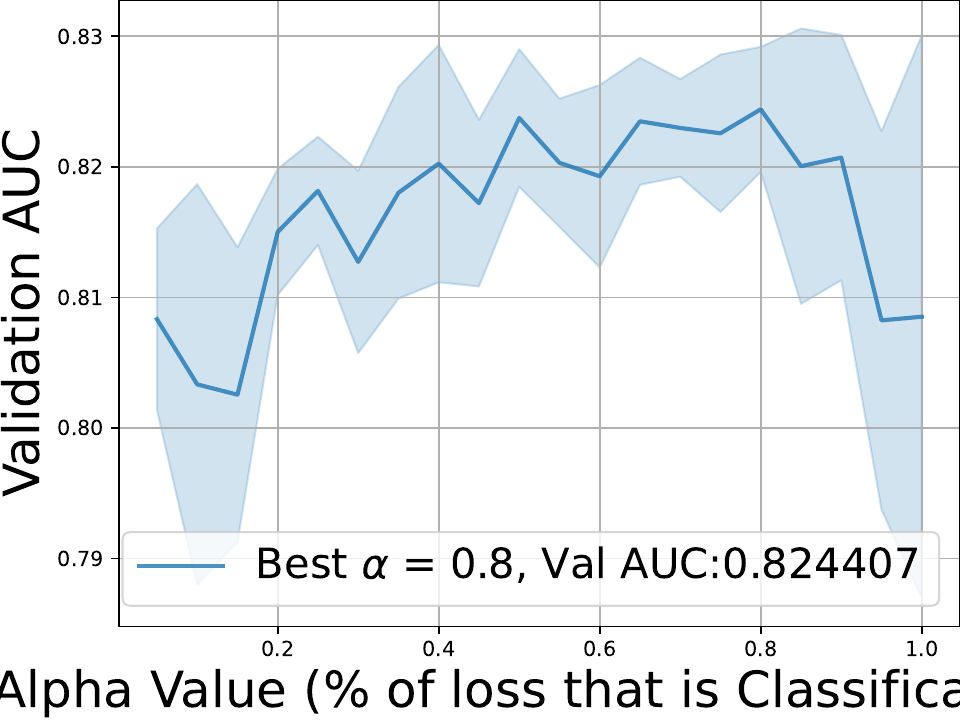}
          \caption{SSIM\&MSE}
      \end{subfigure}
      \caption{DenseNet}
  \end{subfigure} 
  \begin{subfigure}[b]{1\textwidth}
      \begin{subfigure}[b]{0.19\textwidth}
          \centering
          \includegraphics[width=1\columnwidth]{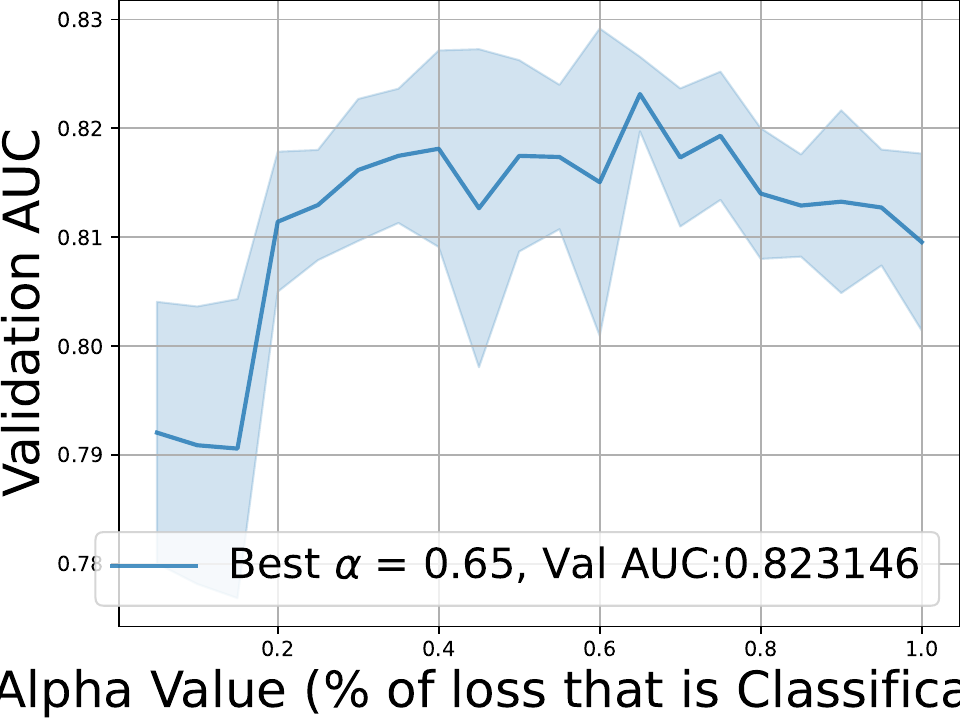}
          \caption{L1}
      \end{subfigure}
      \hfill
      \begin{subfigure}[b]{0.19\textwidth}
          \centering
          \includegraphics[width=1\columnwidth]{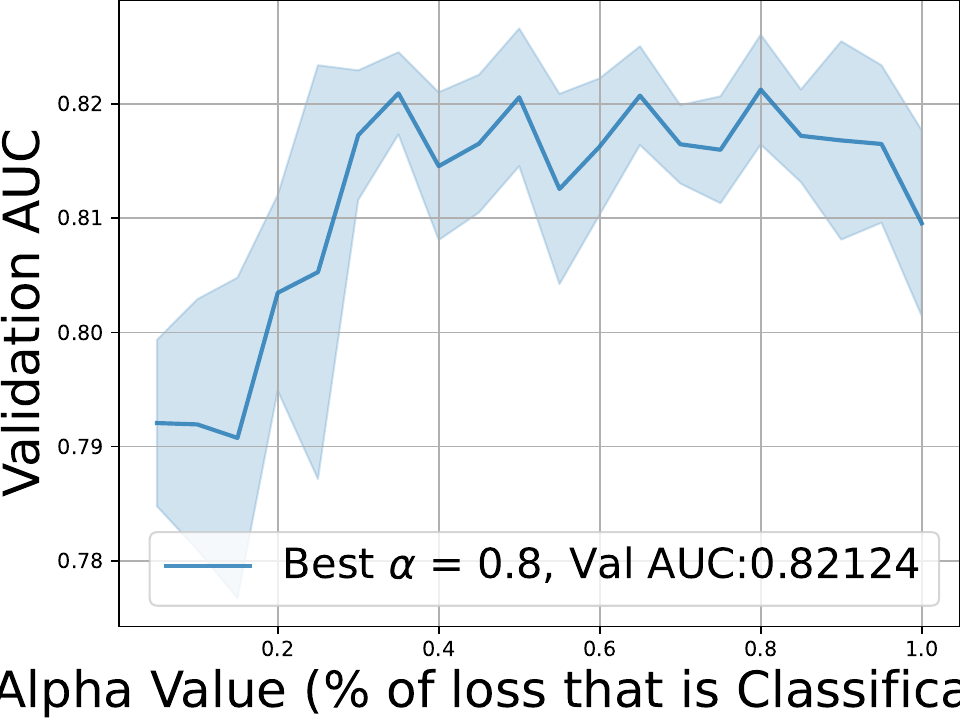}
          \caption{MSE}
      \end{subfigure}
      \hfill
      \begin{subfigure}[b]{0.19\textwidth}
          \centering
          \includegraphics[width=1\columnwidth]{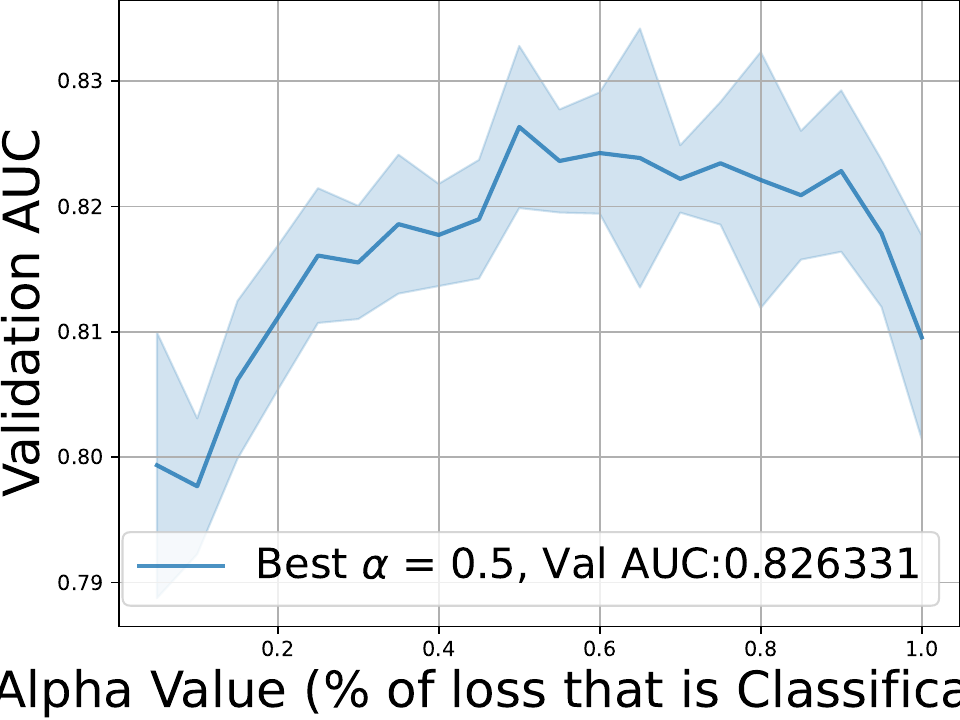}
          \caption{SSIM}
      \end{subfigure}
      \hfill
      \begin{subfigure}[b]{0.19\textwidth}
          \centering
          \includegraphics[width=1\columnwidth]{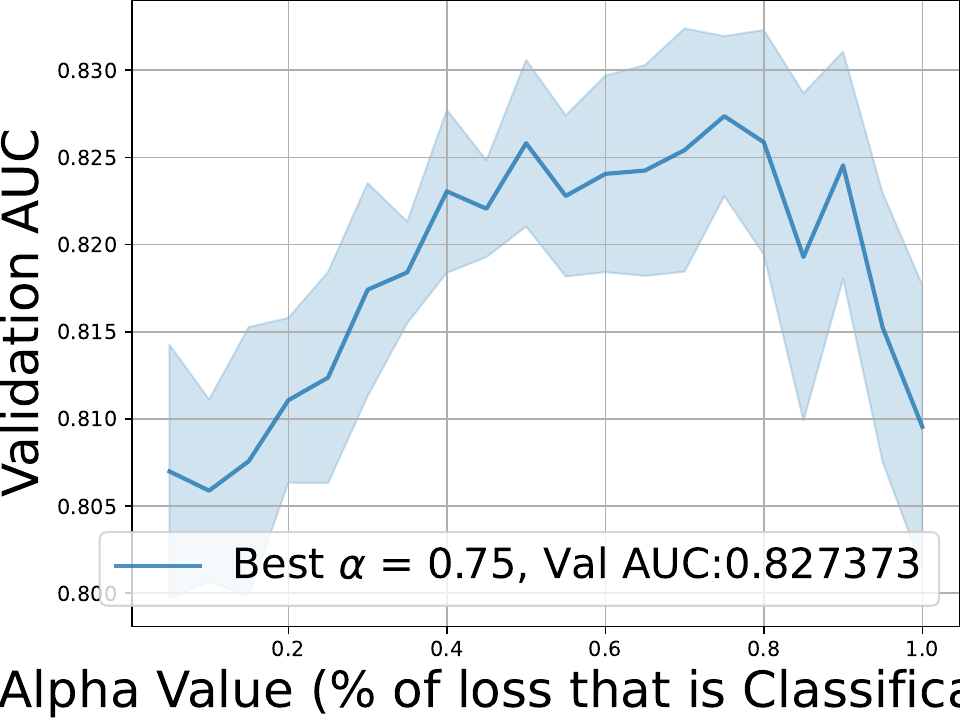}
          \caption{SSIM\&L1}
      \end{subfigure}
      \hfill
      \begin{subfigure}[b]{0.19\textwidth}
          \centering
          \includegraphics[width=1\columnwidth]{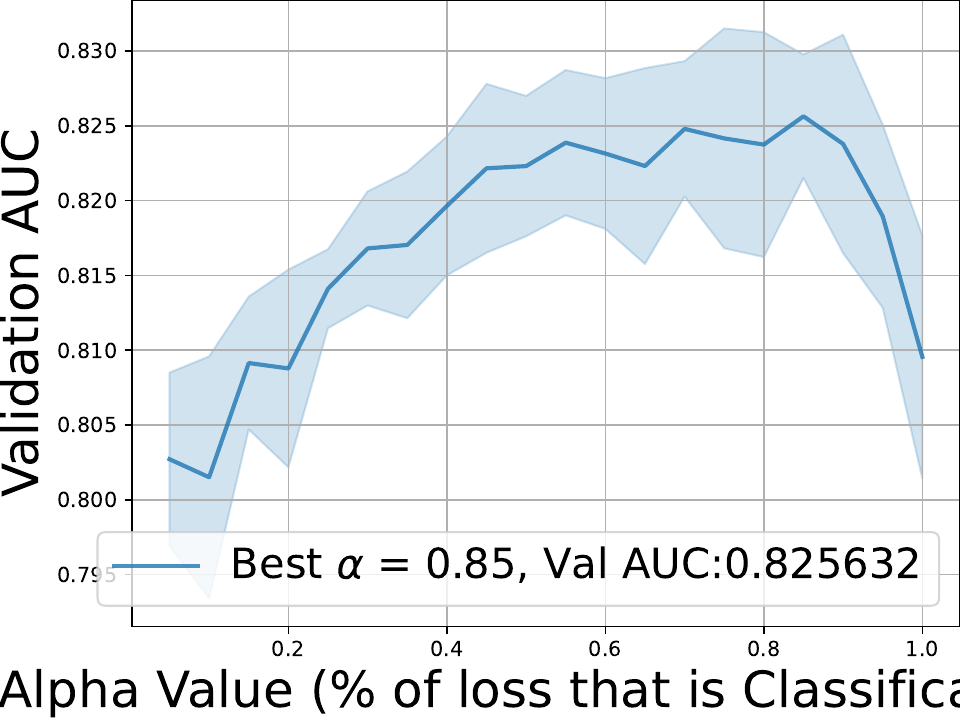}
          \caption{SSIM\&MSE}
      \end{subfigure}
      \caption{Resnet}
  \end{subfigure} 
\begin{subfigure}[b]{1\textwidth}
      \begin{subfigure}[b]{0.19\textwidth}
          \centering
          \includegraphics[width=1\columnwidth]{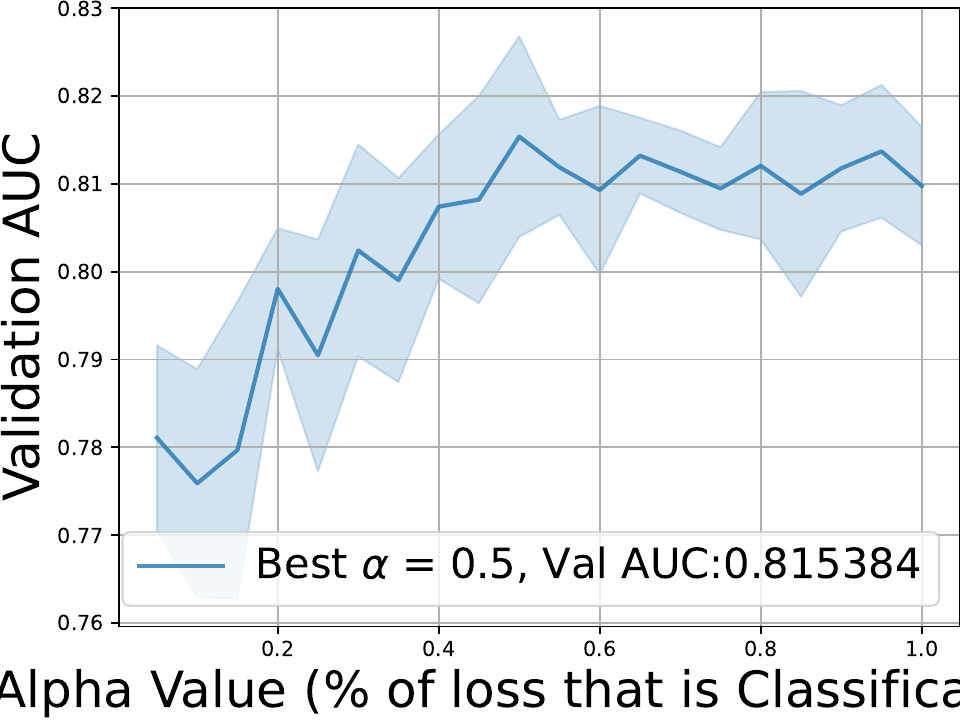}
          \caption{L1}
      \end{subfigure}
      \hfill
      \begin{subfigure}[b]{0.19\textwidth}
          \centering
          \includegraphics[width=1\columnwidth]{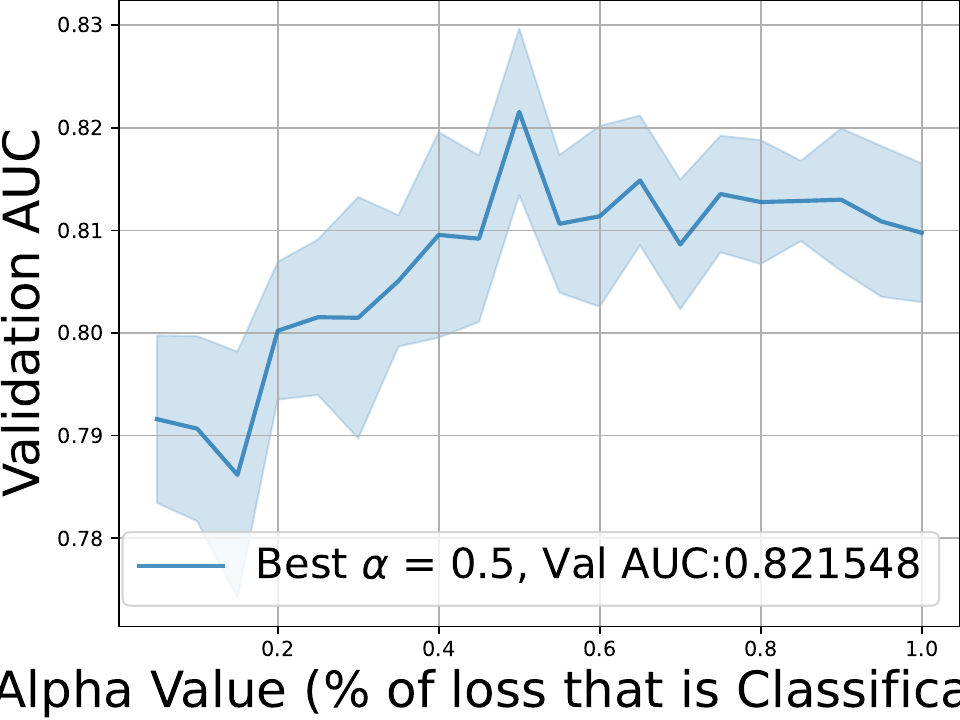}
          \caption{MSE}
      \end{subfigure}
      \hfill
      \begin{subfigure}[b]{0.19\textwidth}
          \centering
          \includegraphics[width=1\columnwidth]{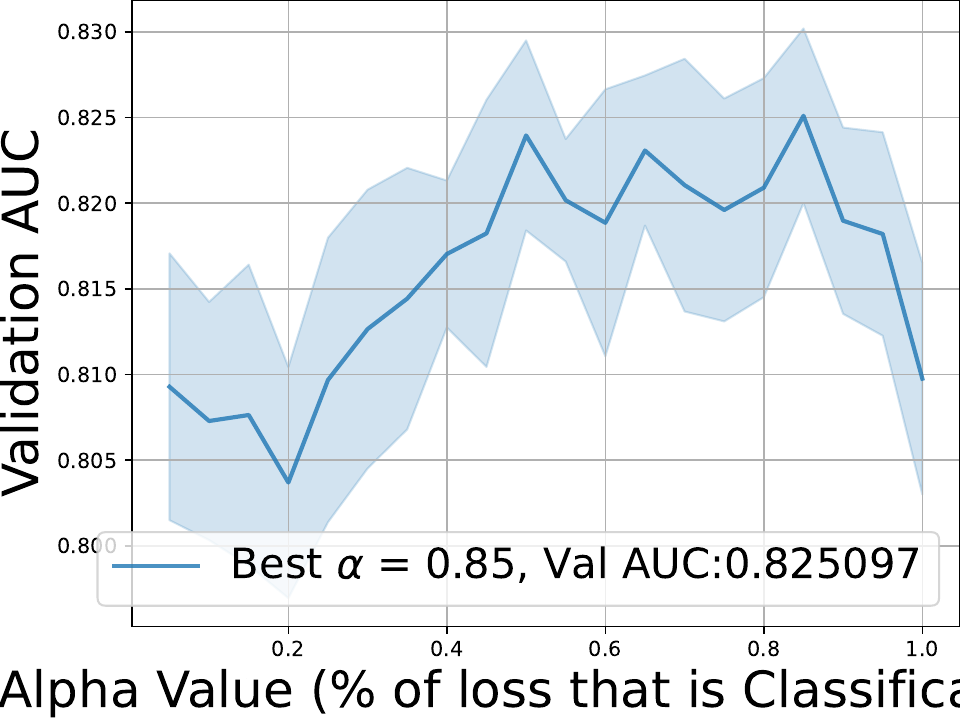}
          \caption{SSIM}
      \end{subfigure}
      \hfill
      \begin{subfigure}[b]{0.19\textwidth}
          \centering
          \includegraphics[width=1\columnwidth]{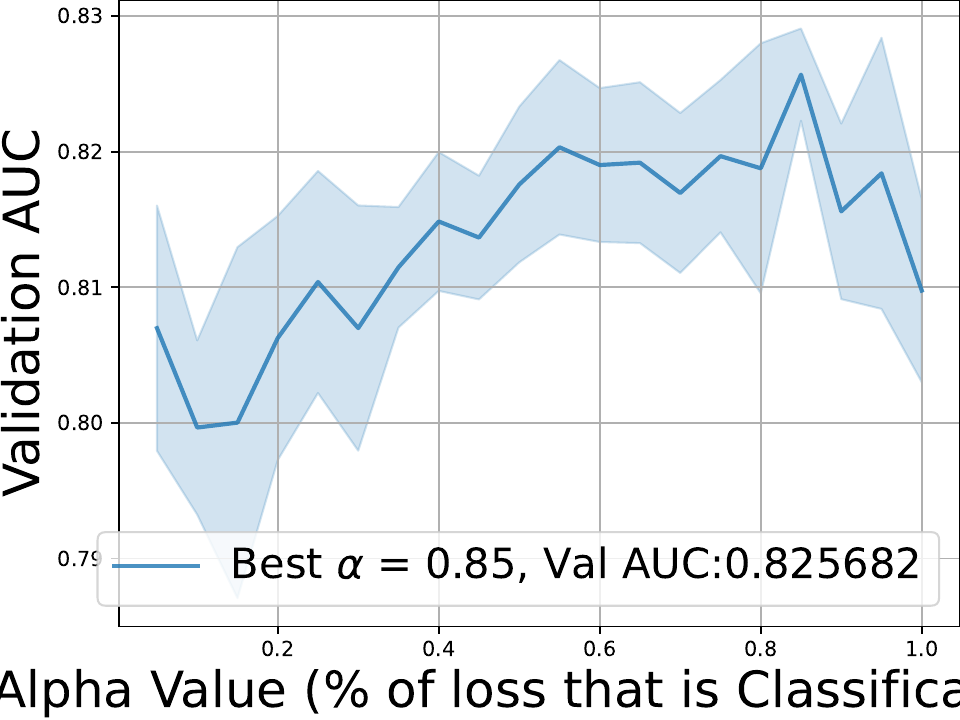}
          \caption{SSIM\&L1}
      \end{subfigure}
      \hfill
      \begin{subfigure}[b]{0.19\textwidth}
          \centering
          \includegraphics[width=1\columnwidth]{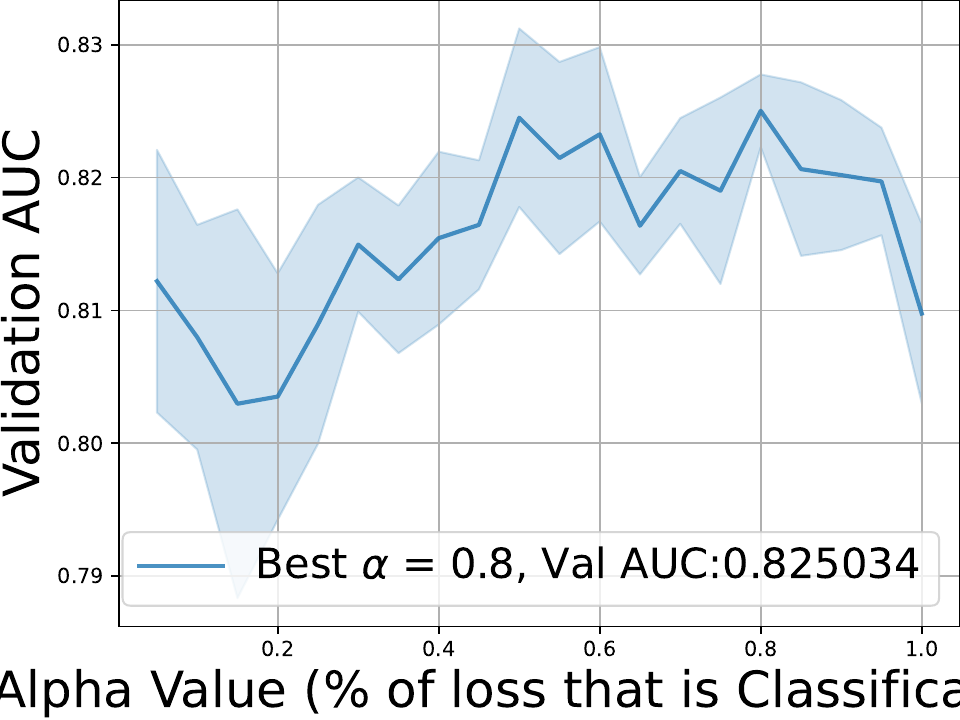}
          \caption{SSIM\&MSE}
      \end{subfigure}
      \caption{Inception}
  \end{subfigure} 
  \caption{CXR.}
  \label{fig:cxr-ablation}
  \null\vskip-5mm
\end{figure*}

This supplemental materials contain all plots used to determine the optimal combination of $\alpha$ and loss penalty for synthetic face detection (Fig. \ref{fig:face-ablation}), iris presentation attack detection (Fig. \ref{fig:iris-ablation}) and abnormality from chest x-ray (Fig. \ref{fig:cxr-ablation}). Plots show the AUC on the validation set at each alpha step (x-axis) for each of the five studied loss penalties for each of the three studied architectures. Optimal values are detailed in Tab. 1 and Fig. 2 in the main paper.

\section*{Plotting Train and Validation Accuracy During Training}

\begin{figure*}[t]
    \begin{subfigure}[b]{1\textwidth}
      \begin{subfigure}[b]{0.32\textwidth}
          \centering
            \includegraphics[width=1\columnwidth]{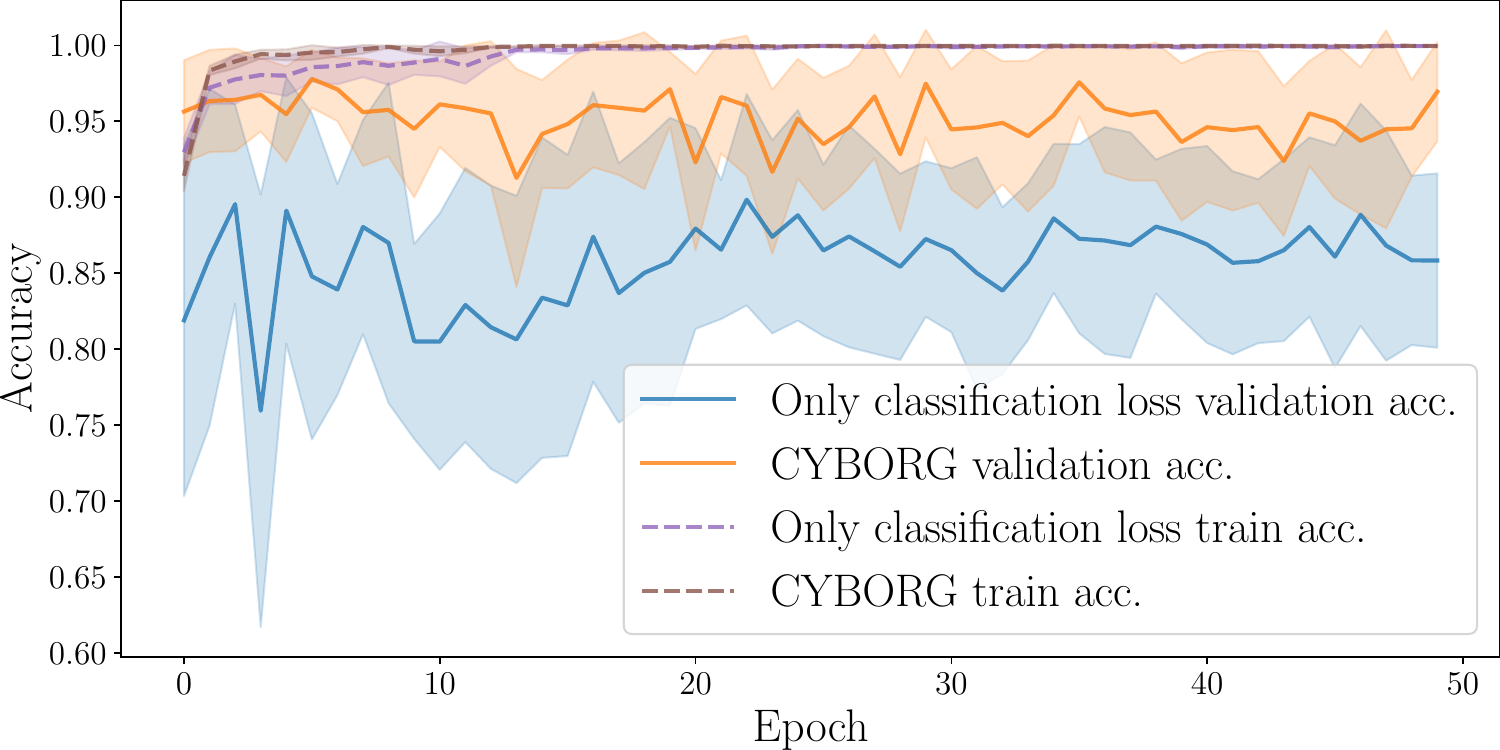}
            \caption{DenseNet121}
      \end{subfigure}
      \hfill
      \begin{subfigure}[b]{0.32\textwidth}
          \centering
          \includegraphics[width=1\columnwidth]{figures/face/resnet_training.pdf}
          \caption{ResNet50}
      \end{subfigure}
      \hfill
      \begin{subfigure}[b]{0.32\textwidth}
          \centering
          \includegraphics[width=1\columnwidth]{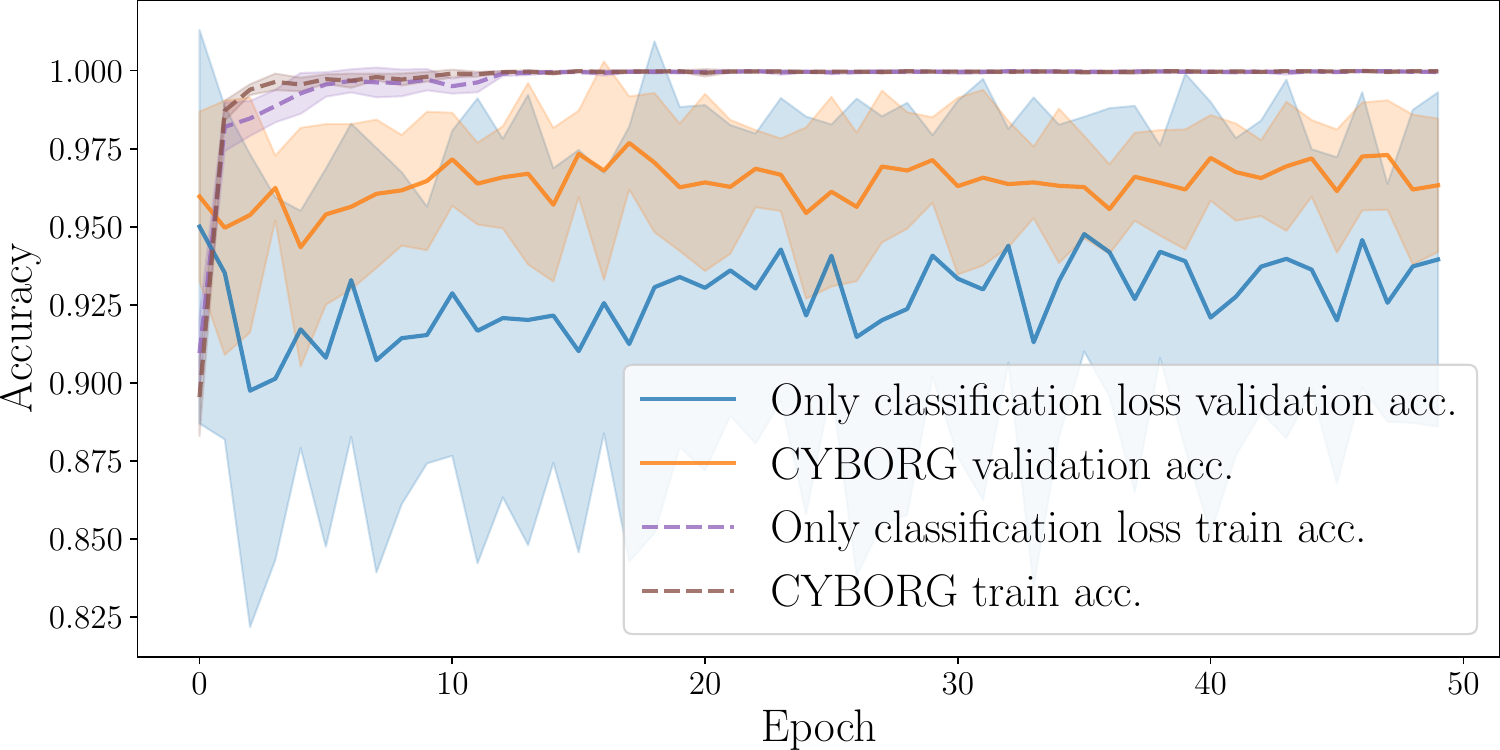}
          \caption{Inception v3}
      \end{subfigure}
    
  \end{subfigure} \vskip3mm
  \caption{Comparison of training and validation accuracy for CYBORG versus traditional training for \textbf{synthetic face detection}. 
  CYBORG training achieves higher validation accuracy, indicating more effective learning.
  Shaded area represents $\pm1$ standard deviation of the accuracy by epoch. }
  \label{fig:train_acc_plots_face}
\end{figure*}

\begin{figure*}[t]
    \begin{subfigure}[b]{1\textwidth}
      \begin{subfigure}[b]{0.32\textwidth}
          \centering
            \includegraphics[width=1\columnwidth]{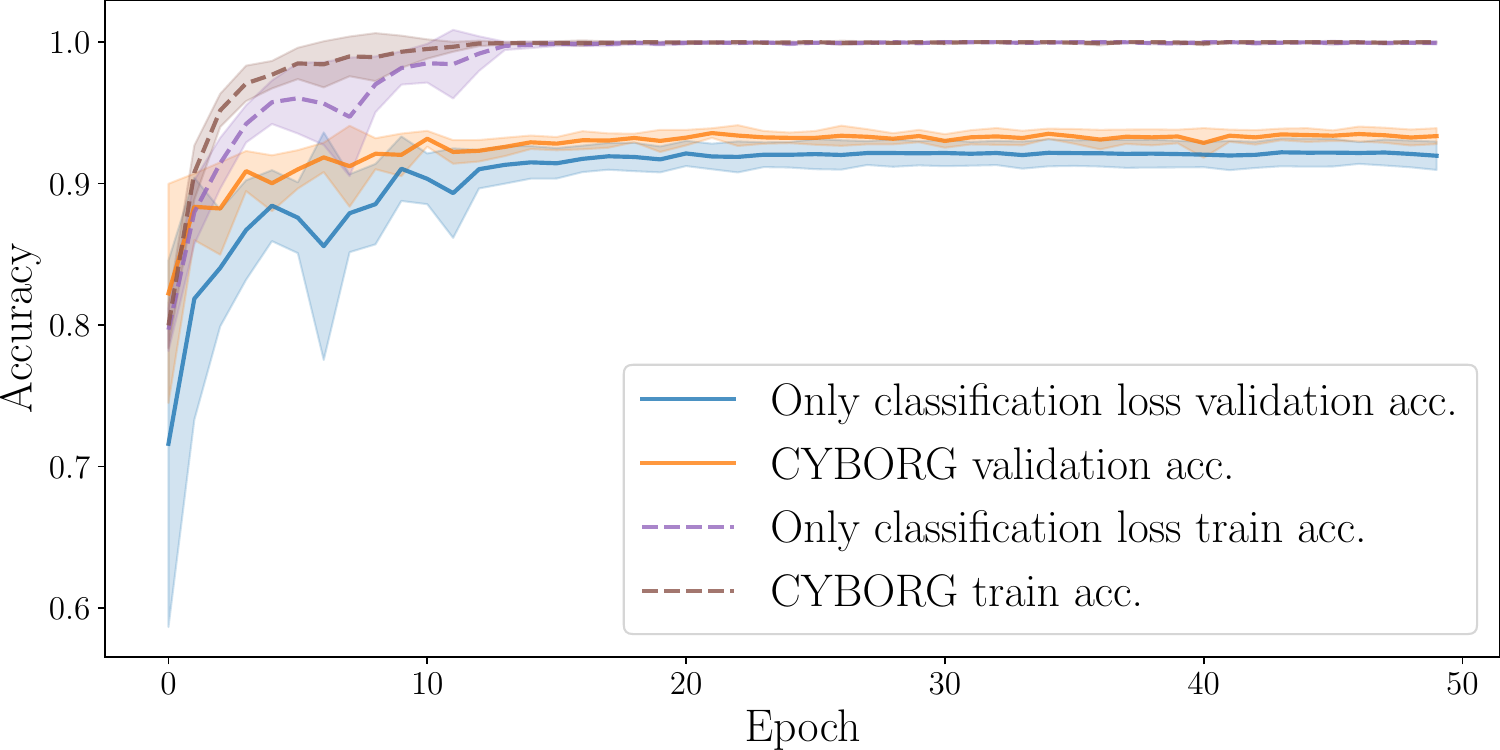}
            \caption{DenseNet121}
      \end{subfigure}
      \hfill
      \begin{subfigure}[b]{0.32\textwidth}
          \centering
          \includegraphics[width=1\columnwidth]{figures/iris/resnet_training.pdf}
          \caption{ResNet50}
      \end{subfigure}
      \hfill
      \begin{subfigure}[b]{0.32\textwidth}
          \centering
          \includegraphics[width=1\columnwidth]{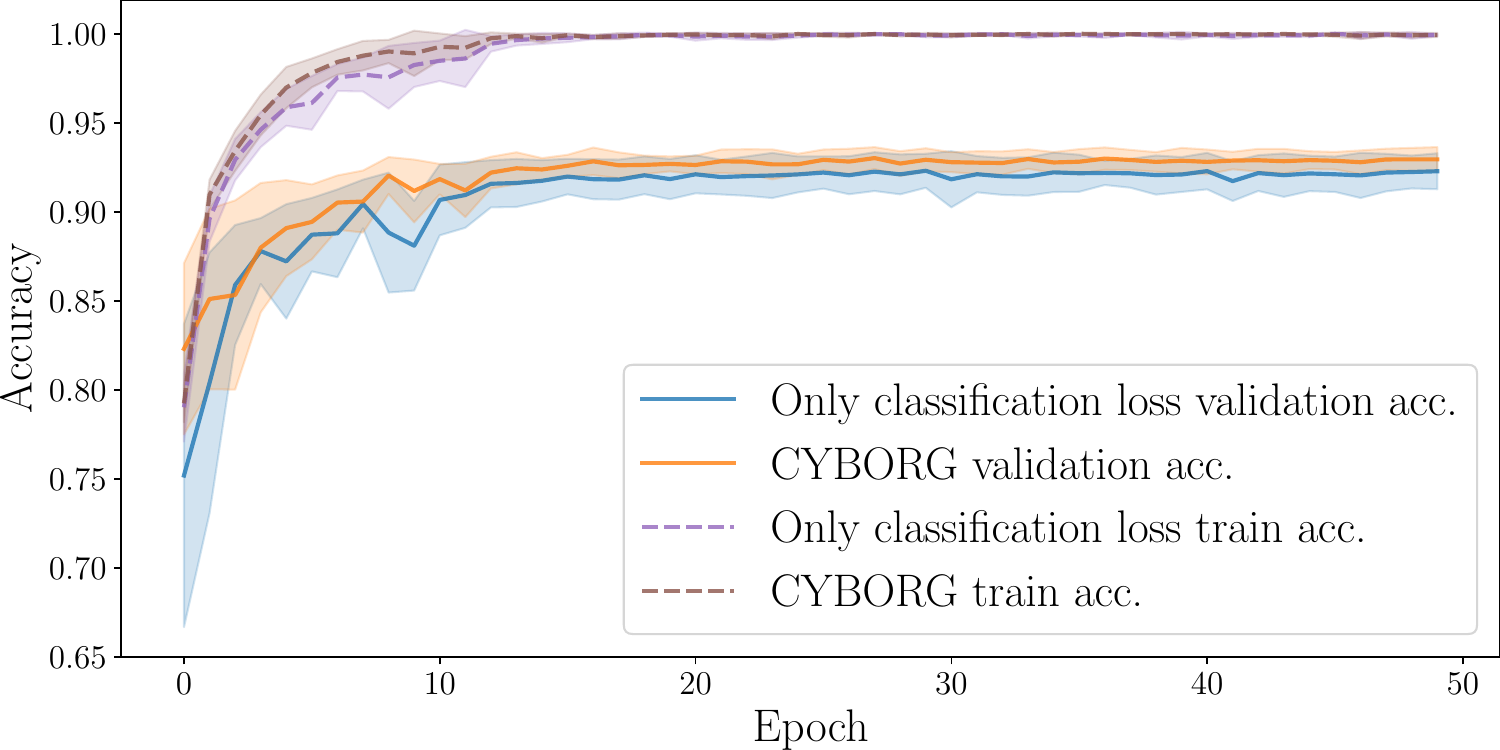}
          \caption{Inception v3}
      \end{subfigure}
    
  \end{subfigure} \vskip3mm
  \caption{Comparison of training and validation accuracy for CYBORG versus traditional training for \textbf{iris presentation attack detection}. 
  CYBORG training achieves higher validation accuracy, indicating more effective learning.
  Shaded area represents $\pm1$ standard deviation of the accuracy by epoch. }
  \label{fig:train_acc_plots_iris}
\end{figure*}

\begin{figure*}[t]
    \begin{subfigure}[b]{1\textwidth}
      \begin{subfigure}[b]{0.32\textwidth}
          \centering
            \includegraphics[width=1\columnwidth]{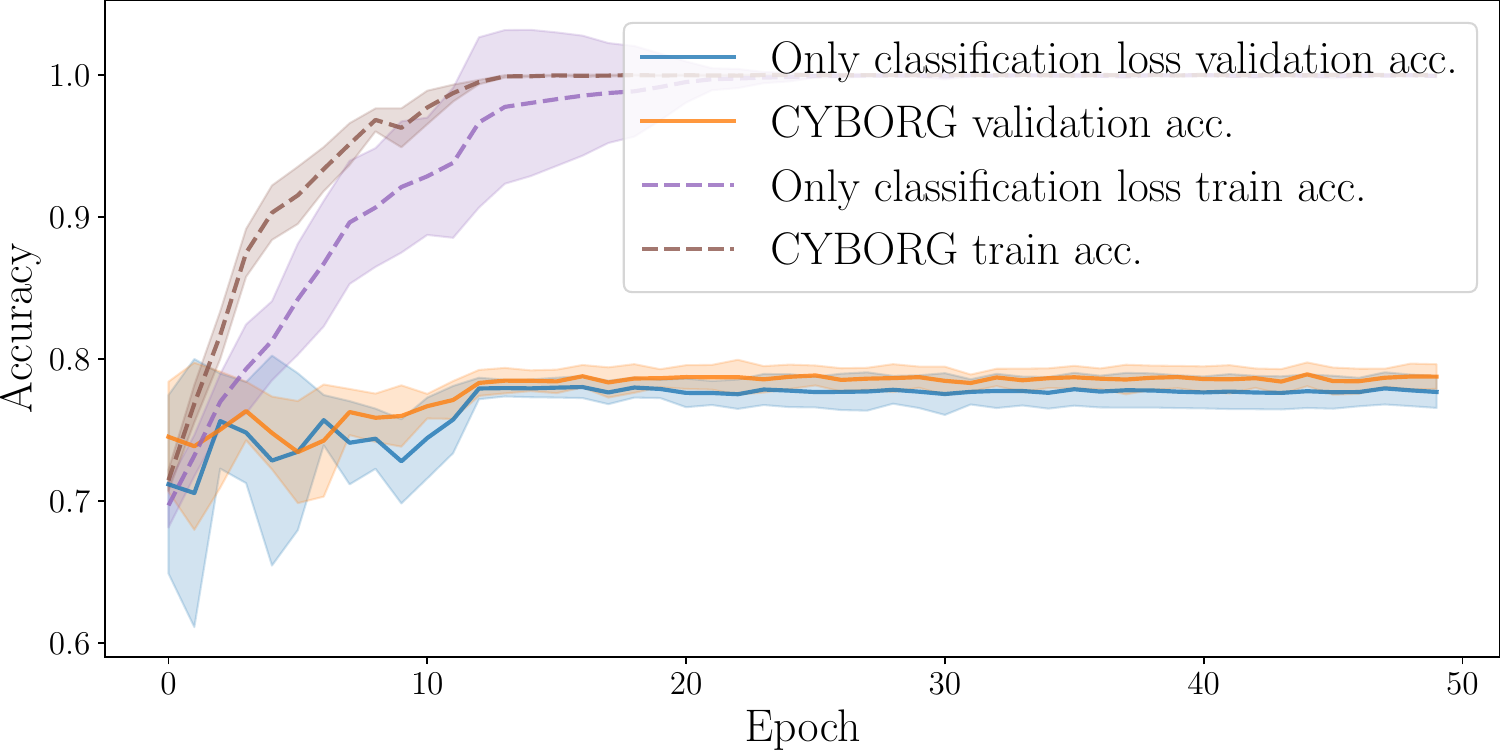}
            \caption{DenseNet121}
      \end{subfigure}
      \hfill
      \begin{subfigure}[b]{0.32\textwidth}
          \centering
          \includegraphics[width=1\columnwidth]{figures/cxr/resnet_training.pdf}
          \caption{ResNet50}
      \end{subfigure}
      \hfill
      \begin{subfigure}[b]{0.32\textwidth}
          \centering
          \includegraphics[width=1\columnwidth]{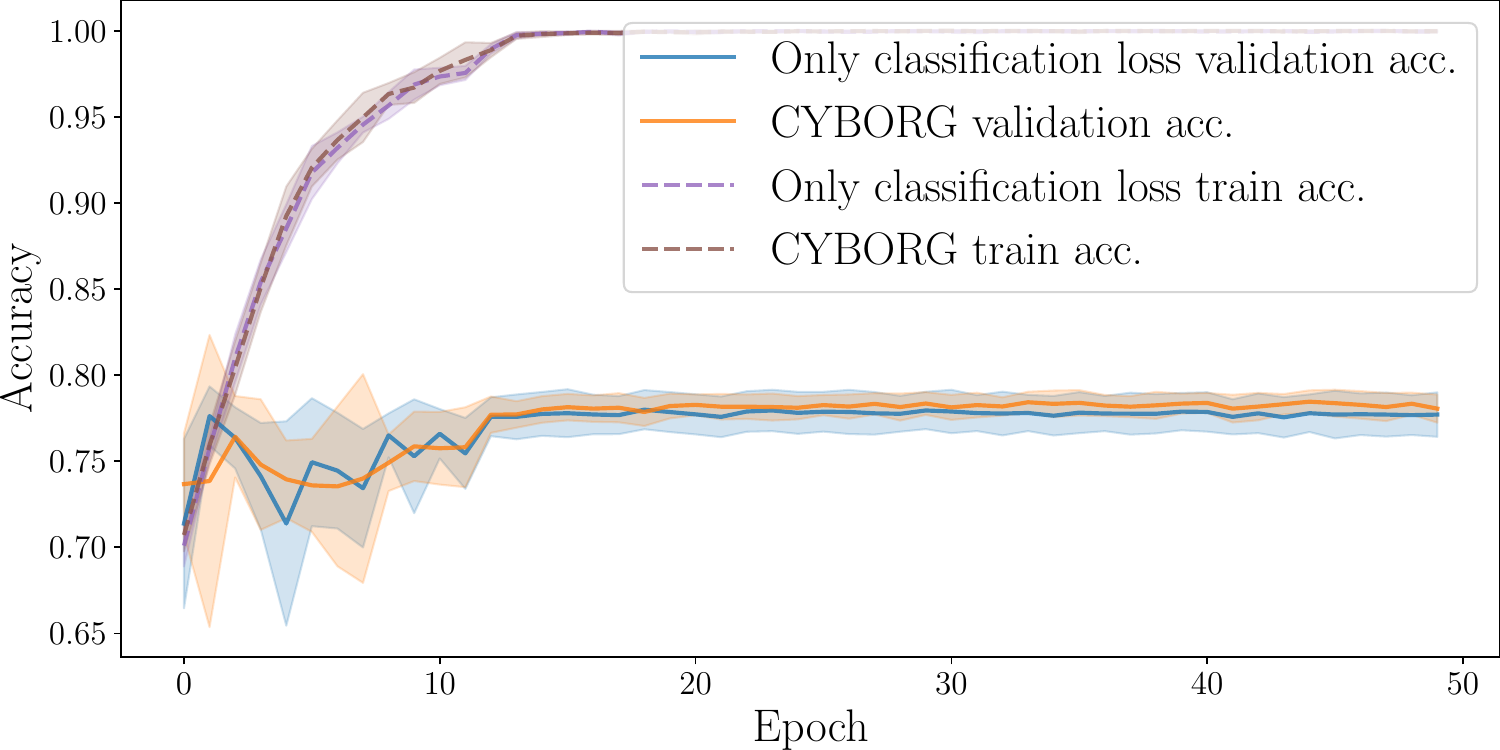}
          \caption{Inception v3}
      \end{subfigure}
    
  \end{subfigure} \vskip3mm
  \caption{Comparison of training and validation accuracy for CYBORG versus traditional training for \textbf{abnormality detection from chest x-ray}. 
  CYBORG training achieves higher validation accuracy, indicating more effective learning.
  Shaded area represents $\pm1$ standard deviation of the accuracy by epoch. }
  \label{fig:train_acc_plots_cxr}
\end{figure*}

The training and validation accuracy during training for synthetic face detection (Fig. \ref{fig:train_acc_plots_face}), iris presentation attack detection (Fig. \ref{fig:train_acc_plots_iris}), and abnormality detection from chest x-ray (Fig. \ref{fig:train_acc_plots_cxr}).

\subsection{Synthetic Face Detection Parameter Search}

\subsection{Iris PAD Parameter Search}

\subsection{Abnormality Detection Parameter Search}

\end{document}